\documentclass[lettersize,journal]{IEEEtran}

\usepackage{stfloats}

\usepackage{graphicx} 
\usepackage{xcolor}
\usepackage{amsmath}
\usepackage{amssymb}

\usepackage{multirow}
\usepackage{graphicx}
\usepackage{tcolorbox}

\usepackage{subfigure}
\usepackage{subcaption}
\usepackage{hyperref} 

\usepackage{todonotes}
\newcommand{\Rev}[1]{\textcolor{black}{{#1}}}

\newcommand{\TR}[2]{#2}
\newcommand{\MinRev}[1]{\textcolor{black}{{#1}}}

\begin{document}

\title{MetaSel: A Test Selection Approach for Fine-tuned DNN Models}

\author{Amin Abbasishahkoo,
        Mahboubeh Dadkhah,
        Lionel Briand,~\IEEEmembership{Fellow,~IEEE,}
        Dayi Lin
\thanks{Amin Abbasishahkoo is with the School of EECS, University of Ottawa,
Ottawa, ON K1N 6N5, Canada (e-mail: aabba038@uottawa.ca).}
\thanks{Mahboubeh Dadkhah is with the School of EECS, University of Ottawa,
Ottawa, ON K1N 6N5, Canada (e-mail: mdadkhah@uottawa.ca).}
\thanks{Lionel Briand is with the School of EECS, University of Ottawa, Ottawa, ON K1N 6N5, Canada, and also with the Research Ireland Lero centre for software, University of Limerick, Ireland (e-mail: lbriand@uottawa.ca).}
\thanks{Dayi Lin is with the Huawei Canada,
Kingston, ON K7K 1B7, Canada (e-mail: dayi.lin@huawei.com).}

\thanks{Manuscript received March 1, 2025.}}
\maketitle

\begin{abstract}
Deep Neural Networks (DNNs) face challenges during deployment due to \MinRev{covariate shift, i.e., data distribution shifts between development and deployment contexts}. Fine-tuning adapts pre-trained models to new contexts requiring smaller labeled sets. However, testing fine-tuned models under constrained labeling budgets remains a critical challenge. This paper introduces MetaSel, a new approach tailored for \MinRev{DNN models that have been fine-tuned to address covariate shift}, to select tests from unlabeled inputs. MetaSel assumes that fine-tuned and pre-trained models share related data distributions and exhibit similar behaviors for many inputs. However, their behaviors diverge within the input subspace where fine-tuning alters decision boundaries, making those inputs more prone to misclassification. Unlike general approaches that rely solely on the DNN model and its input set, MetaSel leverages information from both the fine-tuned and pre-trained models and their behavioral differences to estimate misclassification probability for unlabeled test inputs, enabling more effective test selection. Our extensive empirical evaluation, comparing MetaSel against \TR{C1.6, C3.8}{11} state-of-the-art approaches and involving 68 fine-tuned models across weak, medium, and strong distribution shifts, demonstrates that MetaSel consistently delivers significant improvements in Test Relative Coverage (TRC) over existing baselines, particularly under highly constrained labeling budgets. MetaSel shows average TRC improvements of 28.46\% to 56.18\% over the most frequent second-best baselines while maintaining a high TRC median and low variability. Our results confirm MetaSel's practicality, robustness, and cost-effectiveness for test selection in the context of fine-tuned models.

\end{abstract}

\section{Introduction}
\label{sec:Introduction}

Deep Neural Networks (DNNs) face challenges to widespread adoption, particularly due to data distribution shifts between development and deployment contexts. 
Training DNNs for new deployment contexts from scratch typically requires extensive labeled input sets, which can be prohibitively expensive. Transfer learning techniques, such as fine-tuning, have emerged as an effective and widely used solution by leveraging a model trained on a large and diverse input set, fine-tuning it to related but distinct contexts with substantially smaller labeled input sets. 
Testing fine-tuned DNN models faces similar labeling challenges, requiring effective strategies for selecting test subsets under highly constrained labeling budgets to ensure the reliability and performance of the fine-tuned model while addressing practical resource limitations.
In this paper, we introduce MetaSel, an effective test selection approach specifically designed for fine-tuned, classification DNN models that consistently outperforms State-of-the-Art (SOTA), general test selection approaches in terms of misclassification detection, given the same budget, particularly under highly constrained budgets. 
\TR{C2.1}{MetaSel is specifically designed and evaluated for a scenario where there is a data distribution shift between the development and deployment contexts, while the classification task and thus the model's output classes remain the same. This scenario, referred to as covariate shift, is framed as a key transfer learning scenario in the comprehensive survey by Pan and Yang~\cite{pan2010survey}, who classify covariate shift as a type of common and important domain adaptation.}

Several test selection approaches have been introduced for DNN models in recent years~\cite{hu2024test}. Some of these approaches, such as DeepGD~\cite{aghababaeyan2024deepgd}, are designed to maximize the identification of misclassification inputs within a predefined labeling budget. Another strategy involves leveraging prioritization approaches to rank inputs based on their probability of being misclassified and thus enable testers to select inputs according to available labeling budgets.  
An early contribution in this area is the Surprise Adequacy (SA) criteria introduced by Kim \textit{et al.}~\cite{kim2019guiding, kim2023evaluating}, which measures how surprising test inputs are to the model compared to its training inputs. 
Among the most effective and widely used strategies for test input prioritization are uncertainty-based approaches that aim to estimate the model's confidence in predicting unlabeled test inputs.
These strategies can be categorized into probability-based uncertainty metrics, which rely on the model's output probability vector~\cite{feng2020deepgini, Weiss2022SimpleTechniques, hu2024test, ma2021test}, and neighbor-aware uncertainty metrics, which make use of the input's nearest neighbors~\cite{li2024distance, bao2023defense}.
Recently, learning-based techniques~\cite{demir2024test, li2021testrank} have emerged that entail training an independent model to estimate the probability for an unlabeled test input to be misclassified by DNN Model Under Test (MUT).

Though existing test input prioritization approaches rely solely on MUT and its training set, there are application contexts, such as fine-tuned models, where more information is available for prioritization purposes.  
In this paper, we propose a new learning-based test input prioritization and selection approach tailored for fine-tuned models. Central to this approach is MetaSel, an independent model trained on the outputs of both the fine-tuned model and its pre-trained counterpart. MetaSel leverages the relationship between a fine-tuned model and its pre-trained counterpart to estimate the probability that an unlabeled test input will be misclassified by the fine-tuned model. Fine-tuning is particularly effective when the source and target contexts share related input distributions, resulting in fine-tuned and pre-trained models often exhibiting similar behaviors across many inputs in the new context. This shared behavior reflects the transferred knowledge from the pre-trained model. 
When both models exhibit similar outputs, the predictions are more likely to be correct. In contrast, when the behavior of the pre-trained and fine-tuned models differs, regardless of how we measure this divergence, the corresponding inputs can be expected to lie where the decision boundary has changed between the two models and thus are more likely to be misclassified.
Building on this, MetaSel relies on information from the fine-tuned model, its pre-trained counterpart, and their differences in behavior.
Our results demonstrate that by integrating such insights, MetaSel consistently outperforms SOTA baselines regarding test selection for fine-tuned models.

To comprehensively evaluate MetaSel, we conducted an extensive empirical study, comparing it against \TR{C1.6, C3.8}{11} SOTA test input selection baselines. Additionally, we analyzed the impact of varying levels of data distribution shifts on MetaSel's performance. Specifically, our experiments involved 68 fine-tuned models and corresponding input sets, which were subjected to weak, medium, and strong severity levels of distribution shift between the pre-trained and fine-tuned contexts. 
\Rev{Performing such experiments took approximately 48 days (not accounting for interruptions) using a cloud computing environment provided by the Digital Research Alliance of Canada~\cite{computecanada}. This extended duration was} due to both the large number of baselines and subjects, as well as the time-consuming nature of deploying some of the investigated baselines, like the learning-based approach proposed by Demir \textit{et al.}~\cite{demir2024test}. Furthermore, we included deeper networks with more complex architectures in our experiments, such as ResNet152, and input sets with a larger number of output classes, such as Cifar-100 with 100 classes.
The results demonstrate that MetaSel is a robust solution, consistently outperforming all baselines in selecting more misclassifications within the same budget. Importantly, MetaSel maintains its performance effectiveness across different subjects, selection budgets, and distribution shift levels. Furthermore, our analysis reveals that the second-best performing approach varies across subjects, severity levels, and test selection budgets. This variability highlights MetaSel’s reliability as the only approach that delivers consistently superior performance across diverse conditions.

It is important to note that while all our evaluations in this paper were conducted on image classification tasks, the scope of MetaSel is not limited to image-based inputs. MetaSel can be applied to other data types with limited adjustments, primarily in estimating how the behavior of the pre-trained and fine-tuned models differ on a given input. The core concept underlying MetaSel---leveraging the behavioral relationship between a fine-tuned DNN model and its pre-trained counterpart to enhance test selection---is broadly applicable to classification tasks beyond image input sets.
However, we focus on classification tasks, as MetaSel inherently designed around features that are well-defined and meaningful in classification settings such as logits---the raw, unnormalized scores before softmax activation layer---which are not present in regression models.
Adapting MetaSel for regression tasks would require the development of alternative strategies for assessing behavioral divergence between the pre-trained and fine-tuned models.

In the context of fine-tuned models, the effectiveness of test selection approaches becomes particularly critical when the selection budget is highly constrained. To evaluate this, we conducted an in-depth analysis of MetaSel’s performance compared to SOTA baselines under selection budgets of 1\%, 3\%, 5\%, and 10\% of the unlabeled target test set’s size. We analyzed MetaSel's effectiveness both in absolute terms---by measuring its achieved test Relative Coverage (TRC), capturing the capacity of a technique to identify misclassified inputs within a budget---and in relative terms, by evaluating its relative TRC improvement over the SOTA baselines. Our results show that MetaSel remains effective, consistently achieving a high TRC median with low variability, across different subjects, selection budgets, and levels of data distribution shift between the pre-trained and fine-tuned models. Moreover, MetaSel demonstrates significant beneficial performance compared to baselines, delivering an average TRC improvement percentage ranging from 28.46\% to 56.18\% over the most frequently occurring second-best baselines.

Additionally, we evaluated the efficiency of MetaSel, by comparing its execution time with that of SOTA baselines. Our analysis reveals that while MetaSel incurs a small increase in runtime compared to the most efficient baselines, such as probability-based uncertainty metrics~\cite{feng2020deepgini}, it remains highly practical, even for large input sets and deep networks. The slight computational overhead is more than justified by MetaSel’s ability to deliver substantial improvements in test input selection. Since MetaSel consistently provides robust and reliable performance improvements across a wide range of subjects, varying levels of distribution shift severity, and selection budgets, it provides a good trade-off between execution time and effectiveness.

To summarize, the key contributions of this paper are as follows:
\begin{itemize}
    \item We introduce MetaSel, a new learning-based approach for test input prioritization and selection, specifically designed for fine-tuned DNN models. MetaSel leverages the relationship between a fine-tuned model and its pre-trained counterpart to enhance the estimation of the probability that an unlabeled test input will be misclassified by the fine-tuned model.

    \item We evaluate MetaSel against \TR{C1.6, C3.8}{11} SOTA test selection approaches for DNNs as baselines. The results indicate that MetaSel consistently outperforms all baselines by detecting a higher number of misclassified inputs within the same selection budget. Moreover, the findings underscore MetaSel's reliability, as the second-best performing approach fluctuates across different subjects and selection budgets, further confirming that MetaSel is the only consistent and dependable solution across various scenarios.

    \item Our extensive empirical study involving 68 subjects---encompassing weak, medium, and strong distribution shift severity levels between pre-trained and fine-tuned models---demonstrates that MetaSel's performance remains consistent, confirming MetaSel's robustness regardless of the shift severity level.

    \item We evaluate the execution time of MetaSel in comparison to baseline approaches, confirming its practicality and cost-effectiveness, even when handling large input sets and deep complex network architectures.

    \item \TR{C1.4, C2.2, C3.2}{Our comprehensive ablation study validates the contribution of each training feature in MetaSel's performance, thereby confirming the effectiveness of the selected feature set.}
\end{itemize}

The remainder of this paper is organized as follows: Section II provides background and a concise review of test selection baselines utilized in our experiments. Section III defines the problem and outlines our research objectives. Section IV introduces MetaSel. Section V describes the experiments we performed to assess MetaSel. Section VI discusses the results for each research question. Section VII discusses related work, and Section VIII concludes the paper.

\section{Background} 
\label{sec:Background}

This section provides an overview of \TR{C1.6, C3.8}{the five} main categories of SOTA prioritization approaches used as baselines in our evaluations. These approaches, known for their effectiveness and efficiency, include surprise-based criteria, probability-based uncertainty scores, neighbor-aware uncertainty scores, learning-based methods, \TR{C1.6, C3.8}{and diversity-based approaches.} 

For the reasons discussed below, we deliberately exclude some of the existing methods from our evaluations, namely mutation-based~\cite{wang2021prioritizing} techniques, due to their computational impracticality in the context of fine-tuned models and the enormous computational impact they would have on our experiments. 
\Rev{Mutation-based approaches, such as }PRIMA (PRioritizing test inputs via Intelligent Mutation Analysis)~\cite{wang2021prioritizing}, aim to select test inputs that kill the most mutants under the assumption that such inputs are more likely to be mispredicted. 
However, mutation-based techniques are among the most time-consuming approaches as they require generating mutants and executing each unlabeled test input against all mutants. 
The scalability challenges of PRIMA have been empirically demonstrated by Hu \textit{et al. }~\cite{hu2023evaluating}, who reported a drastic drop in efficiency when this method was applied to larger datasets such as Cifar-100. Their evaluations show that deploying PRIMA for test selection on Cifar-100 requires execution times 50 times longer, respectively, than Distance-based SA (DSA)~\cite{kim2019guiding}, one of the computationally expensive baselines considered in our study. In our specific context, where a large number of unlabeled test inputs can be collected from the target context and is thus expected, as detailed in Section~\ref{sec:Objectives}, these approaches often become impractical. Furthermore, Hu \textit{et al. }\cite{hu2023evaluating} reported that simple probability-based approaches such as DeepGini consistently outperform PRIMA, which are much more computationally efficient. These findings further reinforce the rationale for excluding these methods from our study.

\subsection{Surprise-based approaches}
Kim \textit{et al.}~\cite{kim2023evaluating} proposed three Surprise Adequacy (SA) metrics to measure how surprising a test input is with respect to the DNN model's training input set: 

\textbf{Distance-based SA (DSA)} employs Euclidean distance to compare the activation traces of test inputs with those observed during training. This approach is tailored to classification tasks, as inputs near decision boundaries are prioritized for testing. 

\textbf{Likelihood-based SA (LSA)} uses Kernel Density Estimation (KDE)~\cite{wand1994kernel} to estimate the likelihood of the activation patterns of the test inputs given the distribution of activation patterns of the training inputs. Inputs with lower likelihoods are considered more surprising, indicating that the DNN is less familiar with such inputs, increasing the chances of being mispredicted.

\textbf{Mahalanobis Distance SA (MDSA)} was proposed recently and utilizes the Mahalanobis distance to measure the difference between activation traces of the test input and those of the training inputs. 

We include all three metrics in our experiments.

\subsection{Probability-based uncertainty metrics}
\label{sec:ProbabilityBacedUncertainty}

These metrics measure the DNN model's confidence in predicting a given test input, exclusively relying on the model's predicted probability distribution~\cite{hu2024test}.
In our experiments, we employ DeepGini~\cite{feng2020deepgini}, Margin~\cite{hu2024test}, and Vanilla-Softmax~\cite{Weiss2022SimpleTechniques} metrics. Their calculation formulas are presented in Table~\ref{tab:uncertaintyScores}. In these formulas, $P_i$ is the probability that input $x$ belongs to class $C_i$, $C$ is the total number of output classes, and $m$ and $n$ correspond to the indices of the most and second-most probable classes, respectively. 

\begin{table}[b]
    \centering   
    \small
    \caption{Probability-based uncertainty scores}
    \resizebox{0.99\columnwidth}{!}{
    \begin{tabular}{|c|    c|  c|  c|    }
    \hline 
   
Score    &Gini (x)  &Vanilla (x) &Margin (x) \\  \hline        
Formula  &$1- \sum_{i=1}^{C} {P_i(x)}^{2}$ &$1 -  \max_{i=1}^{C} {P_i(x)}$ &$p_m(x) - p_n(x)$  \\     \hline
         
    \end{tabular}
    }
    \label{tab:uncertaintyScores}
\end{table}

DeepGini~\cite{feng2020deepgini} is a widely recognized metric that has demonstrated superior performance over coverage-based approaches~\cite{feng2020deepgini, Weiss2022SimpleTechniques}. It has been extensively used as a baseline in DNN testing research~\cite{aghababaeyan2024deepgd, gao2022adaptive, dang2023graphprior, hu2022empirical, wang2021prioritizing}.
DeepGini prioritizes inputs with a greater spread of softmax values across classes since they have a higher level of classification uncertainty.
The Vanilla Softmax metric~\cite{hu2023evaluating} is calculated by subtracting the highest activation value from the output softmax layer from 1, resulting in a straightforward measure of uncertainty. With minimal computational and theoretical complexity, it serves as a simple yet effective baseline for test input prioritization. Interestingly, Weiss \textit{et al.}~\cite{Weiss2022SimpleTechniques}, in their extensive study, reported that the more sophisticated DeepGini metric does not consistently outperform the Vanilla Softmax metric, highlighting the robustness of this simple approach.
The Margin score specifically measures the degree of separation between the top two predicted probabilities. While DeepGini has been shown to outperform other metrics in most scenarios, studies have reported cases where the Margin score yields better results on certain subjects~\cite{aghababaeyan2024deepgd, li2024distance}. Given these mixed findings and the complementary insights provided by each metric, we included all three uncertainty metrics---DeepGini, Vanilla Softmax, and Margin score---as baselines in our experiments to ensure a comprehensive evaluation.

\subsection{Neighbor-aware uncertainty metrics}
Recent studies have proposed leveraging information from a test input's nearest neighbors to provide an accurate estimation of the model's uncertainty when predicting that input~\cite{li2024distance, bao2023defense}. 
Bao \textit{et al. }~\cite{bao2023defense} proposed Nearest Neighbor Smoothing (NNS), a new test selection approach that combines the DNN model's prediction on a test input with its predictions for the input's nearest neighbors within the test set. To achieve this, they use a smoothed prediction distribution instead of relying solely on the probability distribution output of the DNN model on the test input. For each test input $x$, the new smoothed prediction distribution is calculated as follows: 

\begin{equation}
p = \alpha p_M(x) + (1 - \alpha) p_{kNN}(x).
\end{equation}

\noindent where $p_M(x)$ is the output probability distribution of input $x$ and $p_{kNN}(x)$ is the average probability distribution of $x$'s $k$-nearest neighbors which is calculated as follows:

\begin{equation}
p_{kNN}(x) = \frac{1}{k} \sum_{t \in N_k(x)} p_M(t).
\end{equation}

\noindent where $N_k(x)$ represents the list of $x$'s $k$-nearest neighbors in the test set determined using Euclidean or Cosine distance between the representation of test inputs. Subsequently, any of the existing probability-based uncertainty scores such as DeepGini, Margin, or Vanilla can be calculated using this smoothed probability distribution.

Li \textit{et al.}~\cite{li2024distance} proposed DATIS (Distance-Aware Test Input Selection), which takes a novel approach to defining the uncertainty score of an unlabeled test input without relying on the DNN model's output probabilities.
Instead, it calculates uncertainty solely based on the ground truth labels of the test input's nearest neighbors in the labeled training set. To achieve this, DATIS estimates the support of the training data for the DNN's prediction by computing the normalized sum of the exponentials of the distances between the test input and its nearest training neighbors. The support of the training data for predicting test input ${x}$ as part of class $c \in Y$ based on its $k$-nearest neighbors in the training set is estimated as follows:

\begin{equation}
p^*_c (x) = \frac{\sum_{t \in Train_k(x)} \exp\left(-\|z(x) - z(t)\|^2_2/ \tau\right) \mathbb{I}(y(t) = c)}
{\sum_{t \in Train_k(x)} \exp\left(-\|z(x) - z(t)\|^2_2/ \tau\right)}
\end{equation}

\noindent where $Train_k(x)$ represents the list of $x$'s $k$-nearest neighbors in the training set. $z(x)$ and $z(t)$ are the latent feature representation of the test input $x$ and its nearest neighbor $t$, respectively. $\sigma$ is a scaling parameter controlling the influence of distances in the latent space. $\mathbb{I}(.)$ is an indicator function that equals 1 if the predicted label of the training input $y(t)$ matches class $c$, and 0 otherwise. $\|z(x) - z(t)\|^2_2$ represents the distance between the latent representations of $x$ and $t$. After calculating $p^*_c (x)$ for each class $c \in Y$, an estimation $p^*(x) = \{p^*_1, p^*_2, \dots, p^*_C\}$ for the test input $x$ is obtained. Then the DATIS uncertainty score for $x$ is calculated as follows:

\begin{equation}
\text{DATIS}(x) = p^*_n / p^*_m
\end{equation}

\noindent where $m$ is the class predicted by the DNN model for input $x$, which may or may not correspond to the class with the highest support $p^*_c (x)$. 
$n  =\arg\max_{c \in \mathcal{Y}, c \neq m} p^*(x)$ denotes the most supported prediction distinct from the DNN predicted class $m$.  Consequently, a higher value of $\text{DATIS}(x)$ means that the DNN's prediction on $x$ has low support in the training data, thus, suggesting a higher probability of being misclassified.

\subsection{Learning-based prioritization}
\label{sec:BackgroundLearningBased}
In recent years, learning-based test input prioritization approaches~\cite{hu2024test} have gained attention for DNN testing. These methods rely on training a model on top of the MUT to estimate the misprediction probability for each test input, leveraging diverse information for enhanced prediction accuracy. 
Various learning-based approaches have been proposed utilizing different sources of information, such as mutation testing results~\cite{wang2021prioritizing, dang2023graphprior}, ensemble model outputs~\cite{demir2024test}, and test inputs' nearest neighbors~\cite{li2021testrank}. In our study, we exclude learning-based approaches that either involve high computational costs, such as PRIMA~\cite{wang2021prioritizing} or are not designed for DNNs, such as GraphPrior~\cite{dang2023graphprior}, which was developed specifically for Graph Neural Networks (GNNs).
In our evaluations, we consider two SOTA learning-based approaches, namely TestRank~\cite{li2021testrank} and Meta-model~\cite{demir2024test}.

TestRank~\cite{li2021testrank}, proposed by Li \textit{et al.}, is one of the well-known learning-based methods that employ a Multi-Layer Perceptron (MLP) to effectively rank test inputs based on their misclassification probability. TestRank involves extracting two categories of features from each test input, namely intrinsic and contextual attributes. Intrinsic attributes are derived from the outputs of the MUT's logit layer for the given test input, while contextual attributes capture the relationships between the test input and its $k$-nearest neighbors. By constructing a graph where nodes represent inputs and edges encode their similarities (measured using a cosine distance metric), TestRank leverages GNNs to extract the contextual attributes of unlabeled test inputs. A simple binary classifier is then trained to predict the probability of an unlabeled input being misclassified by MUT. 

Demir \textit{et al.}~\cite{demir2024test} introduced Meta-model, a new learning-based test prioritization approach that enhances uncertainty scores by combining the uncertainty of the MUT with variations observed in the outputs of Deep Ensemble (DE) models for a given input. The DE models share the same architecture as the MUT but are independently trained using different initial parameter values, introducing diversity into their predictions. A key feature of their approach is the DE variation score, which quantifies the total number of ensemble models that produce predictions different from the MUT’s prediction for a given input. This score captures the variability and disagreement within the ensemble, offering additional insights into the uncertainty of the MUT. They leverage the MUT's uncertainty score and the DE variation score to train a Logistic Regression model to estimate the probability that a given test input will be misclassified by the MUT.

\subsection{Diversity-based prioritization}
\Rev{These approaches} 
\TR{C1.6, C3.8}{aim to select a subset of test inputs that not only has a high potential to detect misclassifications but also exhibits broad input diversity. Robust Test Selection (RTS) introduced by Sun \textit{et al.}~\cite{sun2023robust}, proposes a new probability-tier-matrix-based test selection metric. RTS is a hybrid method that integrates diversity, probability-based uncertainty, and noise filtering into a heuristic test prioritization framework. In this recent approach, the model's output probability distributions are used to divide test inputs into three distinct sets: noise, suspicious, and successful (correctly classified) inputs. Noise inputs are identified based on their dissimilarity to the model's training set based on the structural similarity index (SSIM)~\cite{wang2004image}. To identify successful inputs that are likely to be correctly classified by the model and suspicious inputs that are likely to be misclassified, majority voting from similar training inputs is used. Within each divided set, RTS uses a novel probability-tier-matrix-based metric to measure the output diversity among test inputs and combines it with uncertainty to guide a greedy selection process. Test inputs are then selected sequentially from these sets in a fixed order of priority: first from suspicious, then from successful, and finally from noise inputs. }


Adaptive Test Selection (ATS)~\cite{gao2022adaptive}, proposed by Gao \textit{et al.}, adopts an incremental selection strategy to construct a diverse subset of test inputs by iteratively selecting inputs that are the most different from the already selected subset. This approach, however, has been reported as one of the most computationally intensive approaches~\cite{li2024distance, hu2023evaluating}, as it involves calculating pairwise differences between each remaining unlabeled input and the currently selected test subset, thus growing quadratically in cost with the size of the unlabeled test set. \Rev{Hu \textit{et al.}~\cite{hu2023evaluating} reported a significant efficiency drop when applying ATS to larger datasets like Cifar-100, requiring 80 times the execution time of DSA~\cite{kim2019guiding}, which is already one of the most computationally intensive baselines in our study. More importantly, their findings revealed that simple probability-based methods, such as DeepGini, consistently outperformed ATS. Given that MetaSel consistently outperforms DeepGini across all subjects and test selection budgets, as demonstrated in our evaluations, we exclude ATS from our experiments. }


\section{Problem Definition}
\label{sec:Objectives}

\begin{figure}
    \includegraphics[width = \columnwidth]{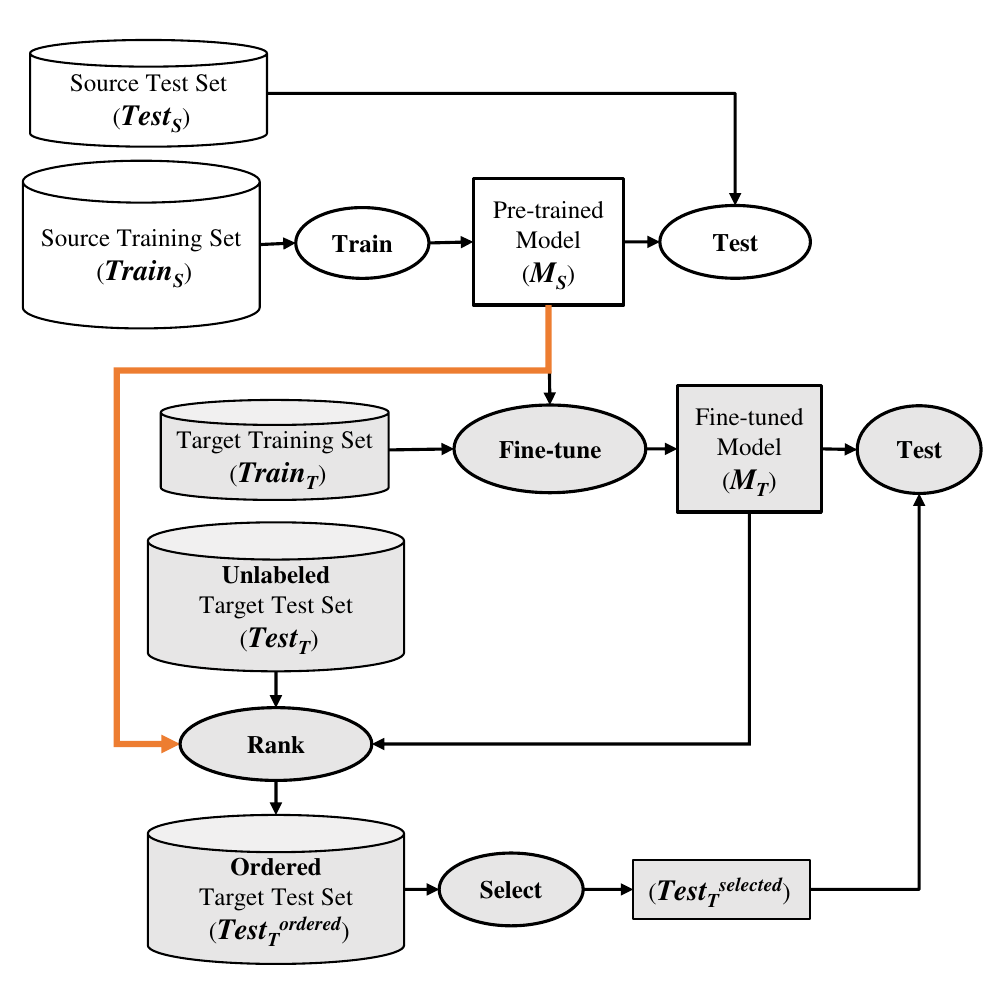}
    \centering
    \caption{Test input ranking and selection for a fine-tuned model ($M_T$) using its corresponding pre-trained model ($M_S$)}
    \label{fig:Objectives}
\end{figure}

In practice, pre-trained DNN models often need to be adapted before deployment in a new context. This typically involves fine-tuning a generic model, trained on a large number of inputs, with a smaller training set specific to the target deployment context. Fine-tuning may also be motivated by observing changes in the deployment context, making it inconsistent with the development context. Such changes can arise from various factors, such as shifts in user behavior or seasonal data patterns, and are referred to as distribution shifts between the development and deployment contexts~\cite{hu2022empirical}. 
The need to fine-tune a pre-trained model for a new context typically arises from a scarcity of labeled target inputs necessary for training a model from scratch. Indeed, despite the availability of a large pool of unlabeled inputs, labeling in many contexts is an expensive activity~\cite{hu2024test}. Similarly, testing a fine-tuned model entails the same challenge with labeling test inputs. As a result, an effective test input selection approach is essential. The ultimate objective of such an approach in this context is to prioritize the labeling of a very small number of test inputs that are more likely to be misclassified by the fine-tuned model and thus minimize the effort required for testing such models across various deployment contexts.

Let us assume a source input set $X_S$ and a related yet distinct target set $X_T$, where the corresponding data distributions of $X_S$ and $X_T$ are assumed to differ from each other and are denoted as $D_S$ and $D_T$, respectively. Let $f_S: x \rightarrow y$ refer to a pre-trained DNN model $M_S$ performing a classification task over input set $X_S$ and labels $Y$, trained on the source training set $Train_S \subset X_S$ with $x \in X_S$ and $y \in Y$. 

Preceding the deployment of the pre-trained model $M_S$ in the target context, the common practice is to fine-tune it by continuing the training process for an additional number of epochs~\cite{hu2022empirical} with a set of labeled inputs from the target set, referred to as the target training set ($Train_T  \subset X_T$). Due to the labeling cost, the size of the target training set is limited and often significantly smaller than the source training set ($Train_T \ll Train_S$).
This fine-tuning process aims to address the distributional shift between $X_S$ and $X_T$ and results in a new model $M_T$ denoted as $f_T: x \rightarrow y$ with $x \in X_T$ and $y \in Y$. We refer to this as the fine-tuned model and assume it has higher accuracy overall on the target input set than the pre-trained model. 

We also assume that a large target test set $Test_T \subset X_T$, containing unlabeled inputs, can be collected from the target context. However, labeling all these test inputs is expensive and time-consuming in most practical situations. 
Therefore, it is essential to prioritize a small-enough subset of unlabeled inputs in $Test_T$ to assess the performance of the fine-tuned model while keeping the testing effort within acceptable bounds. Test prioritization approaches assign a score to all inputs in the target test set, indicating their relative likelihood of being misclassified by the fine-tuned model, resulting in an ordered list of target test inputs $Test^{ordered}_T$. However, ranking the entire target test set is not the primary goal. The ultimate objective in the context of fine-tuned models is, for a given test budget, to select an effective subset of target test inputs $Test^{selected}_T$ for labeling and testing the fine-tuned model.

In this paper, we propose a new approach to effectively rank and select a small subset of test inputs targeting a fine-tuned DNN model ($M_T$) based on information about its pre-trained counterpart ($M_S$). As shown in Figure~\ref{fig:Objectives}, we utilize the pre-trained model to rank target test inputs based on their probability of being misclassified by the fine-tuned model. 
To align with the test selection objective in this context, we presume a constrained test input labeling budget in our evaluations and compare the effectiveness of our approach with SOTA baselines across various realistic budgets for target test inputs.

\begin{figure*}
    \includegraphics[width = \textwidth]{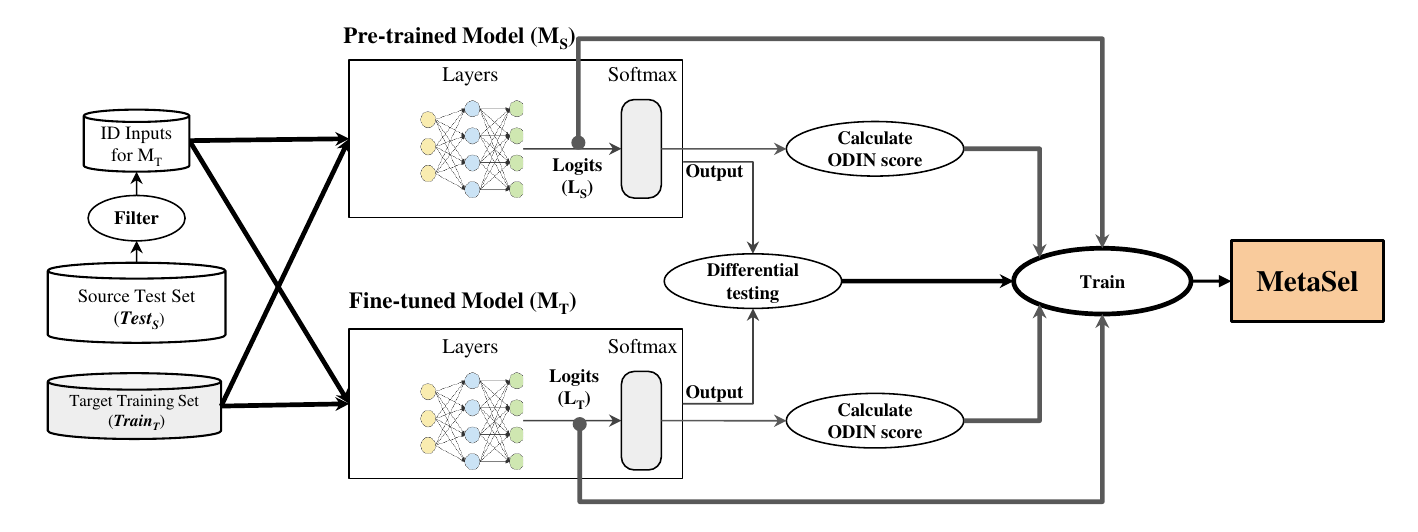}
    \centering
    \caption{The process of training MetaSel}
    \label{fig:MetaSel}
\end{figure*}

\vspace{-10pt}
\section{MetaSel}
\label{sec:MetaSel}

In this section, we propose a new learning-based method for ranking and selecting target test inputs to test a fine-tuned model, leveraging its counterpart pre-trained model. 
Unlike existing general approaches, MetaSel is uniquely tailored for fine-tuned models. While the former offers the advantage of broader applicability across diverse contexts, they fail to leverage the additional artifacts and information available in our specific context, which could significantly enhance the fine-tuned model's testing.

We first conducted a preliminary study to identify relevant artifacts that could be effectively leveraged for test selection in the context of fine-tuned models. This preliminary study was conducted on a limited number of input sets and DNN models, to save time and resources. Building on our initial findings, we developed MetaSel, a solution that leverages the pre-trained model's outputs and its input set alongside the fine-tuned model's outputs, resulting in a significant improvement in target test selection effectiveness over existing general solutions.

MetaSel is a DNN model designed to estimate the probability that an unlabeled target test input will be misclassified by the fine-tuned model. 
\TR{C1.2, C3.3}{We experimented and found that a simple DNN architecture consisting of a 1D convolutional layer for feature extraction, followed by two or three fully connected layers, yields consistently strong results while maintaining efficiency.}
In contrast to simpler Neural Networks (NNs) with a shallow single-layer architecture, DNNs have the capability to capture non-linear relationships from complex input features.
\TR{C1.2, C3.3}{To ensure transparency and facilitate reproducibility, MetaSel models constructed for all investigated subjects are publicly available in our replication package~\cite{replicationpackage}. We should note that using a different or more complex architecture may potentially increase MetaSel’s effectiveness while increasing its training and test selection time. }

We define MetaSel as a function $f_{m}: X_{m} \rightarrow [0, 1]$ where $X_{m}$ represents the input features and the output $f_{m}(X_{m})$ indicates the predicted probability of an input being misclassified by the fine-tuned model. Consequently, $f_{m}(X_{m})$ values closer to 1 indicate a higher likelihood of misclassification.

To predict the outcome for a given test input, MetaSel relies not only on the output of the fine-tuned model but also on the corresponding output of the pre-trained model for that input, since to some extent, knowledge is transferred from the latter to the former. Additionally, to improve MetaSel's training effectiveness, we augment its training set by incorporating a subset of inputs from the source test set $Test_S$ that closely aligns with the target data distribution.
The detailed procedure for constructing MetaSel, including feature selection and training set augmentation, is described in the following sections.

\subsection{Determining MetaSel's feature set}
\label{sec:MetaSelFeatureSet}

To identify an effective set of input features $X_{m}$ for training a model accurately predicting misclassified inputs, we conducted a preliminary study exploring various feature sets. We considered features extracted from the target training inputs (images in our experiments), the final class probabilities of the source ($M_S$) and target ($M_T$) models, and their output labels, as well as a measure of how closely an input conforms to the data distribution on which the source and target DNN models were trained. For the latter, we employed ODIN (Out-of-DIstribution detector for Neural networks)~\cite{liang2017enhancing}, a highly effective yet efficient approach that achieves robust Out-of-Distribution (OOD) detection without requiring any architectural modifications or additional training.

\TR{C1.2}{While MetaSel can be configured to deploy any OOD detection approach, it is crucial to use lightweight solutions to maintain MetaSel's efficiency while ensuring its effectiveness. For instance, Likelihood Regret (LR) OOD~\cite{xiao2020likelihood} has demonstrated higher accuracy than ODIN across a broad range of input sets~\cite{xiao2020likelihood} and could potentially enhance MetaSel’s effectiveness. However, LR OOD entails significantly higher computational cost, which would make MetaSel less efficient, particularly when applied to larger input sets or complex DNNs.
}
Based on the results of our preliminary study and the reasons discussed below, we considered the following features, as illustrated in Figure~\ref{fig:MetaSel}:

\begin{itemize}
    \item The output of the logit layer of $M_S$ (\TR{C1.2}{vector} $L_S$),
    \item The output of the logit layer of $M_T$ (\TR{C1.2}{vector} $L_T$), 
    \item \TR{C1.2}{A vector representing the element-wise absolute} difference between $L_S$ and $L_T$,
    \item The outcome of differential testing by comparing output labels of $M_S$ and $M_T$,
    \item The ODIN score of the given input calculated based on $M_S$, 
    \item The ODIN score of the given input calculated based on $M_T$.
\end{itemize}

At a high level, the assumptions underlying these features include (1) The more uncertain the predictions, the more likely they are to be incorrect, (2) If the two models' predictions are inconsistent, then the corresponding inputs lie in the input subspace where the decision boundary was changed as a result of fine-tuning and are more likely to be misclassified, (3) OOD inputs are more likely to be misclassified. 
Since $M_T$ is a version of $M_S$ specifically tailored to the target context, the two models share a foundational understanding through the knowledge transferred during fine-tuning. Thus, information from both models is potentially relevant.

We leverage logits since they allow for more fine-grained comparisons between the two DNN models and constitute more precise measurements of inconsistency or disagreement between predictions than the final class probabilities or output labels alone.
Compared to the output label for models $M_S$ and $M_T$, using the logit layers offers several advantages.
Labels are determined by selecting the class with the highest probability from the probability distribution produced by the last layer's activation function (softmax layer). In contrast, logits represent the raw scores produced by the layer preceding the softmax layer. Therefore, unlike the output label, which only indicates the predicted class, logits reflect the model’s raw scoring for each output class, providing a more complete view of the model’s internal decision-making.
Consequently, logits provide a richer representation of the pre-trained and fine-tuned models' behavior and a finer indication of their agreement or disagreement. For instance, consider the case where both the pre-trained and fine-tuned models produce the same output class for a given input, while one model produces logits indicating high confidence in its prediction and the other produces logits showing very close values for two or more classes, reflecting uncertainty. In this scenario, the output labels alone suggest agreement between the two models, whereas their logits reveal discrepancies in their prediction confidence. Consequently, we rely on logits as they provide a richer representation of the model's behavior compared to output labels.

The output of the logit layer in a DNN represents raw, unnormalized scores. These scores are then passed through a softmax activation function, which converts them into class probabilities. Class probabilities are normalized values of logits, restricted within the range of [0, 1], and collectively add up to one. However, this normalization procedure can potentially obscure the information originally contained in the logits. Indeed, Aigrain \textit{et al.}~\cite{aigrain2019detecting} further demonstrated that logits have greater discriminative power than softmax values, making the former more effective in detecting misclassifications and adversarial examples in DNNs.
Additionally, the normalization procedure amplifies the differences between logits, making class probabilities vulnerable to outliers present in the logits. In other words, outliers (i.e., extremely high or low logits) can disproportionately affect the output probabilities. In extreme cases, a single outlier can dominate the probability distribution, with its associated class probability approaching 1, while the probabilities for all other classes converge to 0.

However, to ensure that we use all the relevant information available, we also include in our feature set a boolean value indicating whether the output labels of $M_S$ and $M_T$ are identical or differ. \MinRev{This value, representing the differential testing outcome, is calculated solely based on the two models' predictions and assigned a value of 1 when both models predict the same class and 0 otherwise.}
Furthermore, for the reasons presented above, we also feed MetaSel the difference between the logits of two models for each output class, as this may enhance its discriminative power.
The difference between $L_S$ and $L_T$ indicates the magnitude of discrepancies between the prediction of the pre-trained and fine-tuned models for each output class, indicating whether the given input is near the input subspace where the decision boundary has changed.

Additionally, as mentioned earlier, we employ a measure of how closely an input aligns with the data distribution used to train the source and target DNN models, namely ODIN scores for both $M_S$ and $M_T$. 
Inputs that align with the training data distribution of a model are considered in-distribution (ID) and typically exhibit higher ODIN scores, whereas inputs that deviate from the training data distribution (i.e., OOD) tend to have lower ODIN scores.

In the context of fine-tuned models, we expect the data distributions for $M_S$ and $M_T$ not to be identical. However, fine-tuning adapts $M_S$ to a new context while retaining some prior knowledge. As a result, $M_T$ relies on partial knowledge transferred from $M_S$.
Consequently, we expect inputs with low ODIN scores for $M_T$ but relatively high ODIN scores for $M_S$ to have a lower probability of being misclassified than inputs with low ODIN scores for both $M_S$ and $M_T$. 
This is because the transferred knowledge from $M_S$ may still assist $M_T$ in handling inputs that align closely with $M_S$’s distribution, even if they don't with that of $M_T$. 
Furthermore, inputs with high ODIN scores for both $M_S$ and $M_T$, indicating that they closely align with both distributions, are expected to have the lowest probability of being misclassified.
Consequently, we assume that incorporating the ODIN scores of a given input for both $M_S$ and $M_T$ into the feature set can also significantly improve MetaSel's capacity to accurately predict the behavior of the target DNN model.

Notably, we also observed that incorporating representations of target input images derived from a pre-trained VGG-16~\cite{simonyan2014very} model in MetaSel's feature set did not improve its performance.
This may be explained by the fact that information about a given input image is already encoded in the outputs of the logit layer~\cite{zhao2022uniform}. 
\TR{C1.4, C2.2, C3.2}{
To validate the contribution of each selected feature to the overall performance of MetaSel, we conduct a comprehensive ablation study and report the results in the subsequent sections.}

It is important to note that while we validated our feature sets on image-based classification tasks, MetaSel's applicability is not limited to image inputs. MetaSel is designed as a general test selection approach and does not require any architectural modifications when applied to alternative input types, such as text or audio. MetaSel's core features---pre-trained and fine-tuned models' logits, their differences, and their output labels---are readily available from classification models across diverse input types. However, when measuring how well a given test input aligns with a DNN model's underlying data distribution, it is crucial to select an effective and lightweight OOD detection approach to maximize the efficiency and performance of MetaSel. 
In our experiments, we utilize ODIN, an OOD detection method originally developed for image classification with CNNs. ODIN leverages temperature scaling and gradient-based input perturbations applied to pixel inputs. While its core ideas are broadly applicable, they do not naturally translate to all data types and may not yield effective OOD detection performance in domains with different input structures. Therefore, when applying MetaSel on non-image inputs such as text or audio, it may be beneficial to replace the ODIN score with a more suitable OOD detection method tailored to that specific input type. 
Consequently, MetaSel is largely adaptable as it offers the flexibility to incorporate adjustments specific to each input type, further improving its potential effectiveness across diverse data domains.







\subsection{Constructing MetaSel's training set}
To train MetaSel, we require a set of inputs from the target distribution ($D_T$) whose corresponding MetaSel's outputs are known. To be able to determine MetaSel's output, we need to know the following information for each input ($t \in X_T$):
\begin{itemize}
    \item Output label of the fine-tuned model ($f_T(t)$), and
    \item Ground-truth label ($C_t$)
\end{itemize}

To address this challenge, we leverage the target training set ($Train_T \subset X_T$) to train MetaSel since their ground-truth labels are known.  However, in the context of fine-tuned models, the size of such a training set is typically limited since it is often significantly smaller than the source training set. Consequently, this can negatively affect MetaSel's ability to achieve high predictive accuracy. 

To enhance the training input set for MetaSel, additional inputs must meet two critical criteria: (1) their ground-truth labels must be available, and (2) they must closely align with the target set's data distribution. However, labeling new inputs, especially in the context of fine-tuned models, is challenging and resource-intensive. To overcome these limitations, we considered employing the source test set ($Test_S$) alongside the target training set ($Train_T$) to provide additional labeled inputs as illustrated in Figure~\ref{fig:MetaSel}. While all inputs in the source test set meet the first criterion of having known ground-truth labels, not all of them align closely with the target data distribution, as discussed in our problem definition (Section~\ref{sec:Objectives}). 

To address this, we employed once again ODIN scores~\cite{liang2017enhancing}, to filter out source test inputs that do not align with the target data distribution, $D_T$. Source test inputs with ODIN scores lower than a specific threshold are considered OOD and filtered out. To determine an optimized threshold for each $M_T$, we rely on the common procedure introduced by Liang \textit{et al.}~\cite{liang2017enhancing} that involves using a separate validation set. Following this procedure, we select a threshold that minimizes the False Positive Rate (FPR) at a fixed True Positive Rate (TPR) of 95\%, \Rev{as suggested in the original paper~\cite{liang2017enhancing}.} 
This threshold ensures that the model retains a high level of accuracy in distinguishing ID inputs from OOD inputs.
By applying this filtering mechanism, as illustrated in Figure~\ref{fig:MetaSel}, we ensured that only those source test inputs that satisfy both criteria were included in the MetaSel training set.

We then feed these inputs into both $M_S$ and $M_T$ to determine their outputs, which are then used to generate the required data to build MetaSel. The findings of our preliminary study provide additional evidence that, in the context of fine-tuned models, our procedure of filtering inputs $Test_S$ and utilizing inputs that align with $M_T$'s data distribution ($D_T$) to train MetaSel significantly enhances its accuracy in identifying inputs likely to be misclassified by $M_T$.

\section{Experiment Design}
\label{sec:ExperimentalProcedure}

In this section, we outline the methodology for a comprehensive empirical evaluation of MetaSel, comparing it to SOTA baselines described in Section~\ref{sec:Background} across a diverse set of subjects. This includes detailing the research questions guiding our study, the subjects, and the metrics used for evaluation. To answer the first two questions we use source and target input sets, the latter being generated with a medium level of distribution shift severity. To comprehensively address the last research question, we extend our experiments by including two additional levels of distribution shift severity: one representing a weaker shift and the other representing a stronger shift.

\subsection{Research Questions}
\label{sec:ResearchQuestions}

We performed an extensive evaluation to answer the following questions:

\textbf{\textit{RQ1: Does MetaSel demonstrate a significant improvement in performance compared to baselines in guiding test selection for fine-tuned models?}}

In this research question, we investigate the effectiveness of MetaSel in test selection and compare it against the baselines outlined in Section~\ref{sec:Background}.
To achieve this, we compare MetaSel's performance across different selection budgets, with a particular emphasis on smaller budgets, and assess whether the observed performance improvements are both statistically and practically significant. Specifically, we examine MetaSel's improvements under highly constrained budgets of 1\%, 3\%, 5\%, and 10\% of the target test set for each subject. This focus aligns with the practical applications of test selection approaches in the context of fine-tuned models, as outlined in Section~\ref{sec:Objectives}. Our goal is to determine whether, by relying on MetaSel, we can achieve superior performance than the baselines, particularly with highly constrained budgets.

\textbf{\textit{RQ2: Is MetaSel sufficiently efficient?}}

In this research question, we evaluate the efficiency of MetaSel by comparing its total execution time with baselines. 
First, we assess whether the time cost of MetaSel is practical for real-world applications. We then compare this cost with that of baseline methods, particularly simpler uncertainty-based approaches, in order to better understand the trade-offs between accuracy and efficiency.

\textbf{\textit{RQ3: Does MetaSel maintain its effectiveness across varying levels of distribution shift between the source and target sets?}}

In this research question, we investigate the impact of distribution shift severity on the experimental results. To this end, we apply MetaSel to target sets with two additional levels of distribution shift severity: one weaker and one stronger than those used in earlier research questions. This comprehensive analysis allows us to evaluate MetaSel's robustness across varying degrees of distribution shifts.

\Rev{\textbf{\textit{RQ4: How does each selected training feature contribute to MetaSel's performance?}}}

\TR{C1.4, C2.2, C3.2}{To address this research question, we perform a comprehensive ablation study to evaluate the contribution of each training feature to MetaSel’s effectiveness. Specifically, we construct multiple variants of MetaSel, each excluding one or a group of the original training features. We then compare the performance of the original MetaSel with these variants by analyzing the TRC they achieve.
Our goal is to demonstrate that MetaSel, in its original form, maintains superior performance across all variants.}


\subsection{Subjects}
\label{sec:Subjects}
We conducted our experiments using a collection of well-known image input sets, including MNIST~\cite{deng2012mnist}, Fashion-MNIST~\cite{deng2012mnist}, Cifar-10~\cite{krizhevsky2009learning}, and Cifar-100~\cite{krizhevsky2009learning}, that have been utilized in numerous empirical studies on DNN testing~\cite{shen2020multiple, aghababaeyan2024deepgd, abbasishahkoo2024teasma, hu2022empirical, berend2020cats}. 
MNIST is a set of images of handwritten digits, and Fashion-MNIST (FMNIST) contains images of clothes that are associated with fashion and clothing items. Each one represents a 10-class classification problem and consists of 70,000 grayscale images of a standardized 28x28 size. The training set of each input set includes 60,000 images, while each test set includes 10,000 images. Cifar-10 is another widely used input set that includes a collection of images from 10 different classes (e.g., cats, dogs, airplanes, cars). Furthermore, we selected a more complex input set, Cifar-100, containing 100 classes with 60,000 images total, known for its realistic and feature-rich images. Both Cifar-10 and Cifar-100 are widely used benchmarks in computer vision and contain 32x32 cropped colored images. Each input set is divided into 50,000 images for training and 10,000 images for testing.

Using these input sets, for our experiments, we trained four source DNN models using LeNet5~\cite{lecun1998gradient}, ResNet20~\cite{he2016deep}, and ResNet152~\cite{he2016deep}, as detailed in Table~\ref{tab:SourceModels}. Note that, to avoid any selection bias, we reused the training and test sets defined in the original sources~\cite{deng2012mnist, krizhevsky2009learning}.

\begin{table}[t]
    \centering   
    \footnotesize 
    \caption{Information about the input sets and DNN models}
    \resizebox{\columnwidth}{!}{
    \begin{tabular}{|c|      c|  c|  c|    }
    \hline 
         Model    &$Train_S$   &size $Train_S$   &size $Train_T$  \\ \hline 
          LeNet5  &MNIST &60,000    & 6,000 \\    \hline
          LeNet5  &FMNIST &60,000  &  12,000 \\     \hline
          ResNet20  &Cifar-10  &50,000   & 10,000 \\  \hline
          ResNet152  &Cifar-100 & 50,000    &  20,000 \\  \hline
    \end{tabular}
    }
    \label{tab:SourceModels}
\end{table}

To perform our experiments, we require target input sets that have a data distribution different from the source distribution. 
To achieve this, we leverage the corrupted versions of our source test sets as target test sets, which include Cifar-10-C, and Cifar-100-C (created by Hendrycks and Dietterich~\cite{hendrycks2019benchmarking}), as well as MNIST-C~\cite{mu2019mnist}, and FMNIST-C~\cite{Weiss2022SimpleTechniques}. 
These test sets, widely used in studies on DNN robustness, are generated by a diverse range of modifications inspired by real-world input transformations and corruptions.
In our experiments, we focused on a subset of seven corruption types across all sets, including Brightness, Saturate, Spatter, Contrast, Gaussian Noise, JPEG Compression, and Speckle Noise, referred to as C1, C2, ..., and C7. 

Each set also provides five levels of corruption severity. To answer the first two research questions, we focus on the third level of each corruption type, which represents a medium range of severity. To address RQ3, we extend our experiments to include both weaker and stronger corruption levels for each corruption type, specifically using inputs from corruption severity levels 2 and 4 of the corrupted input sets. We excluded the weakest and strongest corruption levels (severity=1 and 5, respectively) since the former includes inputs very similar to the original ones, resulting in the fine-tuned model ($M_T$) closely resembling the pre-trained model with nearly identical accuracies. The latter leads to inputs that are corrupted so severely that even humans struggle to predict them accurately, resulting in a fine-tuning process that is less effective, leading to low-accuracy fine-tuned models. 

Additionally, we observed that for certain corruption types applied to specific input sets, even intermediate severity levels (severity levels 2, 3, and 4) altered the images so severely that fine-tuning cannot effectively produce a high-accuracy DNN model. To maintain alignment with real-world fine-tuning scenarios, we further exclude fine-tuned models created by such corruptions from our experiments.



In addition to corrupted test sets, we require corrupted versions of our source training sets, i.e., target training set, to perform the fine-tuning process and obtain fine-tuned models. However, Cifar-10-C, Cifar-100-C, MNIST-C, and FMNIST-C datasets only provide corrupted test sets. Therefore, we utilized the original source code and settings employed in the creation of these corrupted test sets to generate the corrupted versions of our source training sets. 
However, in practical scenarios, the size of the target training set is often limited and significantly smaller than the source training set due to labeling costs. Therefore, we sampled a subset of these corrupted training sets to serve as the target training set, as detailed in the following section. 
We refer to each fine-tuned DNN model ($M_T$) and its corresponding training and test input sets ($Train_T$ and $Test_T$, respectively) as subjects in our experiments. Consequently, we conducted a comprehensive study across 68 subjects, encompassing four source input sets, seven corruption types, and three levels of corruption severity. \TR{C3.7}{To ensure full transparency and reproducibility, all fine-tuned models, along with their corresponding pre-trained counterparts, are publicly available in our replication package. }

\subsection{Fine-tuning process}
The goal of the fine-tuning process is to enhance the pre-trained model by training it for an additional number of epochs with new inputs from the target context~\cite{hu2022empirical}. 
In practice, the size of the target training set is often limited and significantly smaller than the source training set, due to the labeling cost~\cite{pan2009survey, xu2020transfer}. Indeed, this is the main motivation for adopting transfer learning techniques, such as fine-tuning and retraining a pre-trained model. Therefore, to be aligned with practice, we randomly sample only a portion of corrupted training inputs, constrained by a sampling size $n_s$, and use them for fine-tuning the pre-trained model to adapt it to the target context. 

To define a realistic value for $n_s$, two conditions must be satisfied: (1) the accuracy of the fine-tuned model trained with $n_s$ inputs should exceed that of a DNN model trained from scratch using the same number of inputs, and (2) the accuracy of the fine-tuned model with $n_s$ inputs should also surpass the performance of the pre-trained model $M_S$. This ensures the effectiveness of fine-tuning with a limited training set, resulting in a model that better aligns with the target distribution compared to the pre-trained model. An unseen validation set from the target distribution was used to verify these conditions. We thus determined the size of the target training set for fine-tuning each pre-trained model, as shown in Table~\ref{tab:SourceModels}, to satisfy the above two criteria.

\subsection{Evaluation Measure}
\label{sec:Measures}

To assess the effectiveness of MetaSel and compare its performance against baselines in selecting a small subset of test inputs, we leverage the Test Relative Coverage (TRC) metric. TRC is a well-established measure in the DNN testing literature~\cite{li2021testrank, hu2023evaluating} that quantifies the effectiveness of test selection methods in identifying mispredicted inputs within a defined selection budget, $b$. The TRC value for a selected test subset $T$ containing the top $b$ inputs, selected from an ordered list of inputs produced by prioritization approaches, is computed as follows:  

\begin{equation} \label{Eq:TRC}
\text{TRC (T)} = \frac{Mis_T}{min (b, Mis_{total})}
\end{equation}

\noindent where $Mis_{total}$ represents the total number of failures (i.e., mispredicted inputs) in the entire test set, $Mis_T$ is the number of failures detected by test subset $T$, and $b$ denotes the selection budget, i.e., size of $T$. 
Note that the denominator in the TRC formula, as defined in Equation~(\ref{Eq:TRC}), adjusts for budget sizes smaller than $Mis_{total}$. This ensures that TRC accurately evaluates how effective test selection is compared to the ideal performance within the given budget.
If the denominator were fixed at $Mis_{total}$, regardless of the budget, achieving a TRC value of 1 would be impossible for any subset selected with $b < Mis_{total}$. For example, with $Mis_{total}$=20 and a selection budget of $b$=10, no selected subset could reach a TRC value of 1 due to the inherent budget constraint. The formulation in Equation~(\ref{Eq:TRC}) overcomes this limitation by adapting to the budget, thereby providing a meaningful and general measure of effectiveness across all budgets and regardless of the number of mispredicted inputs.

Another common evaluation metric used in DNN testing literature is the Average Percentage of Fault-Detection (APFD), a standard metric in traditional software test prioritization~\cite{rothermel2001prioritizing, liang2018redefining} that has also been adapted recently for evaluating DNN test prioritization approaches~\cite{Weiss2022SimpleTechniques, feng2020deepgini}. 
However, we decided not to evaluate MetaSel based on APFD for three reasons. 
First, APFD is specifically designed to measure the effectiveness of prioritization techniques over the entire test set, whereas our application context, as clearly described in Section~\ref{sec:Objectives}, is focused on the selection of test inputs within small budgets.
In the domain of fine-tuned models, while practitioners often have access to an abundance of unlabeled inputs that can be easily collected from the target context, the primary objective in such scenarios is not to prioritize the entire test set but to identify a small subset of inputs with the highest likelihood of being mispredicted. 
Therefore, we need to focus on achieving higher TRC within considerably smaller budgets, rather than maximizing prioritization effectiveness across the entire test set.

\begin{figure*}[ht!]
  \centering

  \begin{minipage}[b]{0.20\linewidth}
    \centering
    \includegraphics[width=\linewidth]{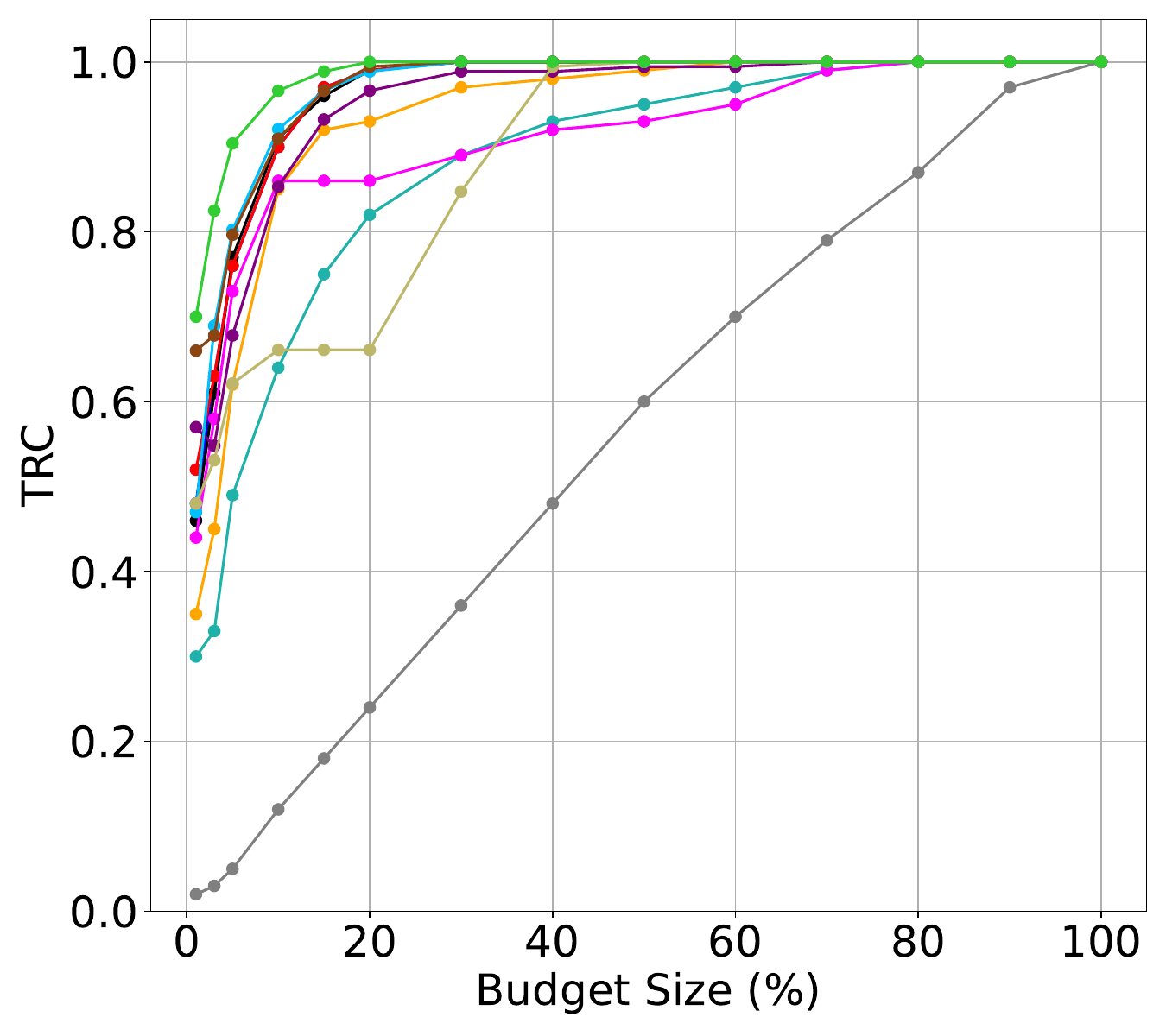}
    \par\vspace{-3pt}
      MNIST Brightness
  \end{minipage}
  \hfill
  \begin{minipage}[b]{0.20\linewidth}
    \centering
    \includegraphics[width=\linewidth]{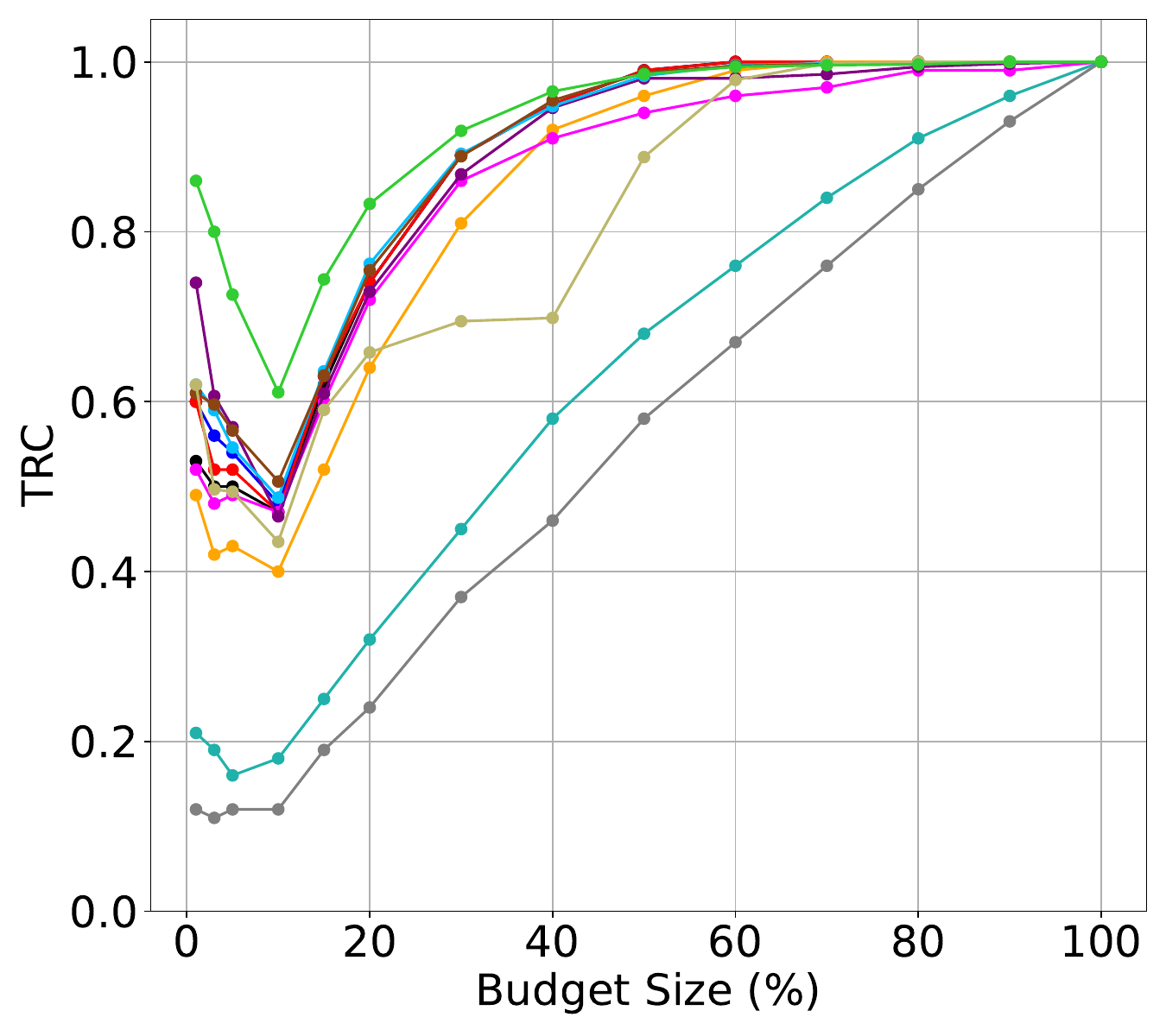}
    \par\vspace{-3pt}
      FMNIST Brightness
  \end{minipage}
  \hfill
  \begin{minipage}[b]{0.20\linewidth}
    \centering
    \includegraphics[width=\linewidth]{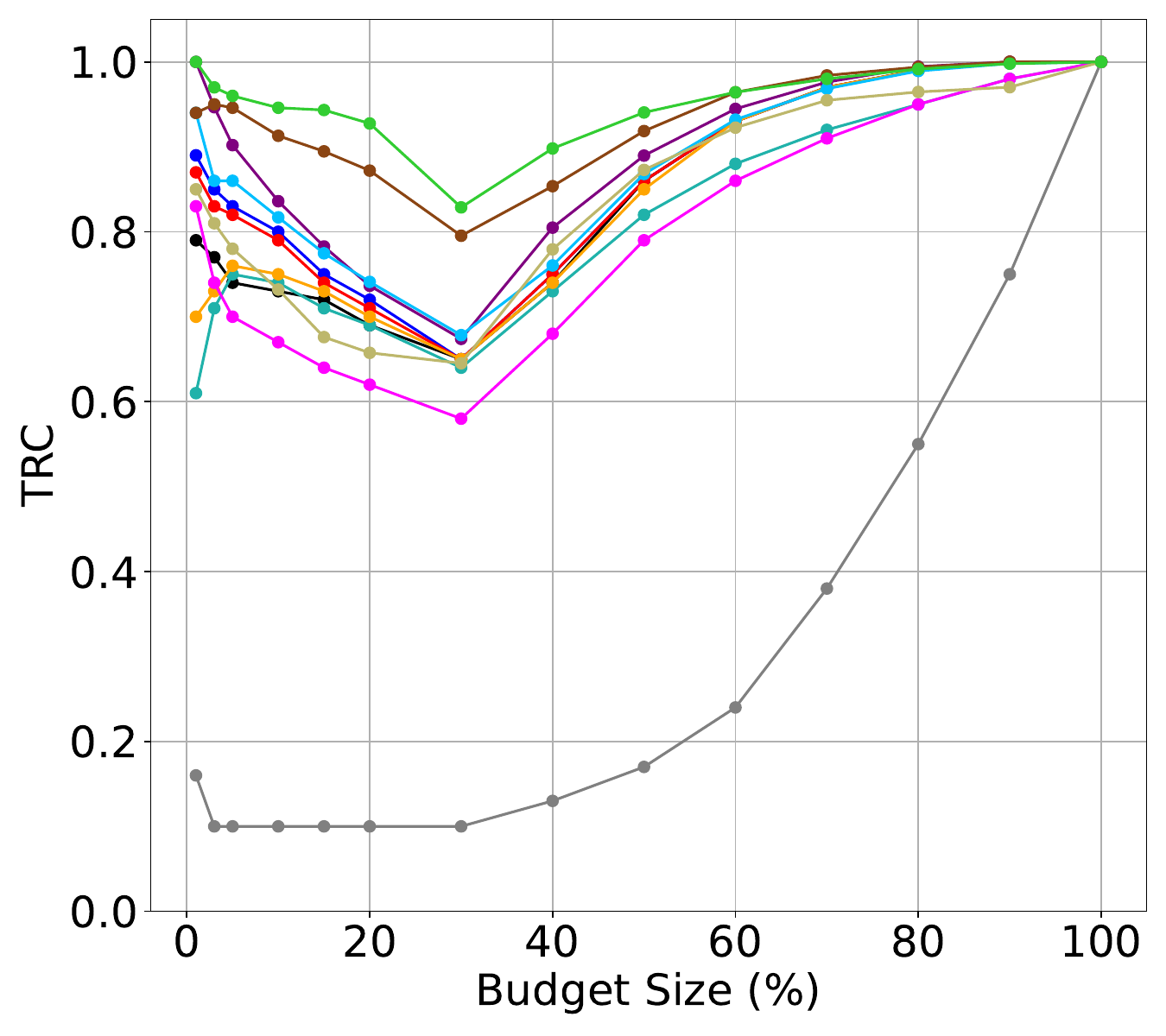}
    \par\vspace{-3pt}
      Cifar-100 Brightness
  \end{minipage}
  \hfill
  \begin{minipage}[b]{0.20\linewidth}
    \centering
    \includegraphics[width=\linewidth]{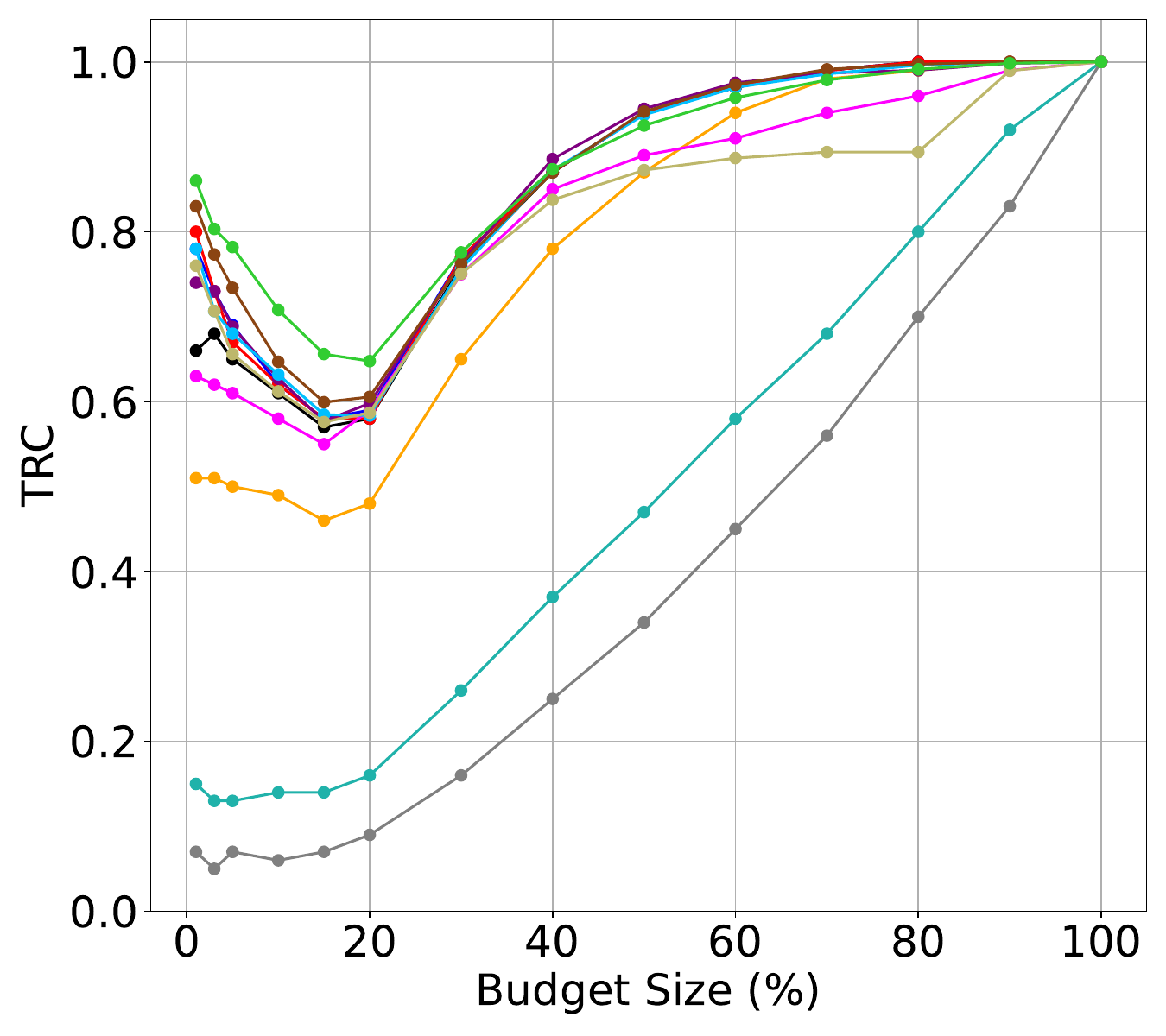}
    \par\vspace{-3pt}
      Cifar-10 Brightness
  \end{minipage}

  \vspace{0.7em} 

  \begin{minipage}[b]{0.20\linewidth}
    \centering
    \includegraphics[width=\linewidth]{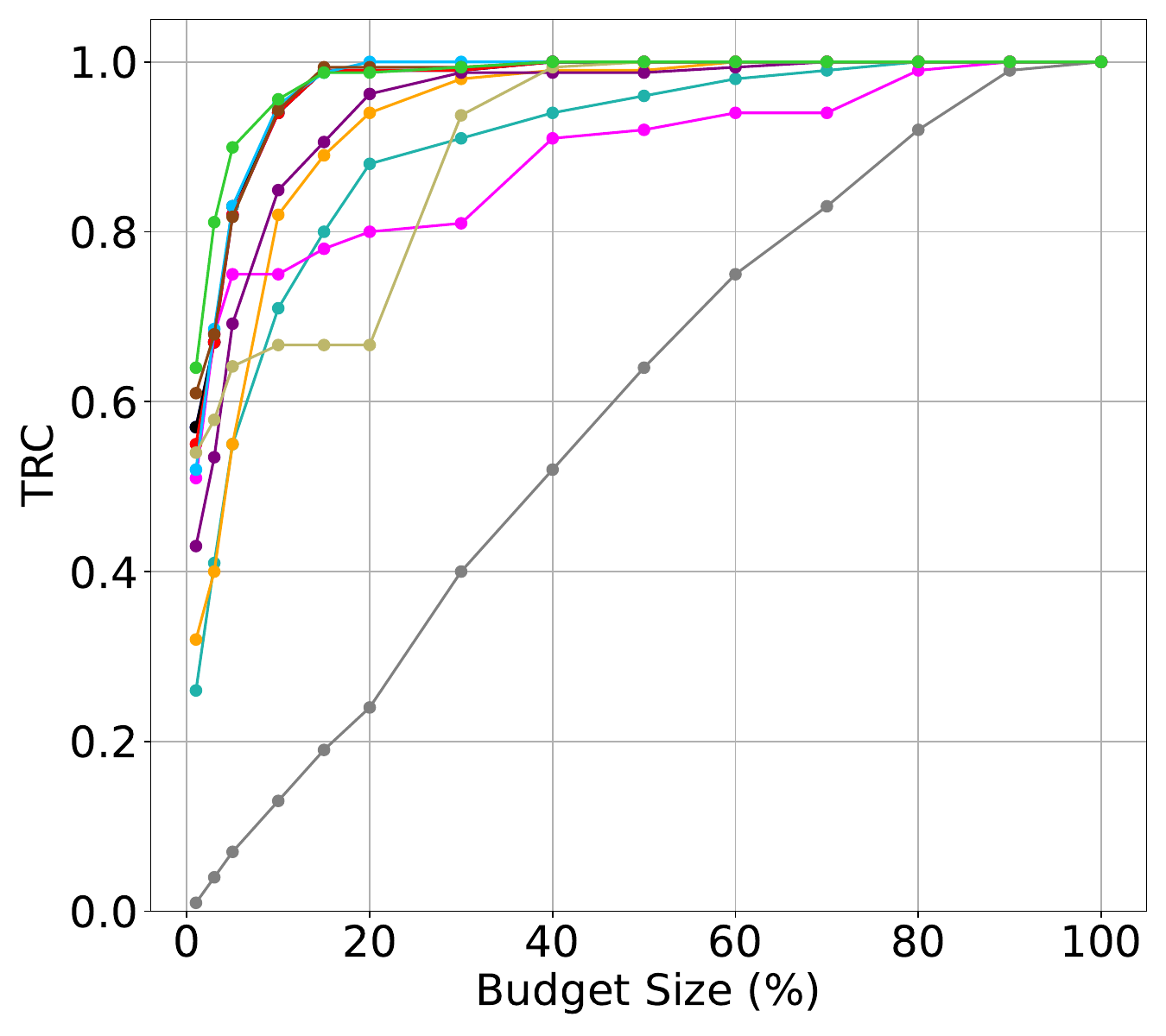}
    \par\vspace{-3pt}
      MNIST Saturate
  \end{minipage}
  \hfill
  \begin{minipage}[b]{0.20\linewidth}
    \centering
    \includegraphics[width=\linewidth]{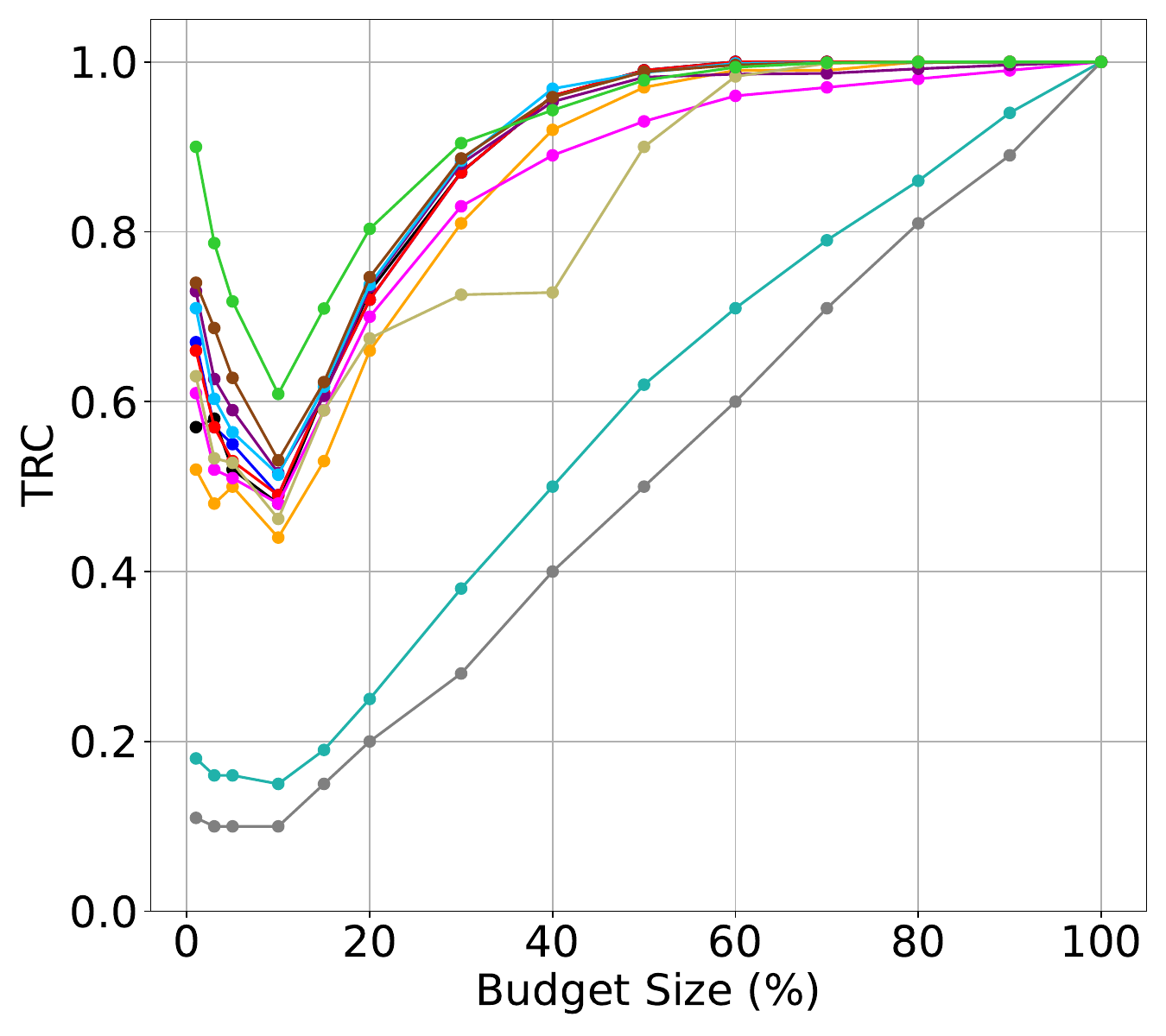}
    \par\vspace{-3pt}
      FMNIST Saturate
  \end{minipage}
  \hfill
  \begin{minipage}[b]{0.20\linewidth}
    \centering
    \includegraphics[width=\linewidth]{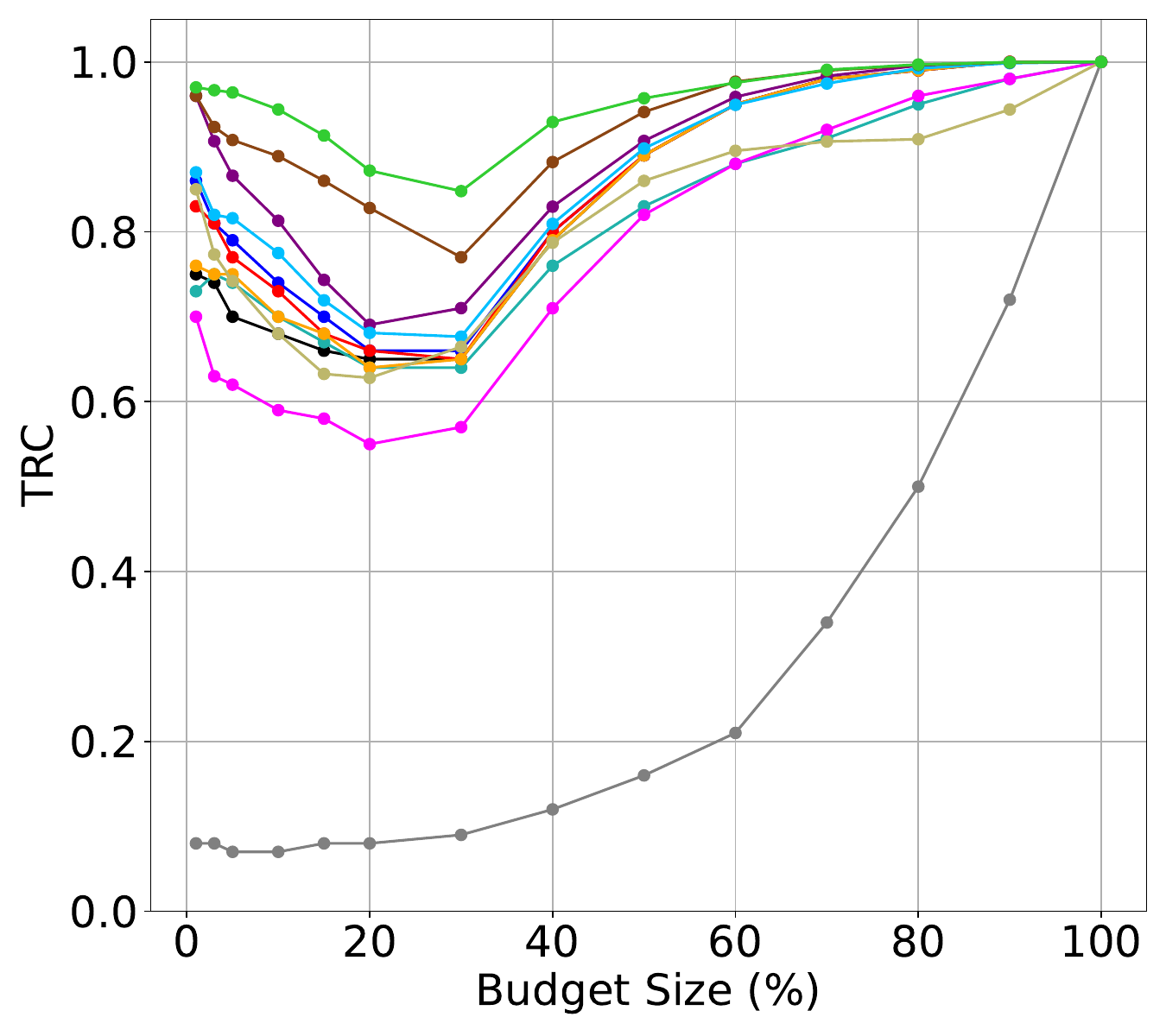}
    \par\vspace{-3pt}
      Cifar-100 Saturate
  \end{minipage}
  \hfill
  \begin{minipage}[b]{0.20\linewidth}
    \centering
    \includegraphics[width=\linewidth]{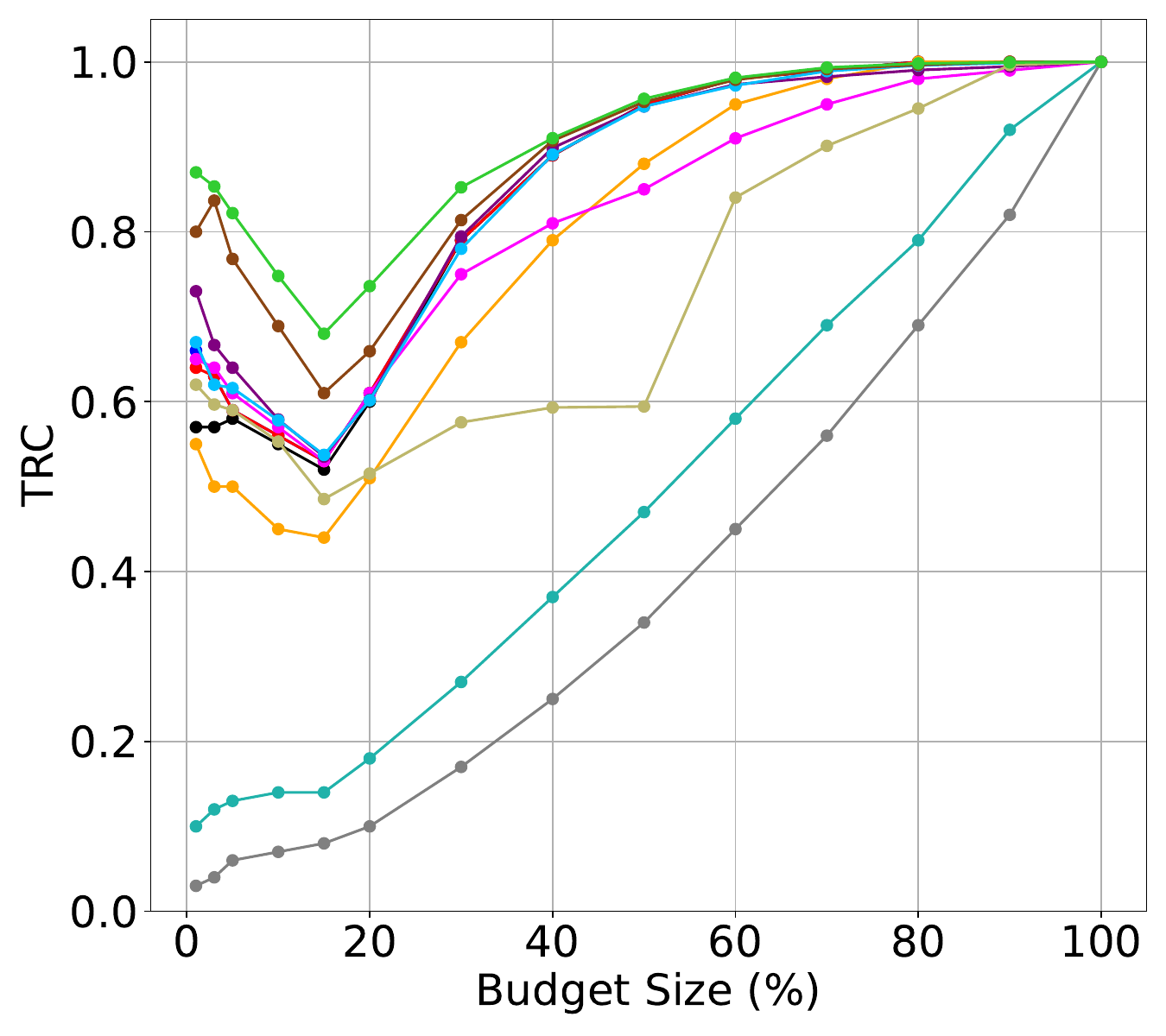}
    \par\vspace{-3pt}
      Cifar-10 Saturate
  \end{minipage}

  \vspace{0.7em} 

  \begin{minipage}[b]{0.20\linewidth}
    \centering
    \includegraphics[width=\linewidth]{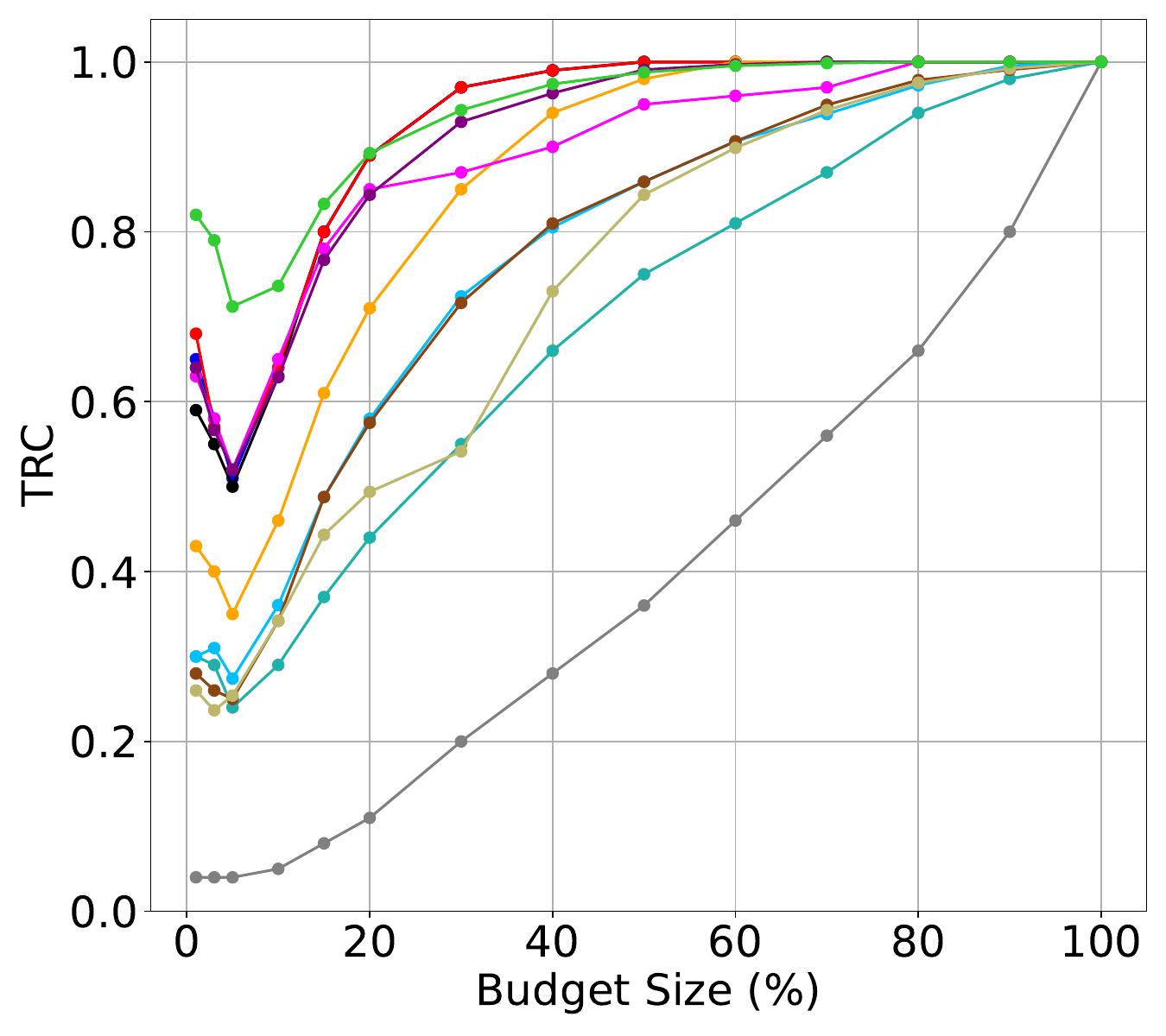}
    \par\vspace{-3pt}
      MNIST Spatter
  \end{minipage}
  \hfill
  \begin{minipage}[b]{0.20\linewidth}
    \centering
    \includegraphics[width=\linewidth]{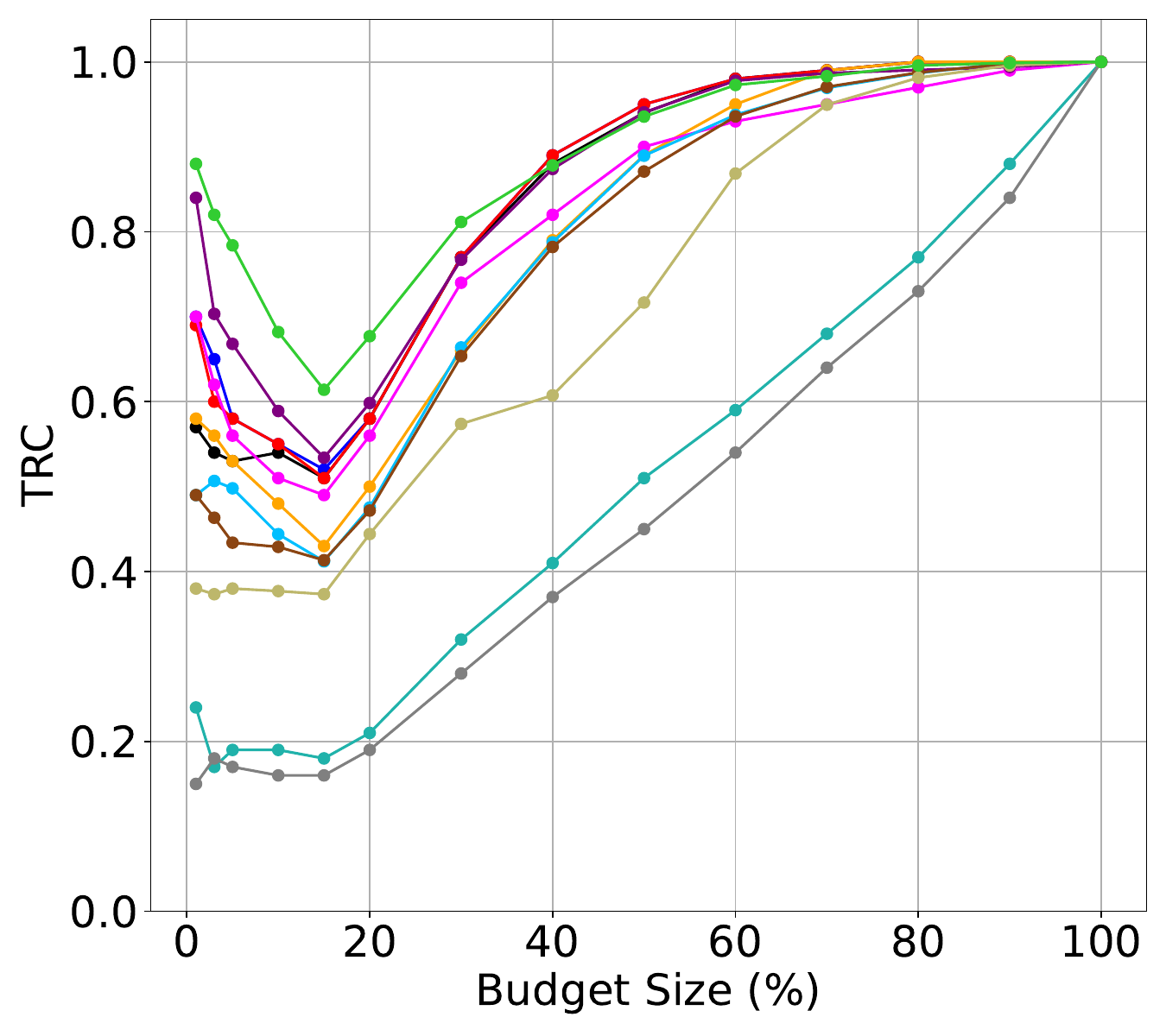}
    \par\vspace{-3pt}
      FMNIST Spatter
  \end{minipage}
  \hfill
  \begin{minipage}[b]{0.20\linewidth}
    \centering
    \includegraphics[width=\linewidth]{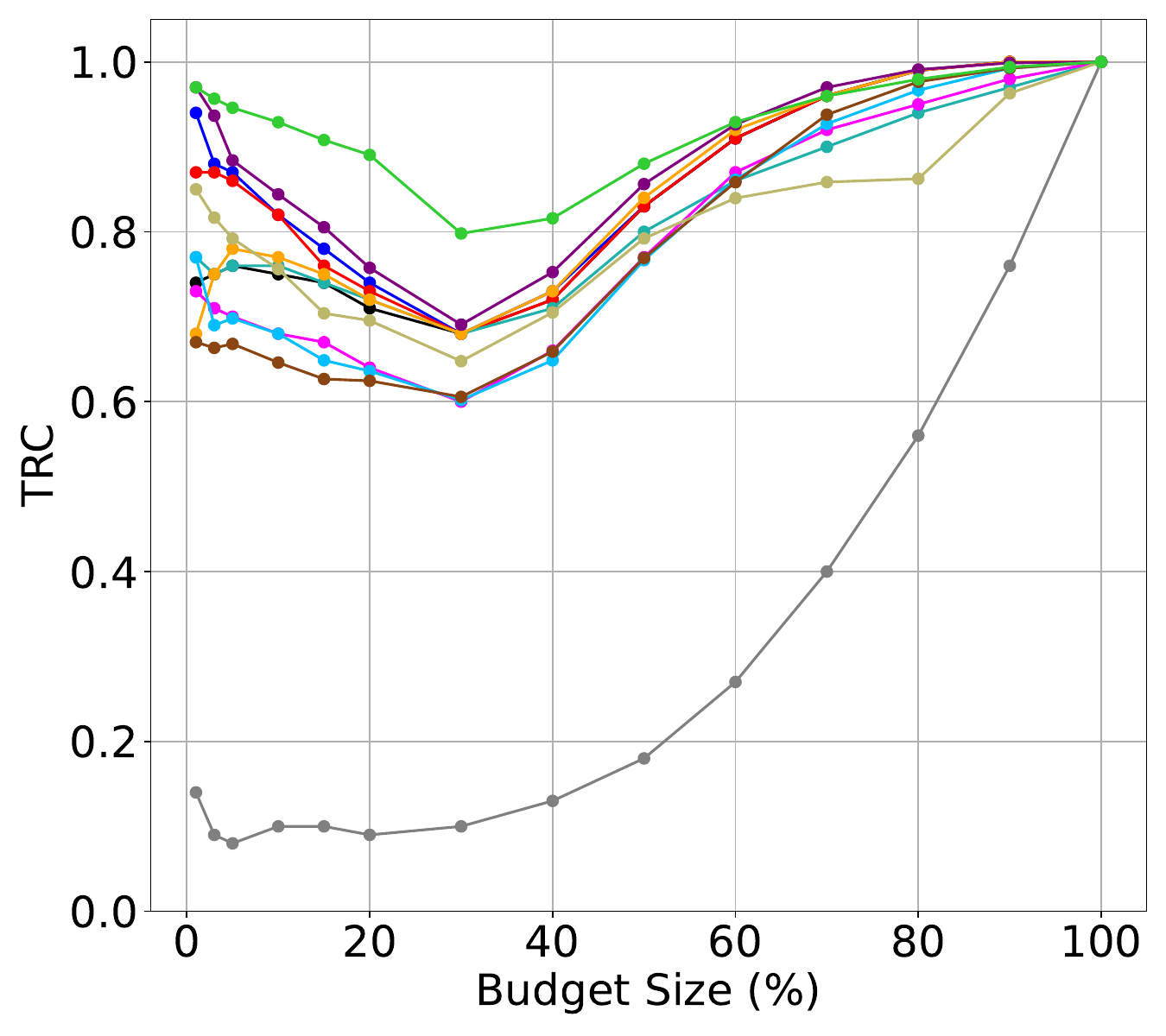}
    \par\vspace{-3pt}
      Cifar-100 Spatter
  \end{minipage}
  \hfill
  \begin{minipage}[b]{0.20\linewidth}
    \centering
    \includegraphics[width=\linewidth]{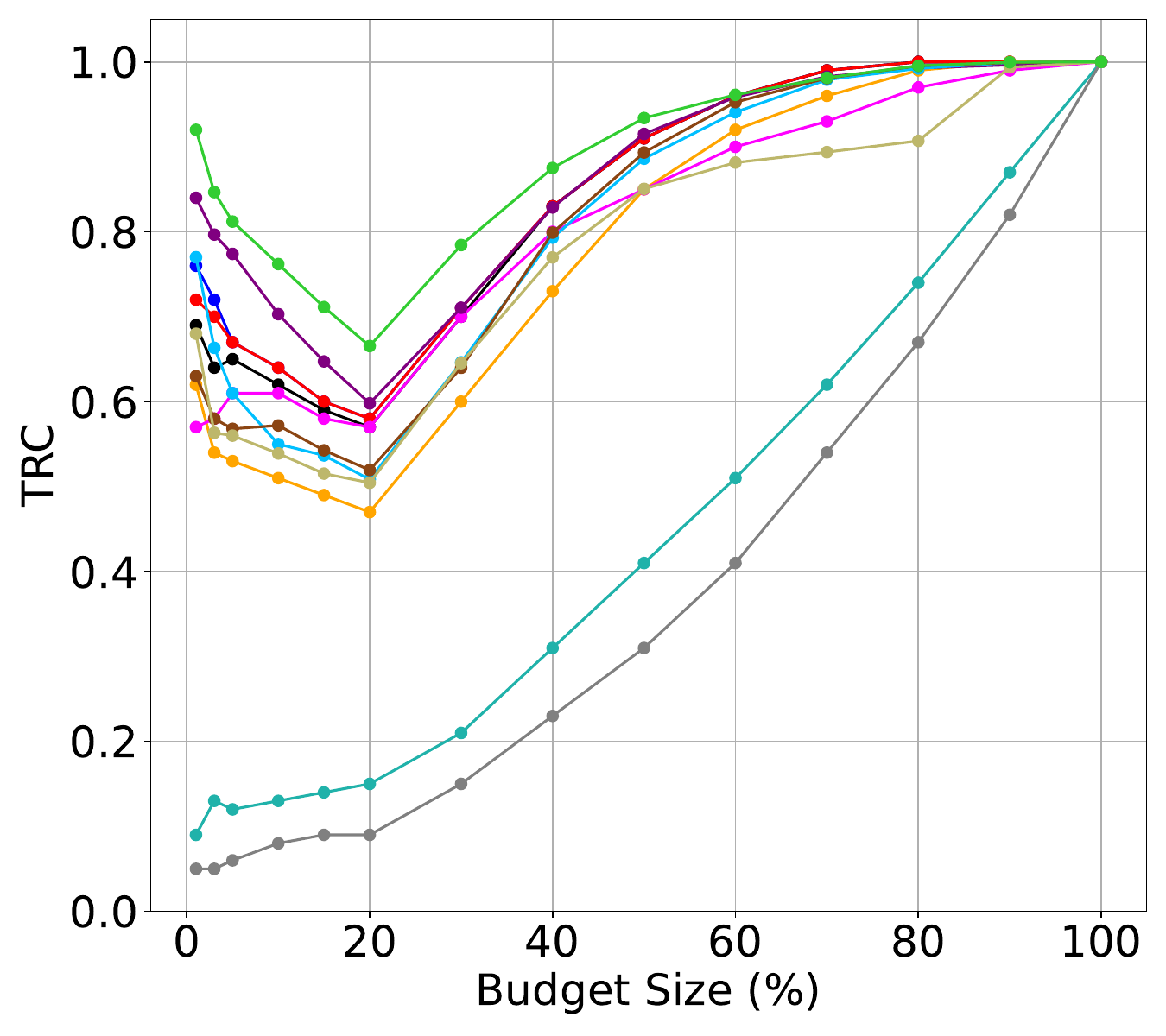}
    \par\vspace{-3pt}
      Cifar-10 Spatter
  \end{minipage}

\vspace{0.7em}

  \begin{minipage}[b]{0.20\linewidth}
    \centering
    \includegraphics[width=\linewidth]{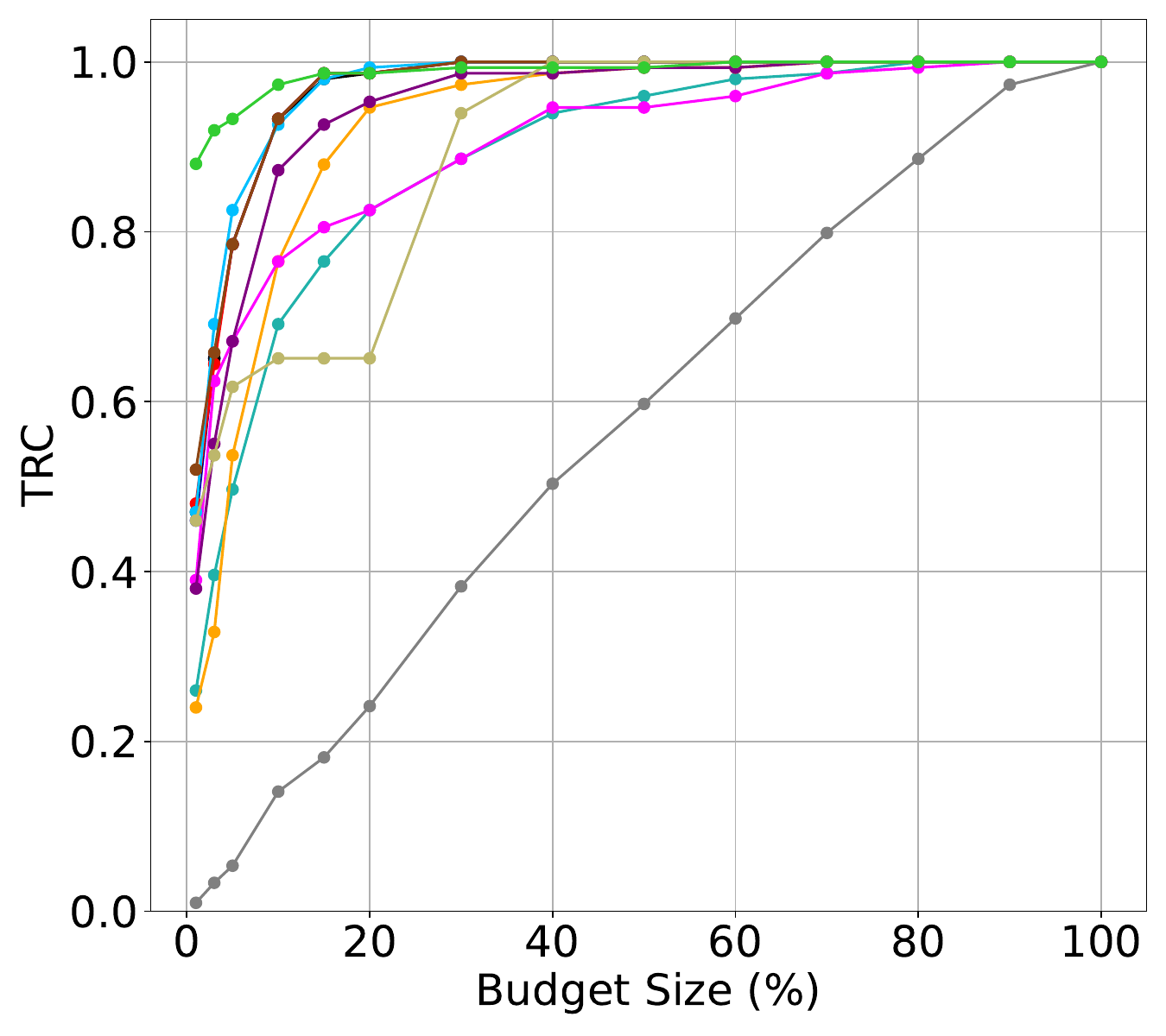}
    \par\vspace{-3pt}
      MNIST jpeg Comp
  \end{minipage}
  \hfill
  \begin{minipage}[b]{0.20\linewidth}
    \centering
    \includegraphics[width=\linewidth]{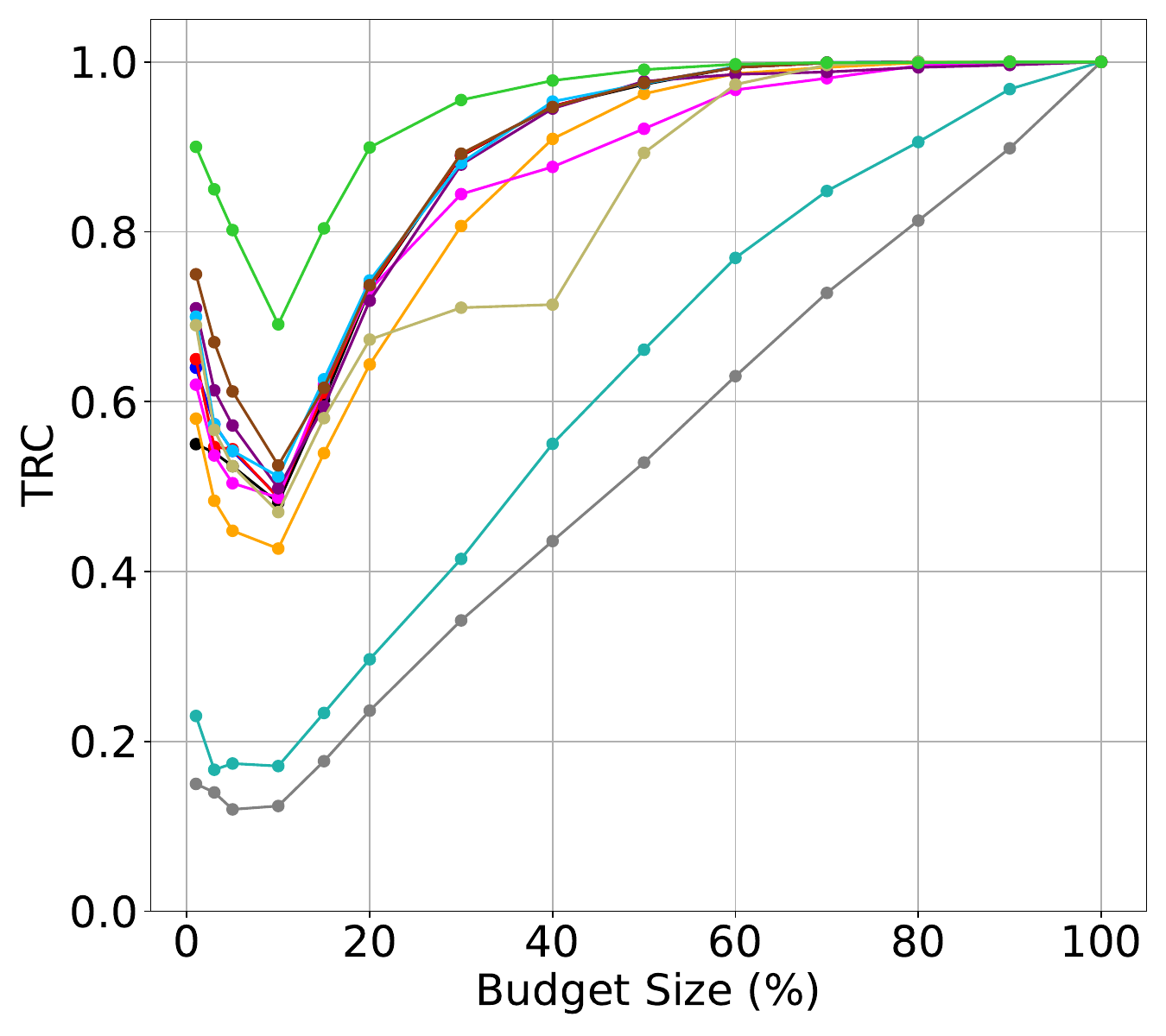}
    \par\vspace{-3pt}
      FMNIST jpeg Comp
  \end{minipage}
  \hfill
  \begin{minipage}[b]{0.20\linewidth}
    \centering
    \includegraphics[width=\linewidth]{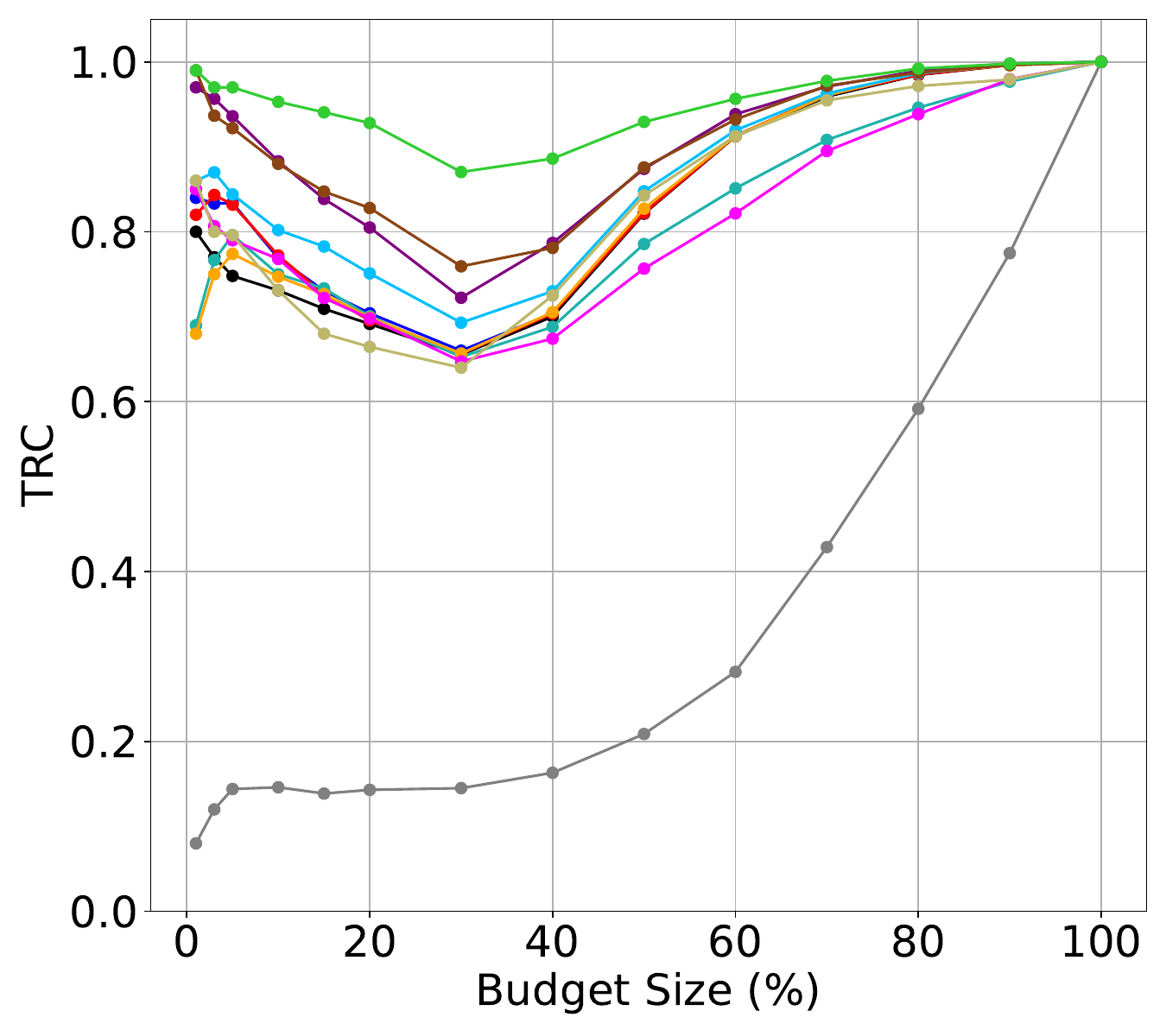}
    \par\vspace{-3pt}
      CIFAR-100 Contrast
  \end{minipage}
  \hfill
  \begin{minipage}[b]{0.20\linewidth}
    \centering
    \includegraphics[width=\linewidth]{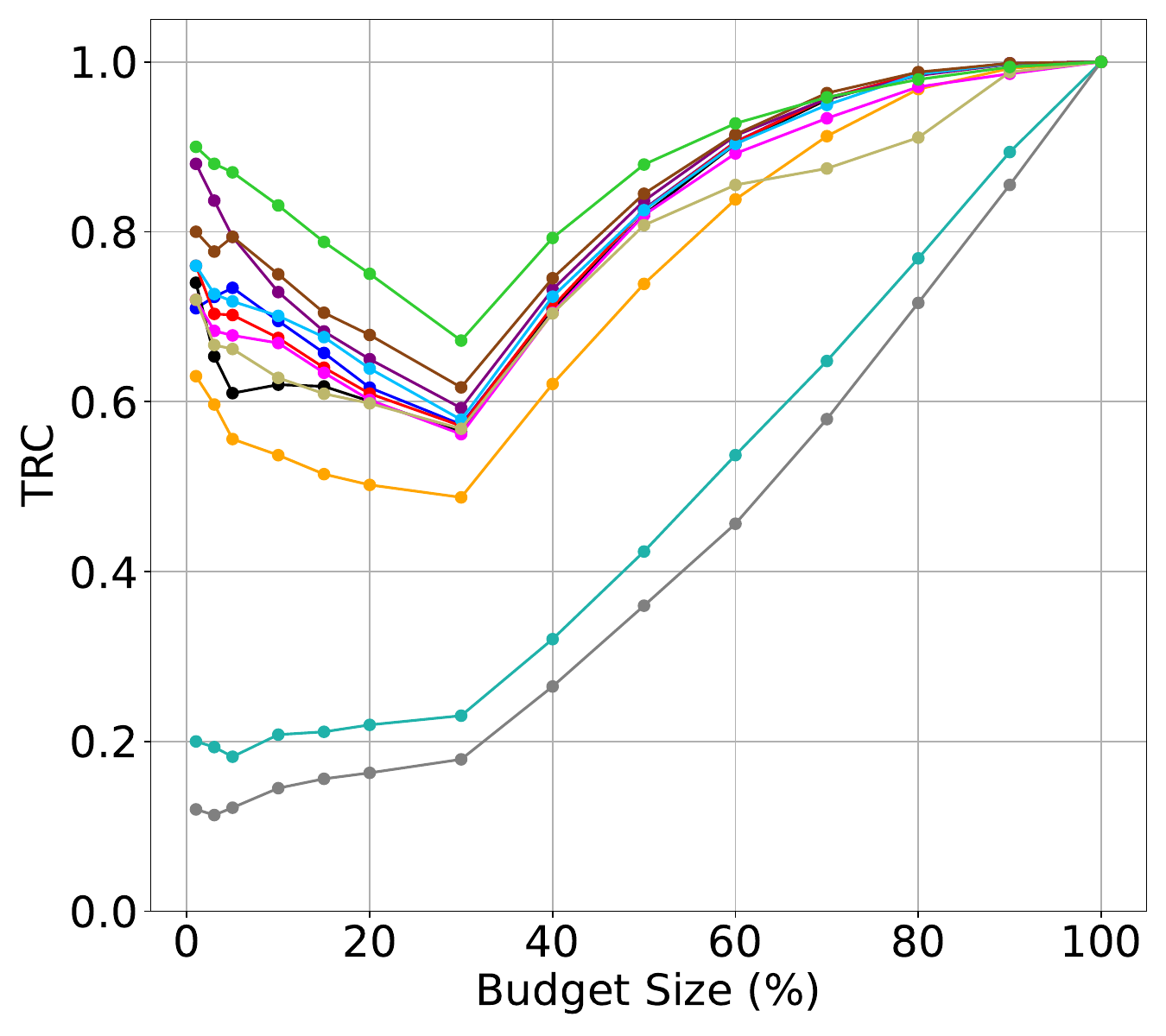}
      CIFAR-10 jpeg Comp
  \end{minipage}

\vspace{0.7em}
  \begin{minipage}[b]{0.20\linewidth}
    \centering
    \includegraphics[width=\linewidth]{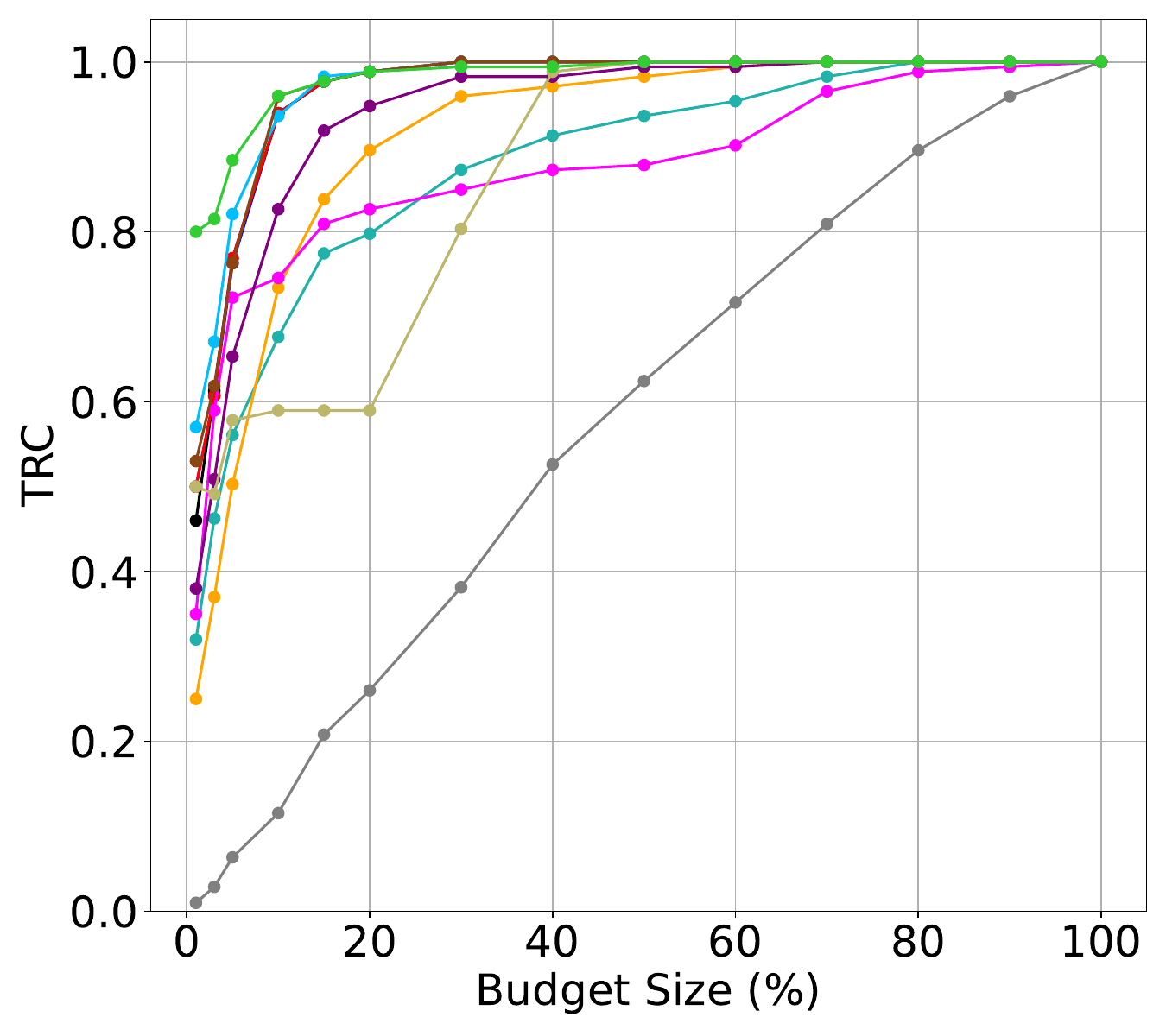}
    \par\vspace{-3pt}
      MNIST Speckle Noise
  \end{minipage}
  \hfill
  \begin{minipage}[b]{0.20\linewidth}
    \centering
    \includegraphics[width=\linewidth]{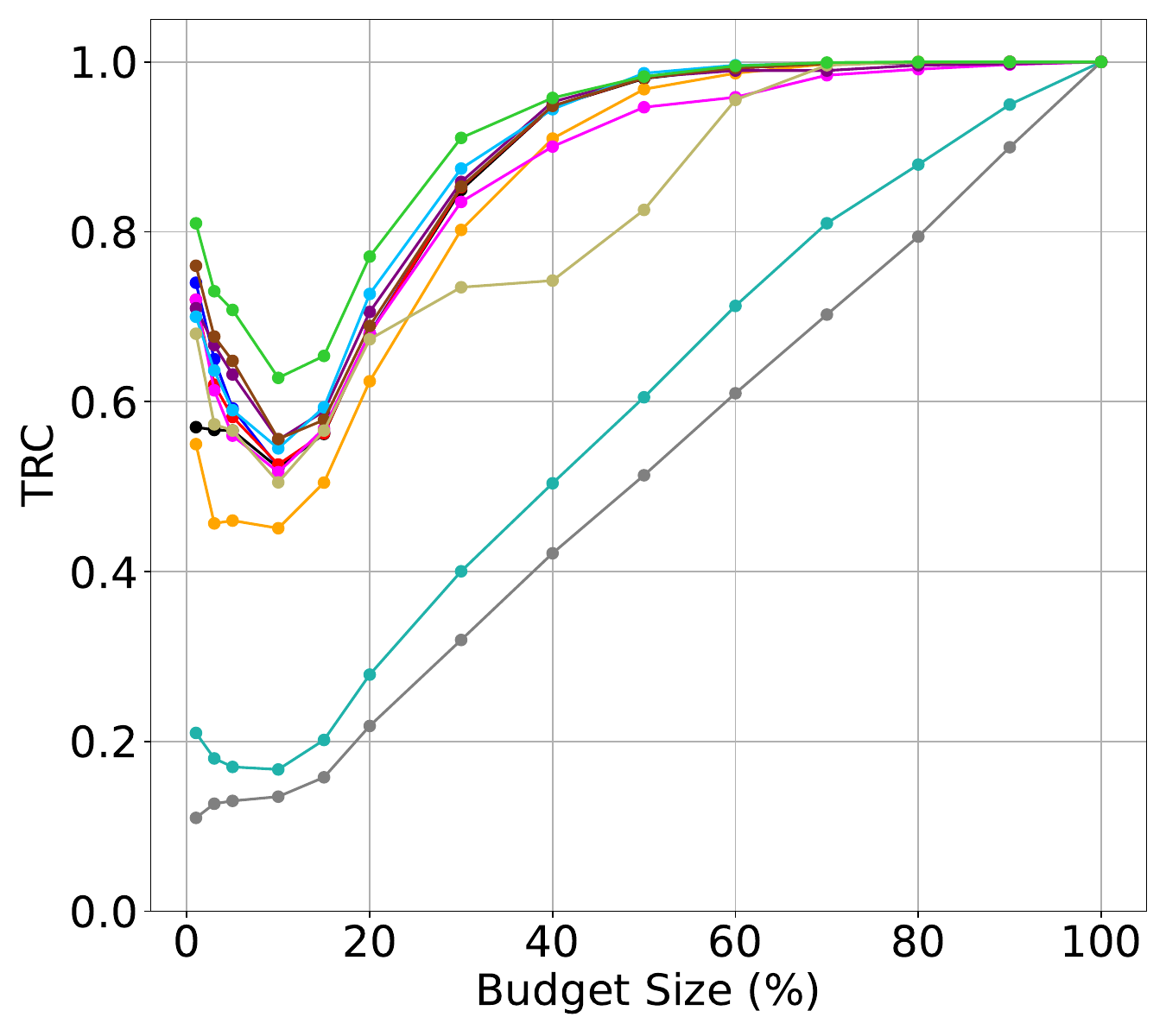}
    \par\vspace{-3pt}
      FMNIST Speckle Noise
  \end{minipage}
  \hfill
\begin{minipage}[b]{0.20\linewidth}
    \centering
    \includegraphics[width=\linewidth]{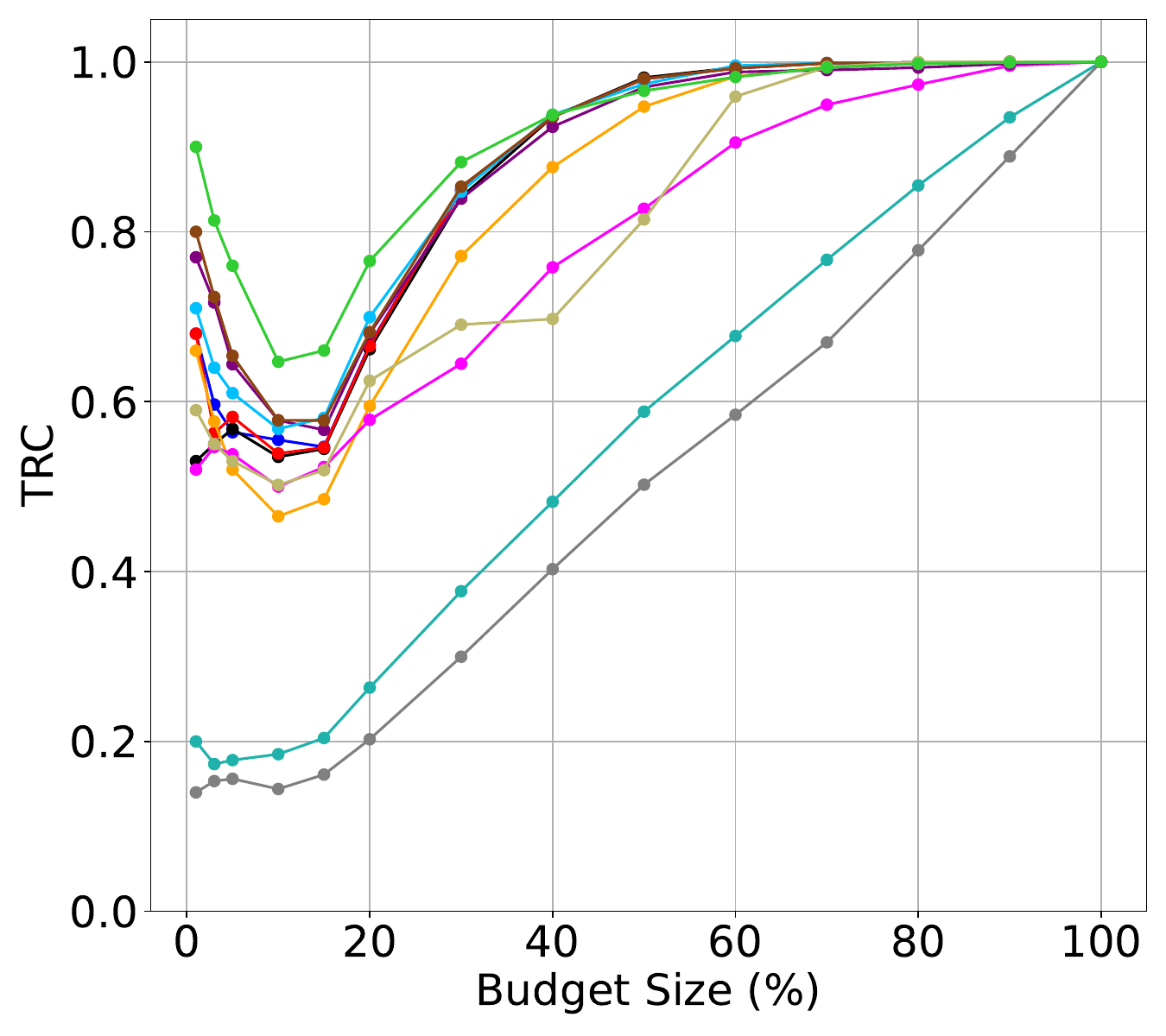}
    \par\vspace{-3pt}
      FMNIST Gaussian Noise
  \end{minipage}
  \hfill
  \begin{minipage}[b]{0.20\linewidth}
    \centering
    \includegraphics[width=\linewidth]{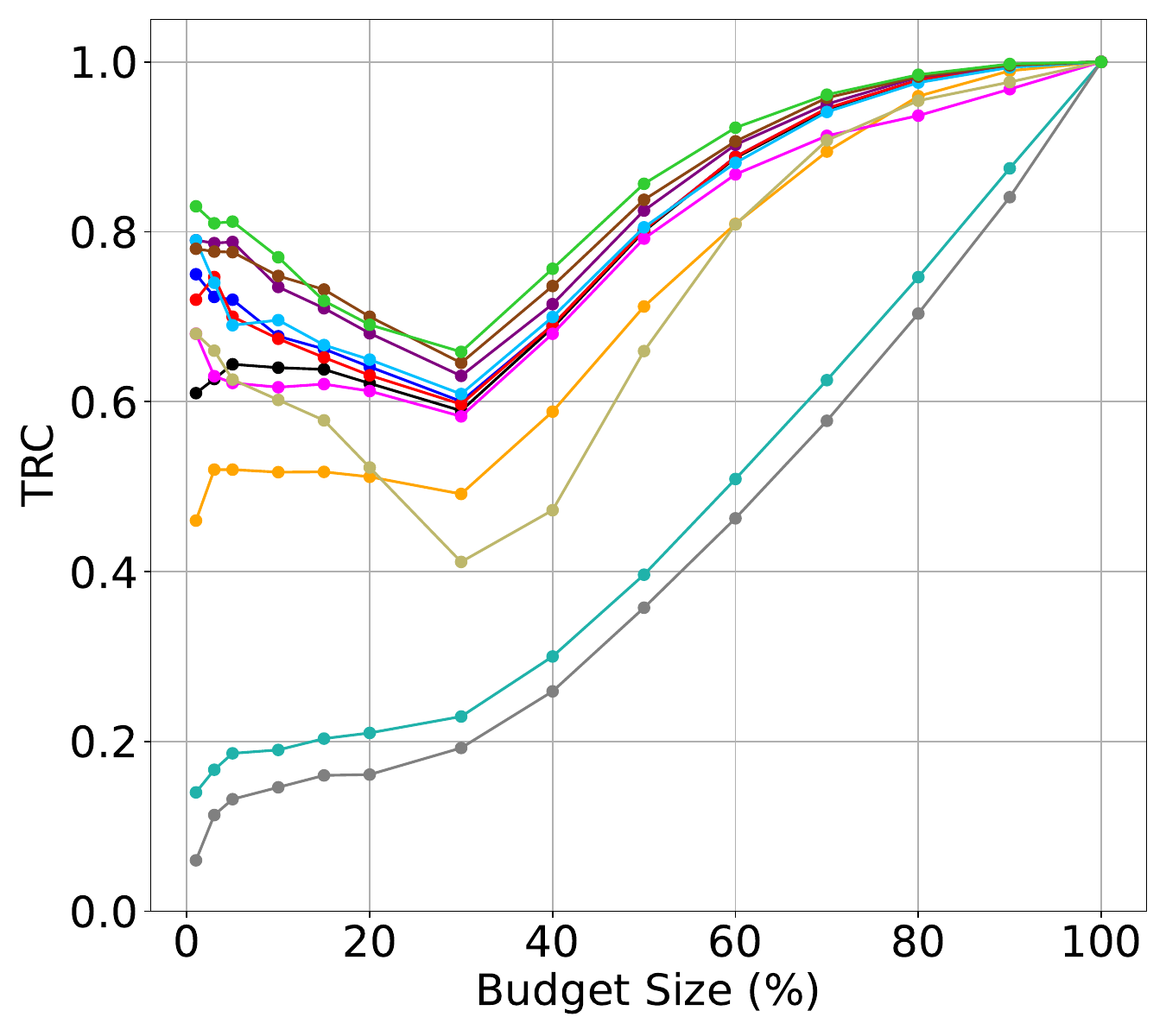}
    \par\vspace{-3pt}
      CIFAR-10 Speckle Noise
  \end{minipage}

\vspace{0.7em}

  \begin{minipage}[b]{0.20\linewidth}
    \centering
    \includegraphics[width=\linewidth]{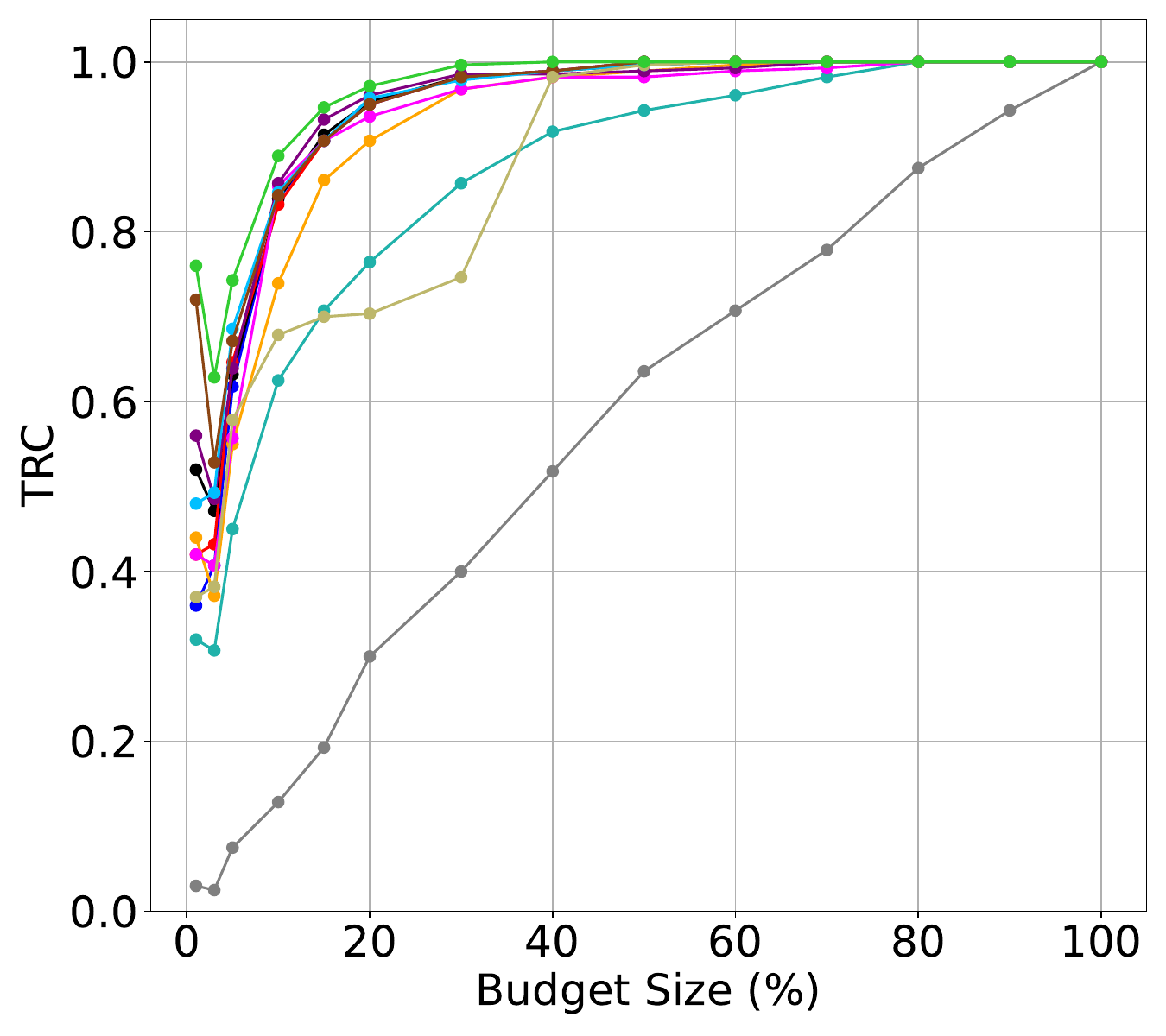}
    \par\vspace{3pt}
      MNIST Contrast
  \end{minipage}
    \hfill
  \begin{minipage}[b]{0.20\linewidth}
    \centering
    \includegraphics[width=\linewidth]{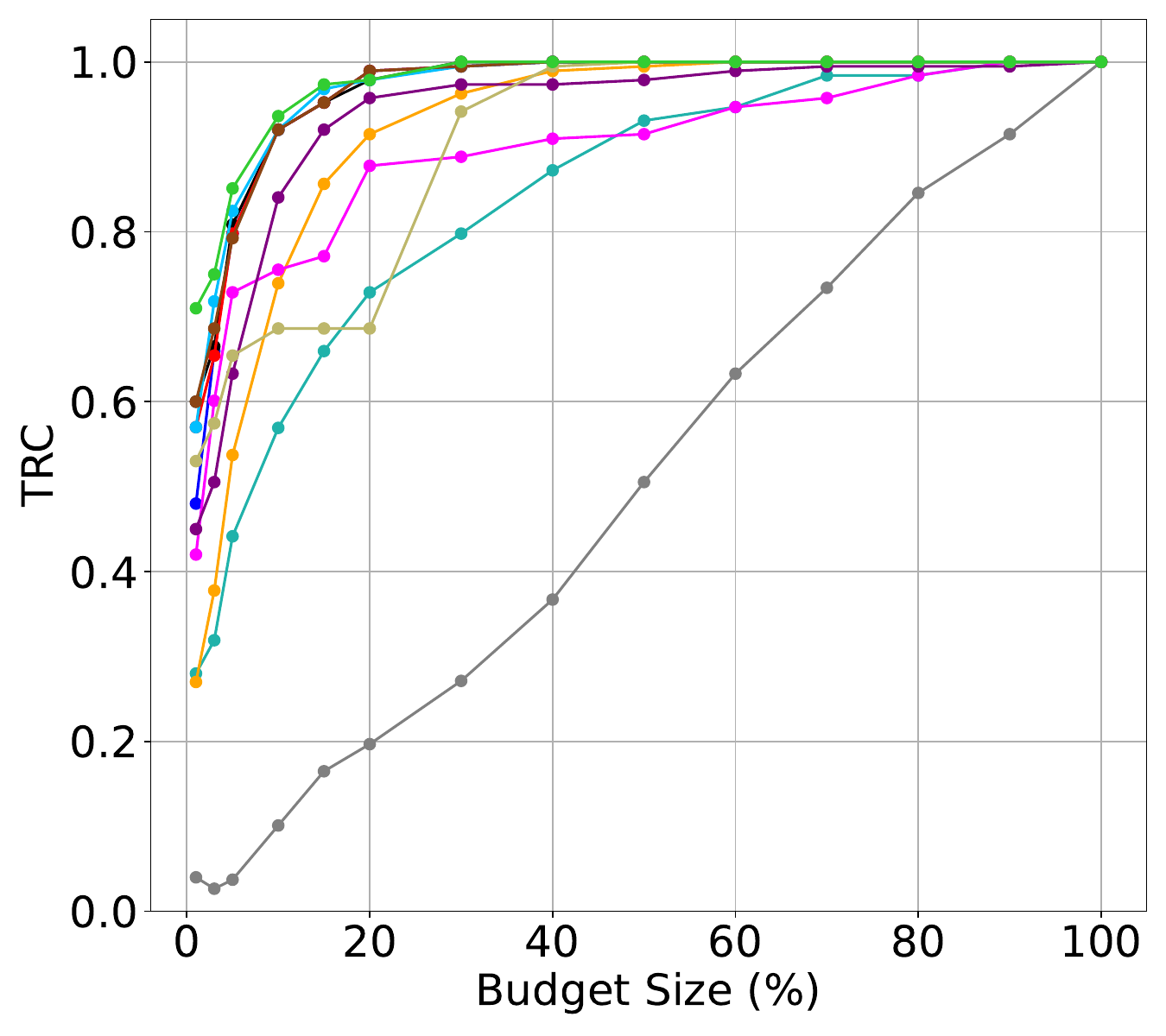}
    \par\vspace{-3pt}
      MNIST Gaussian Noise
  \end{minipage}
  \hfill
    \begin{minipage}[b]{0.20\linewidth}
    \centering
    \includegraphics[width=\linewidth]{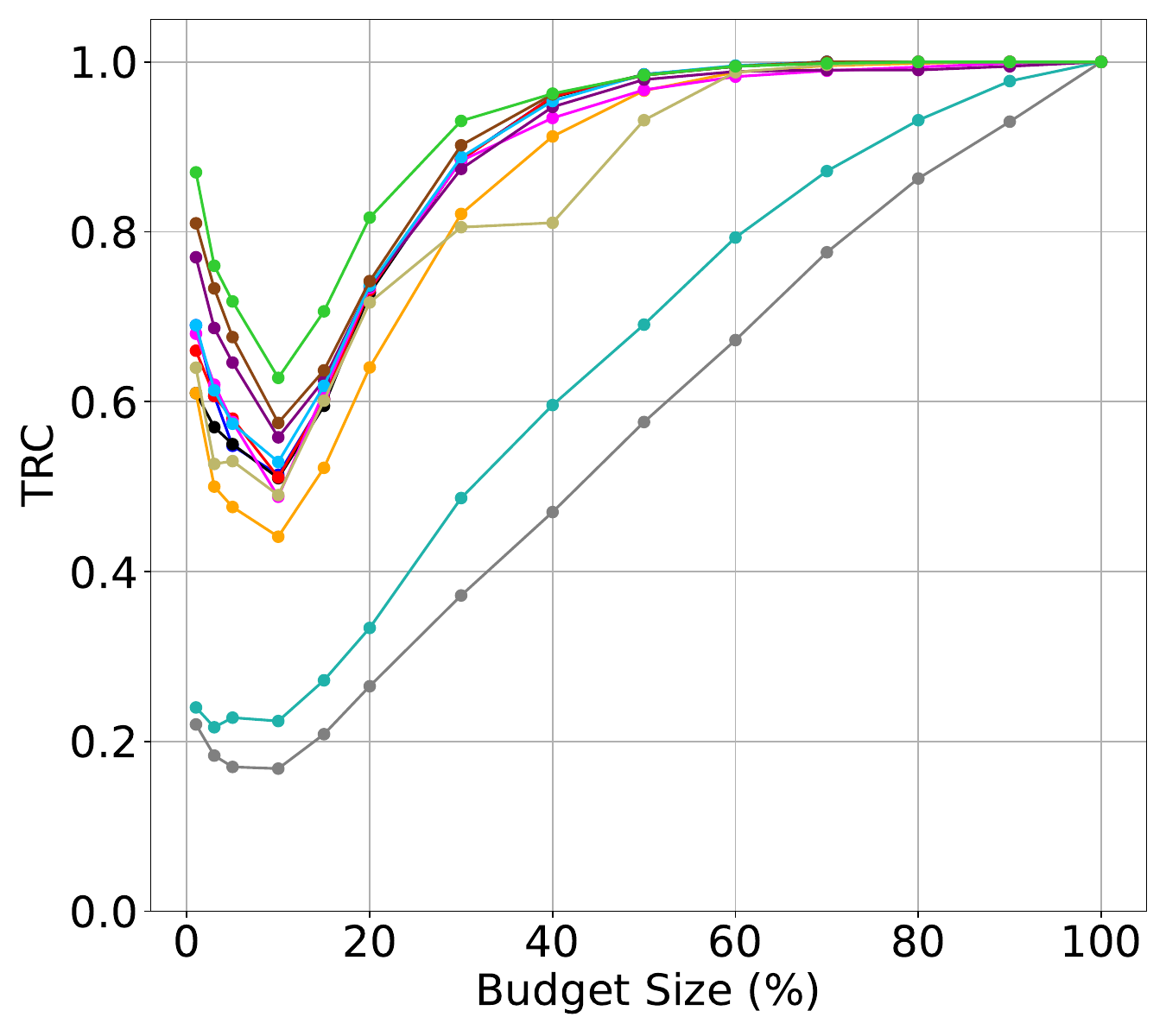}
    \par\vspace{-3pt}
      FMNIST Contrast
  \end{minipage}
   \hfill
    \begin{minipage}[b]{0.20\linewidth}
    \centering
    \includegraphics[width=1.1\linewidth]{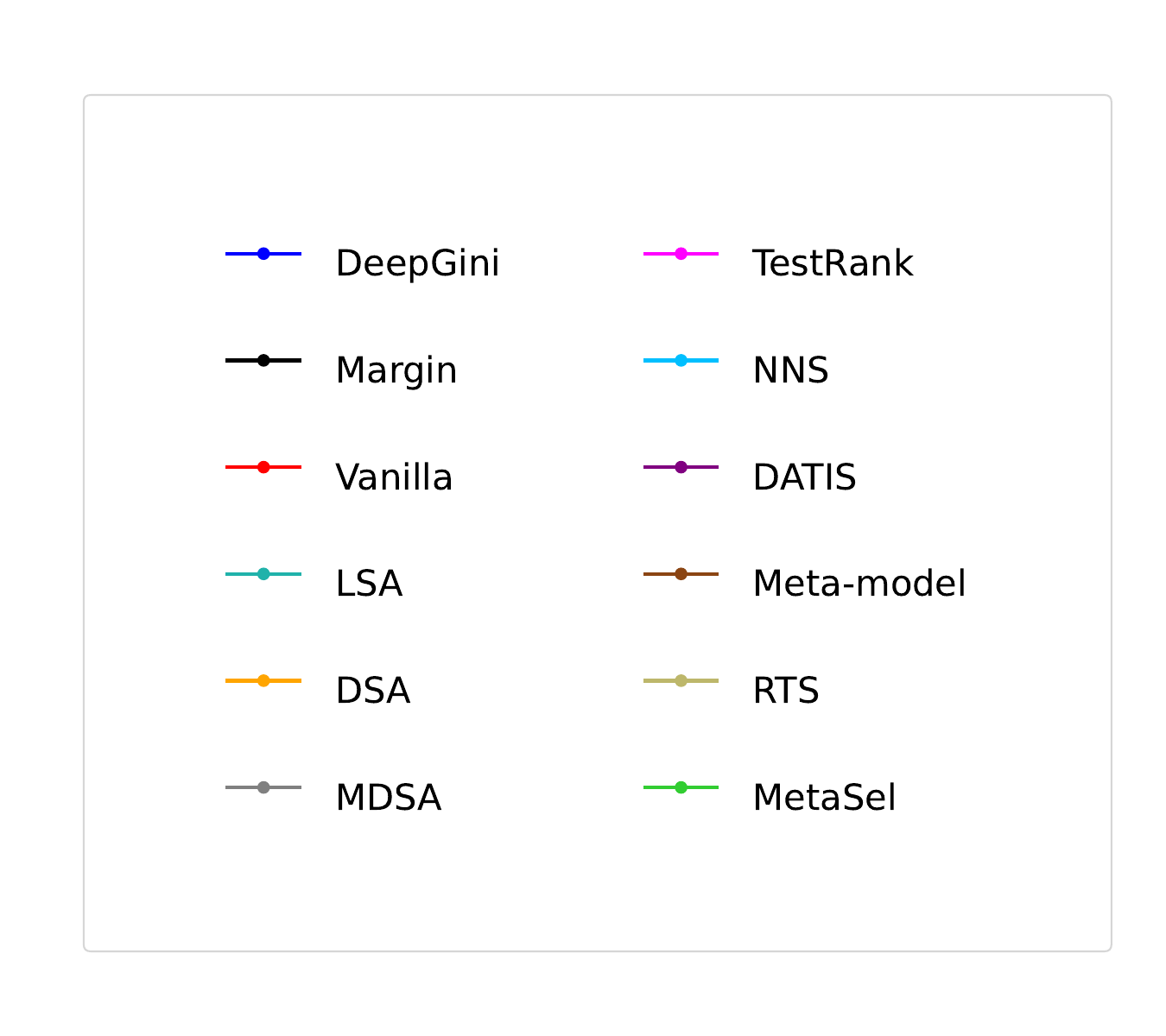}
    \par\vspace{-3pt}
     
  \end{minipage}

    \caption{TRC values for MetaSel and \Rev{11 baseline approaches} across diverse  selection budgets on all subjects with a medium severity level of distribution shift (severity level = 3)}
  \label{fig:ResultsRQ1}
\end{figure*}


Second, APFD cannot be employed to compare the effectiveness of prioritization approaches in selecting a subset of top-priority inputs with a given selection budget.
To clarify this limitation, it is important to recall that APFD values are computed as follows: 

\begin{equation} \label{Eq:APFD}
\text{APFD (T)} = 1 - \frac{\sum_{i=1}^{Mis_T} o_i}{b * Mis_T} + \frac{1}{2b}
\end{equation}

\noindent where $b$ is the total number of test inputs (i.e., selection budget), $Mis_T$ is the number of mispredicted inputs detected by test set $T$, and $o_i$ is the position of the mispredicted test inputs in the ordered list of inputs.
When calculating APFD for different approaches over the entire test set, the term $Mis_T$ remains constant across all methods. However, when focusing on subsets of top-priority inputs selected from the ordered lists generated by various prioritization approaches, the number of mispredicted inputs in each subset can vary. Consequently, the APFD value for one subset might surpass that of another, not because it more effectively prioritizes mispredicted inputs (i.e., mispredicted inputs in the ordered list are located in lower index numbers ($o_i$)), but rather because the subset contains fewer mispredicted inputs overall. This inconsistency renders APFD unsuitable as a metric for comparing the effectiveness of different prioritization approaches in selecting subsets of identical size.

Finally, our primary focus in this context is on evaluating an approach's capability to identify the maximum number of misclassifications within a specified selection budget. The TRC metric effectively captures this objective. Therefore, we employ TRC \TR{C2.5}{as the evaluation metric} in our experiments to assess and compare the performance of MetaSel against the baseline methods. \TR{C2.5}{We nevertheless report APFD results in our replication package~\cite{replicationpackage} for completeness. }

\subsection{Execution Environment}
We conducted our experiments using a cloud computing environment provided by the Digital Research Alliance of Canada~\cite{computecanada}, utilizing the Cedar cluster with 1352 various Nvidia GPUs (P100 Pascal and V100 Volta) with memory ranging from 12 to 32 GB. Conducting our extensive experiment, which involved deploying MetaSel and all baselines across all subjects and distribution shift severity levels, required approximately \Rev{48 days to complete (not accounting for interruptions).}

\section{Results}
\label{sec:Results}

In this section, we present the results related to our research questions and discuss their practical implications.

\subsection{\textbf{RQ1: MetaSel's Performance improvement}}
\label{sec:ResultsRQ1}

To address this research question, we applied MetaSel alongside all baseline methods described in Section~\ref{sec:Background} to prioritize unlabeled inputs in the target test set ($Test_T$). This results in an ordered list of inputs ($Test^{ordered}_T$) for each approach. Subsequently, we calculated the TRC values for test subsets containing the top $b$ inputs. For our analysis, $b$ was set to 1\%, 3\%, 5\%, 10\%, 15\%, 20\%, 30\%, ..., 90\%, and 100\% of $Test_T$. Though we present TRC results for all these budget sizes, we specifically focus on highly constrained selection budgets: 1\%, 3\%, 5\%, and 10\% of the target test set size in this research question to align with the practical application of test selection in the context of fine-tuned models.

We should recall that both NNS and Meta-model proposed by Demir \textit{et al.}~\cite{demir2024test} rely on the model's uncertainty for each input, allowing them to be deployed with different uncertainty metrics. In our experiments, we evaluated both these approaches using three probability-based uncertainty metrics introduced in Section~\ref{sec:ProbabilityBacedUncertainty}: DeepGini, Vanilla, and Margin. Because TRC performance remained largely consistent across these metrics, we only report the results of these two approaches when utilizing DeepGini. The detailed results of applying these approaches with the two other metrics are available in our replication package~\cite{replicationpackage}.
As previously mentioned, we address this research question using subjects with a medium level of distribution shift between their source and target input sets. The TRC values across all subjects and budgets are presented in Figure~\ref{fig:ResultsRQ1}.

\begin{table*}[t]
    \centering   
    \small
    \caption{TRC improvement percentages achieved by MetaSel over the highest TRC achieved by baselines, under highly constrained selection budgets across subjects representing a medium severity level of distribution shift (severity level = 3)}
    \resizebox{0.99\textwidth}{!}{
    \begin{tabular}{   |c|c|    c|c||    c|c||     c|c||     c|c|  }
    \cline{1-10} 
     Input  &Corruption  &\multicolumn{2}{c||}{$b$ = 1\%} &\multicolumn{2}{c||}{$b$ = 3\%} &\multicolumn{2}{c||}{$b$ = 5\%} &\multicolumn{2}{c|}{$b$ = 10\%}\\ \cline{3-10}
     set        &type  &Imp.  &Baseline  &Imp.  &Baseline &Imp.  &Baseline &Imp.  &Baseline \\ \cline{1-10}
    \multirow{7}{*}{MNIST}  &C1  &11.8\% &Meta-model   &43.6\% &NNS      &51.4\% &NNS       &57.1\% &NNS \\ \cline{2-10}
                            &C2  &7.7\%  &Meta-model   &40\%   &NNS      &40.7\% &NNS       &12.5\% &NNS \\ \cline{2-10}
                            &C3  &43.8\% &Vanilla   &50\%   &DeepGini      &40\%   &Vanilla       &24.6\% &TestRank \\ \cline{2-10}
                            &C4  &14.3\% &Meta-model   &21.2\% &Meta-model      &18.2\% &NNS       &22.5\% &DATIS \\ \cline{2-10}
                            &C5  &27.5\% &Margin   &11.3\% &NNS      &15.1\% &NNS       &20\% &DeepGini \\ \cline{2-10}
                            &C6  &75\%   &Meta-model   &73.9\% &NNS      &61.5\% &NNS       &60\% &DeepGini \\ \cline{2-10}
                            &C7  &53.5\% &NNS   &43.9\% &NNS      &35.5\% &NNS       &1.1\% &Meta-model \\ \cline{1-10}

    \multirow{7}{*}{FMNIST} &C1  &46.2\% &DATIS   &49.2\% &DATIS      &36.3\% &DATIS       &21.3\% &Meta-model \\ \cline{2-10}
                            &C2  &61.5\% &Meta-model   &31.9\% &Meta-model      &24.2\% &Meta-model       &16.6\% &Meta-model \\ \cline{2-10}
                            &C3  &25\%   &DATIS   &39.3\% &DATIS      &34.9\% &DATIS       &22.6\% &DATIS \\ \cline{2-10}
                            &C4  &31.6\% &Meta-model   &10\%   &Meta-model      &13\%   &Meta-model       &12.5\% &Meta-model \\ \cline{2-10}
                            &C5  &50\%   &Meta-model   &32.5\% &Meta-model      &30.6\% &Meta-model       &16.4\% &DATIS \\ \cline{2-10}
                            &C6  &60\%   &Meta-model   &54.5\% &Meta-model      &49\%   &Meta-model       &34.9\% &Meta-model \\ \cline{2-10}
                            &C7  &20.8\% &Meta-model   &16.5\% &Meta-model      &17\%   &Meta-model       &16.2\% &Meta-model \\ \cline{1-10}                        

    \multirow{5}{*}{Cifar-10}&C1 &17.6\% &Meta-model   &13.2\% &Meta-model      &18\%   &Meta-model       &17.3\% &Meta-model \\ \cline{2-10}
                            &C2  &35\%   &Meta-model   &10.2\% &Meta-model      &23.3\% &Meta-model       &19\% &Meta-model \\ \cline{2-10}
                            &C3  &50\%   &DATIS   &24.6\% &DATIS      &16.8\% &DATIS       &19.9\% &DATIS \\ \cline{2-10}
                            &C6  &16.7\% &DATIS   &26.5\% &DATIS      &36.9\% &DATIS       &32.4\% &Meta-model \\ \cline{2-10}
                            &C7  &19\%   &NNS   &10.9\% &DATIS      &11.3\% &DATIS       &8.7\% &Meta-model \\ \cline{1-10} 

    \multirow{4}{*}{Cifar-100} &C1&0\%  &DATIS   &40\%   &Meta-model      &25.9\% &Meta-model       &37.9\% &Meta-model \\ \cline{2-10}
                            &C2  &25\%  &DATIS   &56.5\% &Meta-model      &60.9\% &Meta-model       &49.5\% &Meta-model \\ \cline{2-10}
                            &C3  &0\%   &DATIS   &31.6\% &DATIS      &53.4\% &DATIS       &54.5\% &DATIS \\ \cline{2-10}
                            &C4  &0\%   &Meta-model   &30.8\% &DATIS      &53.1\% &DATIS       &59.8\% &DATIS \\ \cline{1-10}
                                                   
    \multicolumn{2}{|c|}{Avg. over second-best}  &\multicolumn{2}{c||}{30.09\%}  &\multicolumn{2}{c||}{33.13\%}  &\multicolumn{2}{c||}{33.35\%}  &\multicolumn{2}{c|}{27.7\%}   \\ \cline{1-10}  

    \multicolumn{2}{|c|}{Avg. over Meta-model}  &\multicolumn{2}{c||}{45.77\%}  &\multicolumn{2}{c||}{42.37\%}  &\multicolumn{2}{c||}{41.06\%}  &\multicolumn{2}{c|}{33.36\%}   \\ \cline{1-10} 
    
    \end{tabular}    
     \label{tab:TRCimprovementsSeverity3}
     }
\end{table*}

We should note that most curves exhibit an initial decrease in TRC values, followed by an increase that eventually reaches TRC = 1. This initial decline can be explained by the specific denominator used in TRC calculations, as defined in Equation (\ref{Eq:TRC}). The denominator is determined as the minimum of the selection budget ($b$) and the total number of misclassifications ($|Mis_{total}|$) within the entire test set. For smaller budgets, where $b < |Mis_{total}|$, the denominator grows as $b$ increases, leading to a temporary decline in TRC. The turning point occurs when $b$ becomes equal to $|Mis_{total}|$ for the first time.
Beyond this point, the denominator remains constant at $|Mis_{total}|$, causing the TRC value to increase as the budget increases. 

As depicted in Figure~\ref{fig:ResultsRQ1}, MetaSel consistently achieves a higher TRC value compared to all baseline approaches across various subjects and budget sizes. The difference between MetaSel and the second-best performing baseline is larger for smaller budget sizes, highlighting the practical advantages of MetaSel in test selection for fine-tuned models under constrained selection budgets, which is a common situation in practice. This difference diminishes as the budget size increases, which is expected since, with larger budgets, the majority of misclassified inputs are already selected, causing all TRC curves to converge toward 1.

To investigate whether the observed performance improvements are statistically significant, we conducted a Wilcoxon signed-rank test at a significance level of $\alpha = 0.05$. 
In this analysis, we collected TRC values for MetaSel and all baseline approaches across all selection budgets for each subject. Paired comparisons of TRC values between MetaSel and each baseline revealed that all p-values were below 0.05, confirming that MetaSel significantly outperforms all SOTA baselines in identifying failures in DNNs.

Furthermore, we conducted a more detailed analysis of performance improvements provided by MetaSel, focusing specifically on highly constrained selection budgets: 1\%, 3\%, 5\%, and 10\% of the target test set size. Table~\ref{tab:TRCimprovementsSeverity3} presents the TRC improvement percentages achieved by MetaSel for each of these budget sizes compared to baselines. The improvement percentage at a given budget $b$ is calculated as follows:

\begin{equation}\label{Eq:Improvements}
\text{Imp} (b) = \left( \frac{TRC_{MetaSel}(b) - TRC_{bestBaseline}(b)}{1 - TRC_{bestBaseline}(b)} \right) \times 100
\end{equation}

\noindent where $TRC_{MetaSel}(b)$ represents the TRC value achieved by MetaSel at budget $b$ and $TRC_{bestBaseline}(b)$ is the highest TRC value achieved by any of the baselines on the same subject and budget size. 
This formula accounts for the maximum possible improvement for each subject by considering the gap between the highest TRC value achieved by baselines and the theoretical maximum (TRC = 1) in the denominator. The improvement percentage is zero if $TRC_{bestBaseline}(b) = 1$, as there is no room for further improvement.

Using a simpler improvement percentage formula\footnote{$\left( \frac{TRC_{MetaSel}(b) - TRC_{bestBaseline}(b)}{TRC_{bestBaseline}(b)} \right) \times 100$}, solely based on the TRC value achieved by baselines, might overstate smaller gains. Consequently, the improvement percentage calculated based on Equation~(\ref{Eq:Improvements}) ensures a more accurate analysis of MetaSel's effectiveness relative to baselines. Furthermore, compared to calculating absolute improvements, this formula normalizes improvements and allows for a meaningful comparison across budgets and subjects.

Most reported improvement percentages in Table~\ref{tab:TRCimprovementsSeverity3} are positive across all subjects and selection budgets, confirming MetaSel's superior performance. 
The only exceptions are three cases for a 1\% budget on subjects derived from the Cifar-100 input set, where there is no improvement. This occurs because both MetaSel and the second-best approach achieve identical TRCs that are very high, equal or close to 1, leaving no room for further improvement. 
Moreover, these results demonstrate that MetaSel consistently delivers a practically significant improvement across all scenarios, as detailed next.

While for certain subjects the TRC improvement achieved by MetaSel is exceptionally high--such as an 
improvement percentage of 75\% on MNIST with corruption type C6 at $b = 1\%$--
MetaSel demonstrates a robust overall performance with average improvements of 30.09\% for a 1\% budget, 33.13\% for a 3\% budget, 33.35\% for a 5\% budget, and 27.7\% for a 10\% budget across all subjects.

Table~\ref{tab:TRCimprovementsSeverity3} additionally identifies the baseline that achieves the highest TRC after MetaSel for each selection budget and subject. It is important to highlight that the second-best performing approach after MetaSel varies significantly across subjects and selection budgets. Among the 92 combinations of subjects and budget sizes presented in Table~\ref{tab:TRCimprovementsSeverity3}, Meta-model, proposed by Demir \textit{et al.}~\cite{demir2024test}, emerged as the second-best performing approach after MetaSel in 43 cases, followed by DATIS~\cite{li2024distance} (27 cases).
This variability underscores the usefulness of MetaSel, making it the only approach we can confidently rely on, regardless of the subject or budget size. In contrast, selecting an alternative method poses substantial risks, as there is no reliable way to predict a priori which baseline will perform best for a given scenario in the absence of MetaSel.

Given such variability, it is reasonable to assume that practitioners would rely on Meta-model, the most frequent second-best performing approach following MetaSel, as an alternative to MetaSel. Consequently, in the last row of Table~\ref{tab:TRCimprovementsSeverity3}, we report the average improvement percentage provided by MetaSel over Meta-model. The improvement percentages are calculated using the same calculation from Equation~(\ref{Eq:Improvements}), replacing $TRC_{bestBaseline}(b)$ with $TRC_{Meta-model}(b)$ for all subjects. Our results show that MetaSel provides substantial performance improvements over Meta-model with an average improvement of 45.77\% for a 1\% budget, 42.37\% for a 3\% budget, 41.06\% for a 5\% budget, and 33.36\% for a 10\% budget across all subjects. These findings further confirm the consistent superiority of MetaSel over Meta-model, demonstrating its effectiveness in maximizing misclassification detection across all subjects and selection budgets.

\begin{figure}[ht!]
  \centering
  \begin{minipage}[b]{0.44\linewidth}
    \centering
    \includegraphics[width=\linewidth]{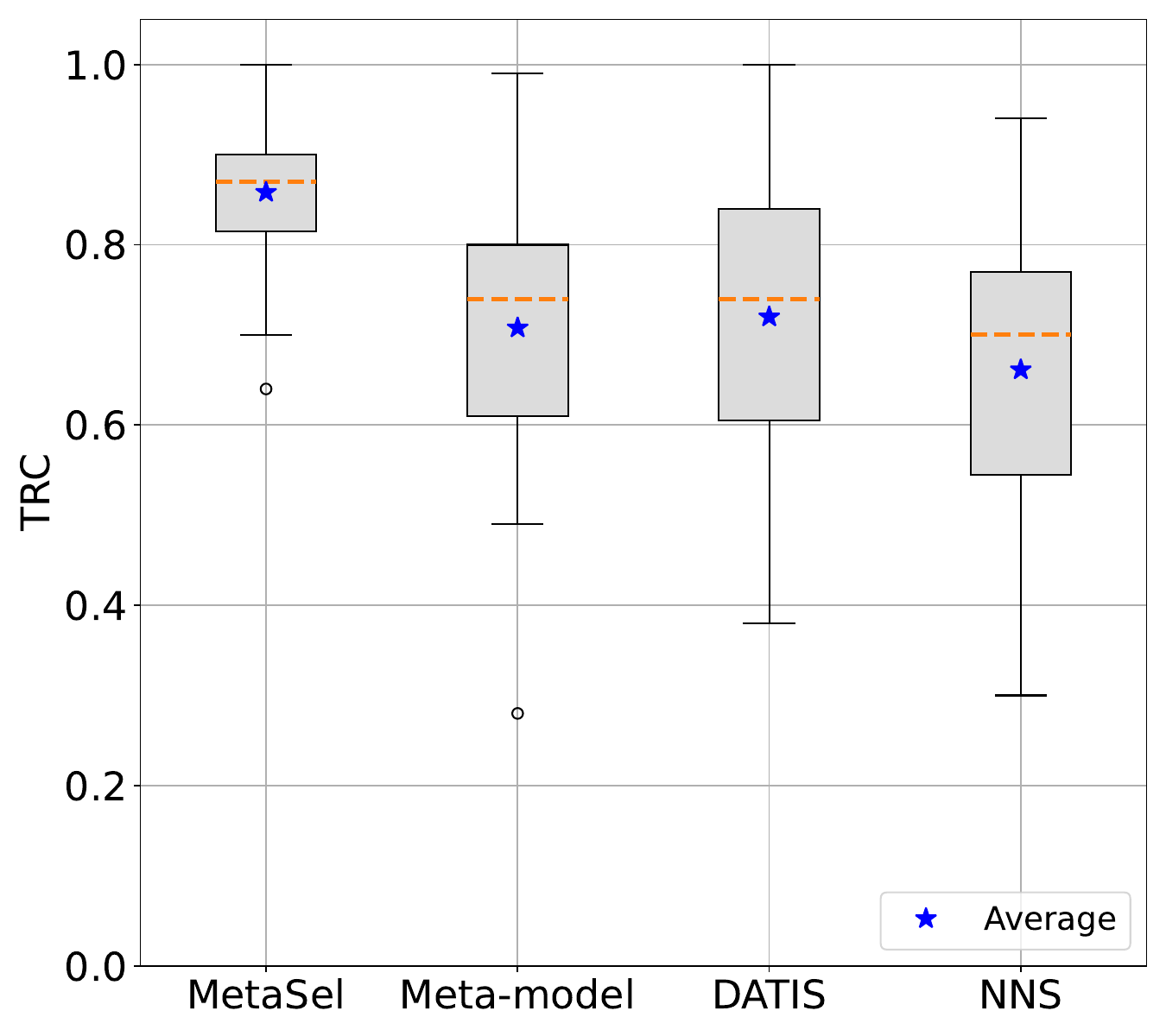}
    \par\vspace{-3pt}
             $b$ = 1\%
  \end{minipage}
  \hfill
  \begin{minipage}[b]{0.44\linewidth}
    \centering
    \includegraphics[width=\linewidth]{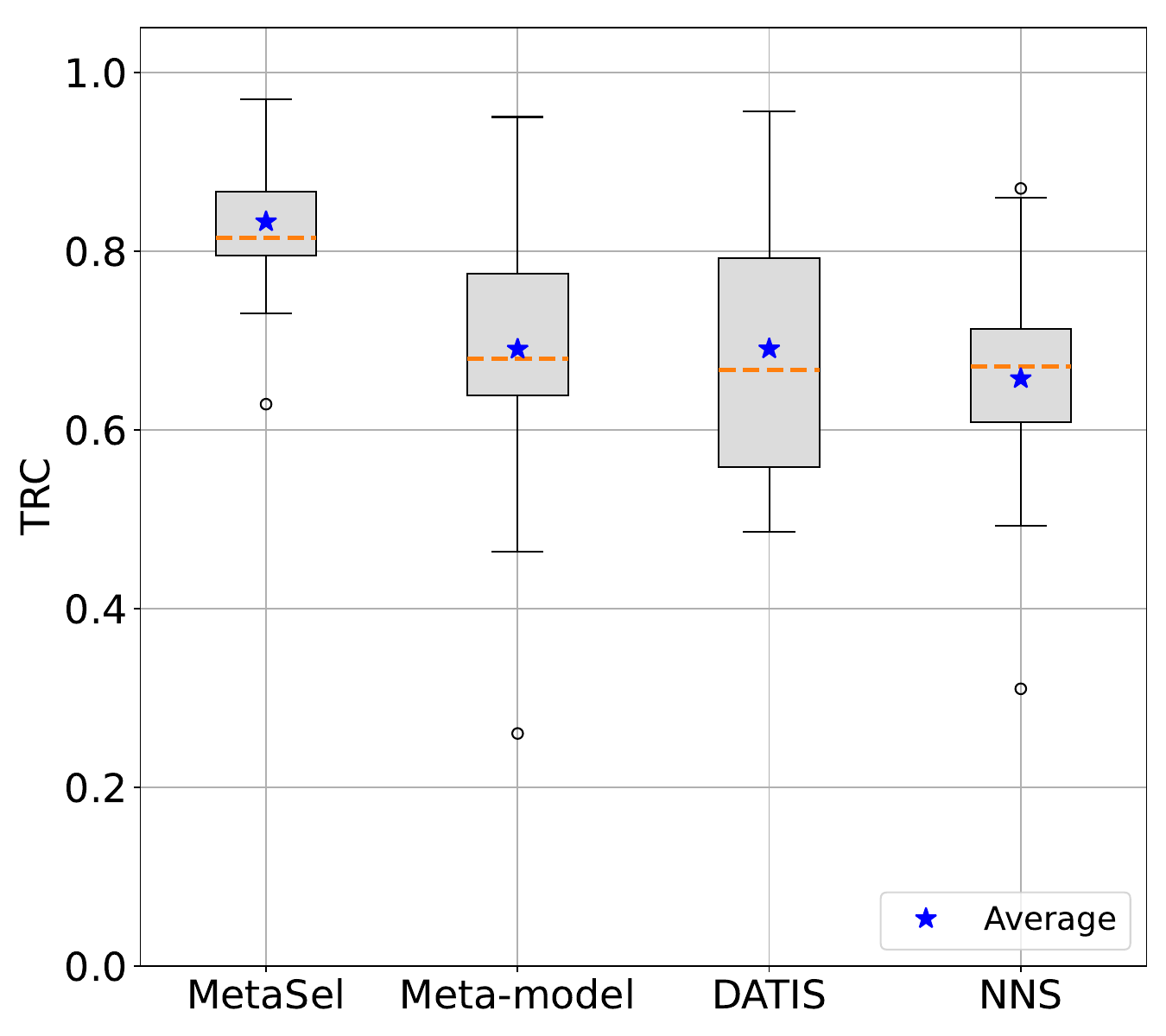}
    \par\vspace{-3pt}
         $b$ = 3\%
  \end{minipage}
  \hfill
    \par\vspace{5pt}
  \begin{minipage}[b]{0.44\linewidth}
    \centering
    \includegraphics[width=\linewidth]{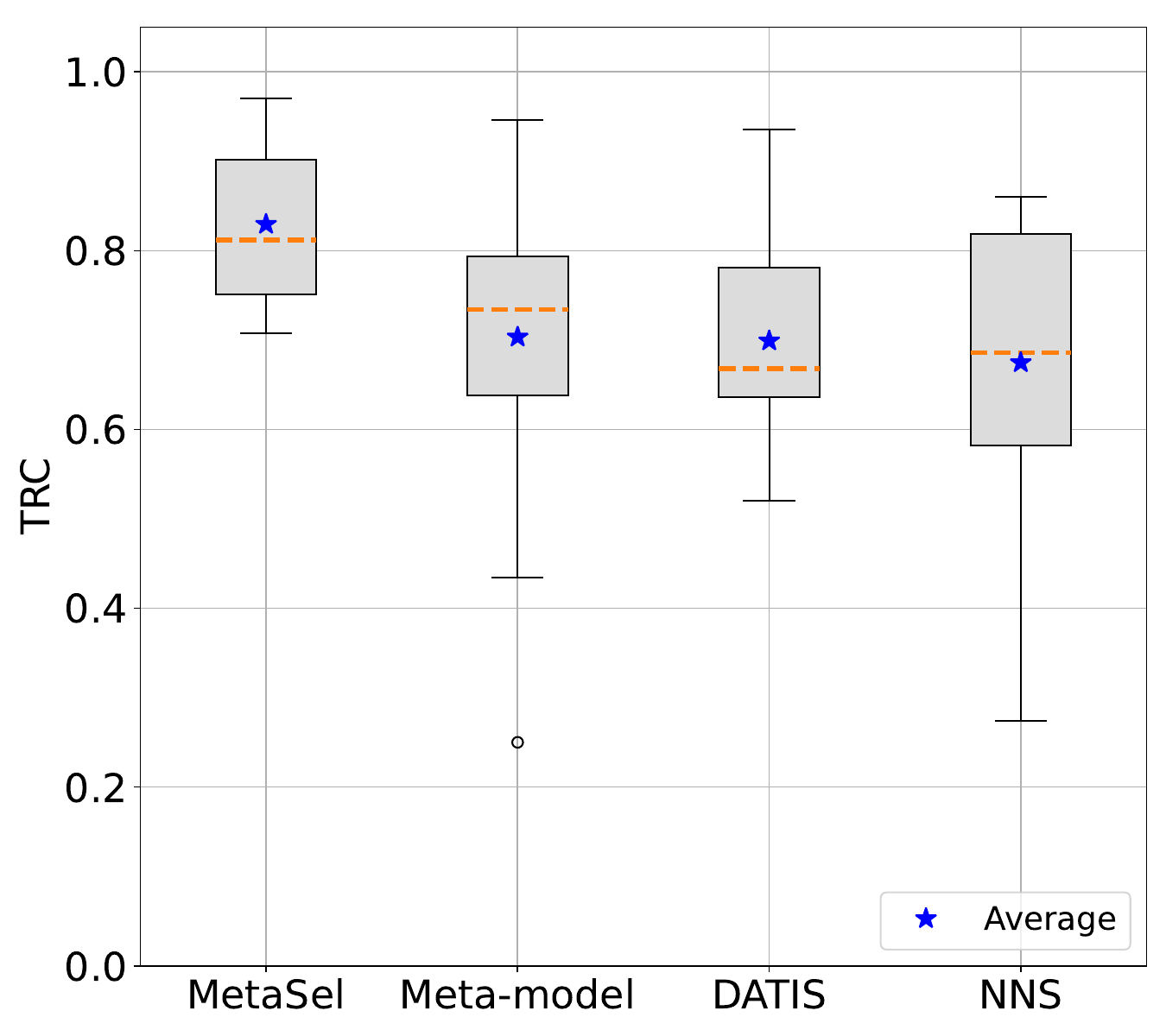}
    \par\vspace{-3pt}
     $b$ = 5\%
  \end{minipage}
  \hfill
  \begin{minipage}[b]{0.44\linewidth}
    \centering
    \includegraphics[width=\linewidth]{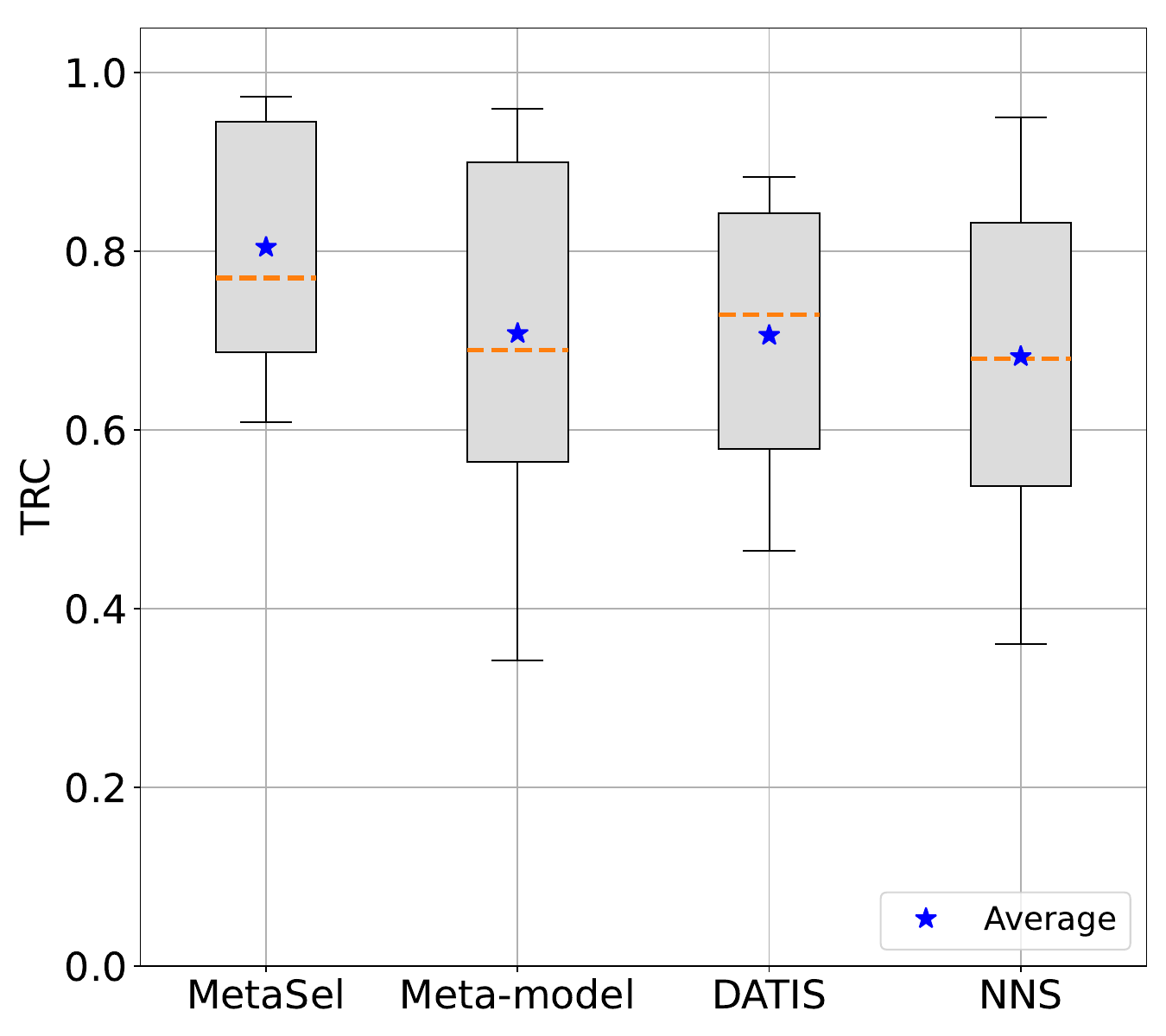}
    \par\vspace{-3pt}
    $b$ = 10\%
  \end{minipage}
    \par\vspace{5pt}

    \caption{TRC distribution across all subjects under highly constrained selection budgets, across subjects representing a medium severity level of distribution shift (severity level = 3)}
  \label{fig:ResultsRQ1_boxplots}
\end{figure}

We further analyze the distribution of TRC achieved by MetaSel in comparison to the three most-frequently occurring second-best baselines in Table~\ref{tab:TRCimprovementsSeverity3}, namely Meta-model, DATIS, and NNS. We do so under highly constrained selection budgets across all subjects, to gain deeper insights into the effectiveness advantages of MetaSel over these alternatives. The results are presented in Figure~\ref{fig:ResultsRQ1_boxplots}.

Each box plot in Figure~\ref{fig:ResultsRQ1_boxplots}, illustrates the TRC distribution achieved by MetaSel, Meta-model, DATIS, and NNS for a given selection budget $b$.
MetaSel consistently achieves the highest TRC median and average values across all budget sizes, demonstrating its superior effectiveness in test selection compared to the other three approaches. 
While TRC values for all approaches show a higher variance at a selection budget of $b = 10\%$, this is primarily due to the initial decline in TRC curves as the budget size increases (see Figure~\ref{fig:ResultsRQ1}), a consequence of the specific denominator used in TRC calculation (Equation~(\ref{Eq:TRC})).
Nevertheless, MetaSel exhibits a significantly smaller variance, thus maintaining a more consistent performance across all subjects.

Moreover, MetaSel guarantees a high minimum TRC of 0.61 across all subjects and budget sizes, thus making it a reliable solution. 
In contrast, Meta-model exhibits greater variability in TRC distributions, with a lower median TRC than MetaSel across all budget sizes. Relying on Meta-model introduces a risk of significantly lower performance, with a notably lower minimum TRC of less than 0.3 for some subjects. This is mainly because, despite being the most frequent second-best approach, Meta-model demonstrates inconsistent performance, delivering significantly poor results on certain subjects, such as Cifar-100 with Spatter corruption type, as shown in Figure~\ref{fig:ResultsRQ1}. Further, compared to Meta-model, MetaSel also enhances the maximum TRC achieved across all subjects and budget sizes, further validating its superiority in test selection effectiveness.

Both DATIS and NNS demonstrate significantly lower TRC medians compared to MetaSel. They also show greater variability in TRC values, introducing the risk of ineffective test selection for some subjects.
These findings confirm MetaSel's superiority and its ability to maintain consistently higher TRC under tight selection budgets, making it the most effective selection approach among all evaluated methods. Moreover, the results highlight the risk of relying on alternative baselines, even the most frequent second-best approaches, like Meta-model, DATIS, or NNS, as they may lead to significantly poor performance.
\TR{C3.1}{
Furthermore, these findings empirically support the core assumption underlying the design of MetaSel---namely, that target test inputs causing behavioral divergence between a fine-tuned model and its pre-trained counterpart are more likely to be misclassified by the fine-tuned model. }

\begin{tcolorbox}   
    \textbf{Answer to RQ1:} MetaSel consistently outperforms all baselines in selecting test subsets that contain a higher number of inputs misclassified by the fine-tuned DNN model, within the same selection budget. Furthermore, MetaSel demonstrates statistically and practically significant performance improvements on highly constrained budgets, delivering average TRC improvement percentages of 30.09\% for a 1\% budget, 33.13\% for a 3\% budget, 33.35\% for a 5\% budget, and 27.7\% for a 10\% budget, relative to the second-best performing approach. 
   Further, the second-best approach after MetaSel greatly varies across subjects and budgets. In contrast, MetaSel maintains a significantly lower variance in TRC values across subjects compared to the most frequently occurring second-best approaches: Meta-Model, DATIS, and NNS.
   These findings further underscore MetaSel's superiority, confirming its reliability as the only approach that consistently delivers superior results, regardless of the subject or selection budget.    
\end{tcolorbox}

\begin{table*}[b]
    \centering   
    \caption{Execution time (in seconds) of MetaSel and each baseline when ranking the entire target test set once.}
    \resizebox{0.99\textwidth}{!}{
    \begin{tabular}{   |cc|    c|    c|c|c|c|c|c|c|c|c|c|c|  }
    \cline{1-14} 
     \multirow{2}{*}{DNN model} &\multirow{2}{*}{Input set}   &\multirow{2}{*}{MetaSel} &\multicolumn{11}{c|}{Baselines}  \\ \cline{4-14}
     &          &   &Gini &Margin &Vanilla   &DSA &MDSA &LSA   &TestRank &Meta-model   &DATIS &NNS &\Rev{RTS}\\ \cline{1-14}
    LeNet5 &MNIST     &10.45   &1.16 &0.95 &0.74  &165.07 &1.08 &5.40   &79.83 &9.23    &8.98 &5.04 &\Rev{141.30} \\ \cline{1-14}
                            
    LeNet5 &FMNIST    &10.64   &1.15 &0.99 &0.97  &172.69 &1.09 &11.03  &97.29 &11.69    &10.11 &4.85 &\Rev{154.76} \\ \cline{1-14}

    ResNet20 &Cifar-10  &51.02   &2.14 &2.02 &2.01  &167.40 &2.98 &9.22   &92.46 &24.81    &9.74 &5.55 &\Rev{184.64} \\ \cline{1-14}
                      
    ResNet152 &Cifar-100 &277.95  &27.14&25.65&25.58  &228.40&48.32&63.32  &118.07&382.05    &91.41 &83.25 &\Rev{1282.75} \\ \cline{1-14}                                      
    \end{tabular}    
     \label{tab:time}
     }
\end{table*}

\subsection{\textbf{RQ2: MetaSel's efficiency}}

To answer this research question, we empirically evaluate the execution time of MetaSel and compare it against SOTA baselines. Specifically, we measure the time required for each approach to estimate the probability of misclassification for all inputs in a target test set. It is important to note that, similar to any learning-based approach, MetaSel involves an initial training phase. However, this initial phase only needs to be performed once for each fine-tuned model. Moreover, compared to other learning-based approaches like Meta-model proposed by Demir \textit{et al.}~\cite{demir2024test}, MetaSel's initial training phase is considerably more efficient. For instance, in our experiments, training MetaSel for our most complex subject DNN model, ResNet152 with the Cifar-100 input set, took approximately 15 minutes. In contrast, the approach by Demir et al.\cite{demir2024test}, which involves training five deep ensemble DNN models with the same architecture as the fine-tuned model, followed by the training of a logistic regression model as the Meta-model, required approximately 130 minutes for the same subject.
Once trained, MetaSel can predict the misclassification probabilities for any number of unlabeled test inputs, making it even more cost-effective for larger test sets. 

Table~\ref{tab:time} provides the execution time (in seconds) for MetaSel and each baseline when ranking the entire target test set. These execution times are measured only for one of the subjects created from each input set, as they remain identical across all subjects derived from the same input set.

As shown in Table~\ref{tab:time}, while MetaSel generally requires more time for test input ranking compared to simpler and more efficient approaches, such as probability-based uncertainty metrics, it remains a practical solution, even for highly complex DNNs and input sets with a large number of output classes. For example, it requires approximately 278 seconds to rank the entire Cifar-100 test set on a ResNet152 model. It is important to note that the required time for manually labeling test inputs, specifically in the context of fine-tuned models with an abundance of unlabeled test inputs, is far more expensive than MetaSel's execution time. Moreover, since test input selection is neither frequent nor a real-time task, within acceptable bounds, an increase in execution time is justifiable when leading to improved test effectiveness.

It is important to note that to ensure a fair and consistent comparison, we report execution times in this table on a server equipped with an Intel(R) Xeon(R) Gold 6234 CPU (3.30GHz) and an Nvidia Quadro RTX 6000 GPU with 24GB of memory. However, MetaSel's execution time can be significantly reduced when deployed in industrial computing environments since they are often more powerful and feature enhanced parallelization, thus further confirming the practicality of MetaSel.

\begin{table*}[b]
    \centering   
    \footnotesize 
    \small
    \caption{Average TRC improvement percentages achieved by MetaSel over the three most frequently occurring second-best baselines, under highly constrained selection budgets, across subjects representing weak (severity level = 2) and strong (severity level = 4) distribution shift severity}
    \resizebox{0.99\textwidth}{!}{
    \begin{tabular}{|c|c|c|c|c||c|c|c|c|}
    \hline
    Average Imp. & \multicolumn{4}{c||}{severity level = 2} & \multicolumn{4}{c|}{severity level = 4} \\
    \cline{2-9}
    over Baselines&$b$ = 1\% &$b$ = 3\% & $b$ = 5\% & $b$ = 10\% & $b$ = 1\% & $b$ = 3\% & $b$ = 5\% & $b$ = 10\% \\
    \hline
    Meta-model &45.07\% &41.29\% &37.22\% &30.97\% &41.23\% &35.72\% &34.70\% &28.46\%  \\
    
    \hline
    DATIS &36.74\% &39.17\% &38.86\% &37.90\% &38.42\% &31.55\% &35.61\% &35.99\% \\
    
    \hline
    NNS &56.18\% &48.38\% &44.02\% &38.87\% &53.46\% &45.45\% &42.45\% &33.42\%  \\
    
    \hline
    \end{tabular}    
     \label{tab:RQ3TRCimprovements}
     }
\end{table*}

While MetaSel, similar to many baseline approaches, exhibits an increase in execution time as the complexity of the DNN model and the size of the test set grow, it is not the most time-consuming approach. When applied to subjects with relatively simple architectures, such as LeNet-5 or ResNet-20, and input sets with only 10 output classes (i.e., MNIST, FMNIST, or Cifar-10), MetaSel demonstrates efficiency outperforming baselines such as TestRank, DSA, \TR{C1.6, C3.8}{and RTS} in terms of execution time. 
The primary factors driving MetaSel's execution time are the depth and complexity of the DNN model, as well as the number of output classes. This is primarily due to the requirement for ODIN score calculations for test inputs based on both the pre-trained and fine-tuned models, which requires significantly more computations for forward and backward passes to calculate the ODIN scores on deeper networks such as ResNet152. 
Additionally, the size of the logit layer, which corresponds to the number of output classes, further increases computational requirements for input sets like Cifar-100. Despite these challenges, MetaSel maintains acceptable efficiency even when applied to deeper networks like ResNet152 and input sets with a larger number of output classes, such as Cifar-100 with 100 classes. Indeed, MetaSel requires approximately 278 seconds to rank the entire test set, a remarkably lower execution time compared to Meta-model, the most viable alternative for practitioners based on effectiveness, which takes around 382 seconds. The increased execution time, combined with the longer initial training phase required by Meta-model, highlights the considerably more time-intensive nature of this method compared to MetaSel. 
\TR{C1.6, C3.8}{
Notably, RTS~\cite{sun2023robust} is the most computationally expensive approach, requiring 4 to 15 times more time than MetaSel to prioritize the entire test set across investigated subjects. This significant overhead primarily stems from its diversity estimation step, which involves computing pairwise similarities between output probability distributions using SSIM for each test input against a sampled subset of training inputs. This computation substantially increases the runtime, particularly for larger input sets. For instance, applying RTS on the Cifar-100 input set takes approximately 1282 seconds.
}

Given our earlier findings, where MetaSel consistently outperforms all baselines, delivering a practically significant effectiveness improvement, particularly under highly constrained selection budgets, MetaSel's execution time is acceptable, even for complex DNNs such as ResNet152 and input sets such as Cifar-100 with 100 classes. 
MetaSel is thus a cost-effective alternative as it not only delivers significant performance improvements but also remains acceptably efficient, even on deep and more complex networks.

\begin{tcolorbox}   
    \textbf{Answer to RQ2:} MetaSel's efficiency, though lower than many baselines, remains acceptable from a practical standpoint, even for subjects with large input sets featuring numerous output classes, and deep, complex DNN architectures. Its efficiency is even better than \Rev{some baselines, including} the second-best baseline in terms of effectiveness, namely Meta-model.  MetaSel is thus a cost-effective approach. 
\end{tcolorbox}

\subsection{\textbf{RQ3: MetaSel's effectiveness across varying levels of distribution shifts}}

To address the first two research questions, we conducted our experiments on subjects representing a medium level of distribution shift (severity level = 3), as mentioned earlier. To answer this research question, we further extend our analysis by repeating the experiments from RQ1 using two additional severity levels representing both weaker (severity level = 2) and stronger (severity level = 4) distribution shifts. Such analysis offers valuable insights into MetaSel's robustness to real-world scenarios where distribution shifts can vary in severity. To this end, we utilized the same models and source inputs detailed in Table~\ref{tab:SourceModels}, and employed the same corruption types we used in our earlier experiments.

Similar to RQ1, we calculated the TRC achieved by MetaSel and each baseline using Equation~(\ref{Eq:Improvements}) across the same range of budget sizes. As detailed below, the results demonstrate that MetaSel consistently and significantly outperforms all baselines, achieving higher TRC values across all severity levels, subjects, and selection budgets. To assess the statistical significance of these performance improvements, we once again conducted Wilcoxon signed-rank tests with a significance level of $\alpha = 0.05$. Results show that all p-values are below 0.05, confirming MetaSel's consistently superior performance across varying severity levels of distribution shift. To provide a comprehensive report, we have included the full results of our statistical analysis, detailed TRC values, and TRC curves for MetaSel and all baseline approaches across all severity levels of distribution shift, subjects, and selection budgets in our replication package~\cite{replicationpackage}.

Consistent with our earlier findings on subjects with medium levels of distribution shift, the second-best performing approach after MetaSel also varies across subjects, selection budgets, and levels of distribution shift. These findings further confirm that MetaSel is reliably outperforming baselines, regardless of subjects, selection budgets, and distribution shift levels.
The three most frequently occurring second-best baselines, in subjects with both weaker and stronger severity levels of distribution shifts, are once again Meta-model, DATIS, and NNS, similar to what was observed in RQ1. Consequently, we only report in Table~\ref{tab:RQ3TRCimprovements} the average improvement percentages provided by MetaSel over these three baselines. Similar to RQ1, we focused on highly constrained budget sizes of 1\%, 3\%, 5\%, and 10\% of the target test set's size. 
Consistent with our earlier findings in RQ1 on subjects with a medium level of distribution shift, MetaSel yields practically significant improvements relative to the best-performing baselines in subjects with both weaker and stronger shift severity levels. For subjects with weaker distribution shifts (severity level = 2), MetaSel achieves average improvement percentages ranging from 30.97\% to 45.07\% over Meta-model, 36.74\% to 39.17\% over DATIS, and 38.87\% to 56.18\% over NNS. Similarly, for subjects with stronger distribution shifts (severity level = 4), MetaSel maintains its superiority, achieving average improvement percentages of 28.46\% to 41.23\% over Meta-model, 31.55\% to 38.42\% over DATIS, and 33.42\% to 53.46\% over NNS.
These results highlight that MetaSel consistently provides a practically significant improvement in misclassification detection across a large distribution shift severity range. They thus confirm MetaSel's wide applicability across diverse distribution shift situations in practice.

\begin{figure}[ht!]
  \centering
  \begin{minipage}[b]{0.48\linewidth}
    \centering
    \includegraphics[width=\linewidth]{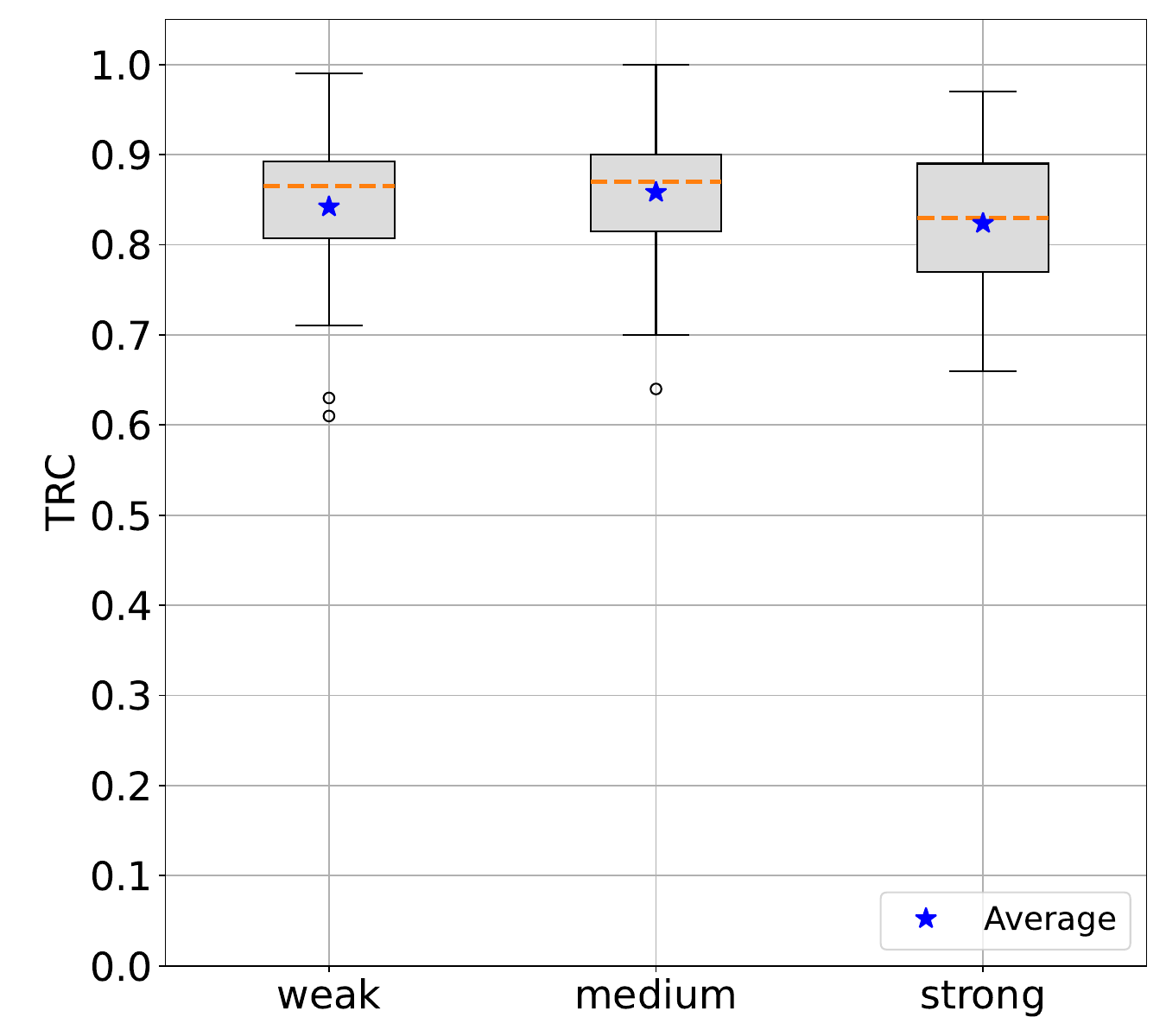}
    \par\vspace{-3pt}
             $b$ = 1\%
  \end{minipage}
  \hfill
  \begin{minipage}[b]{0.48\linewidth}
    \centering
    \includegraphics[width=\linewidth]{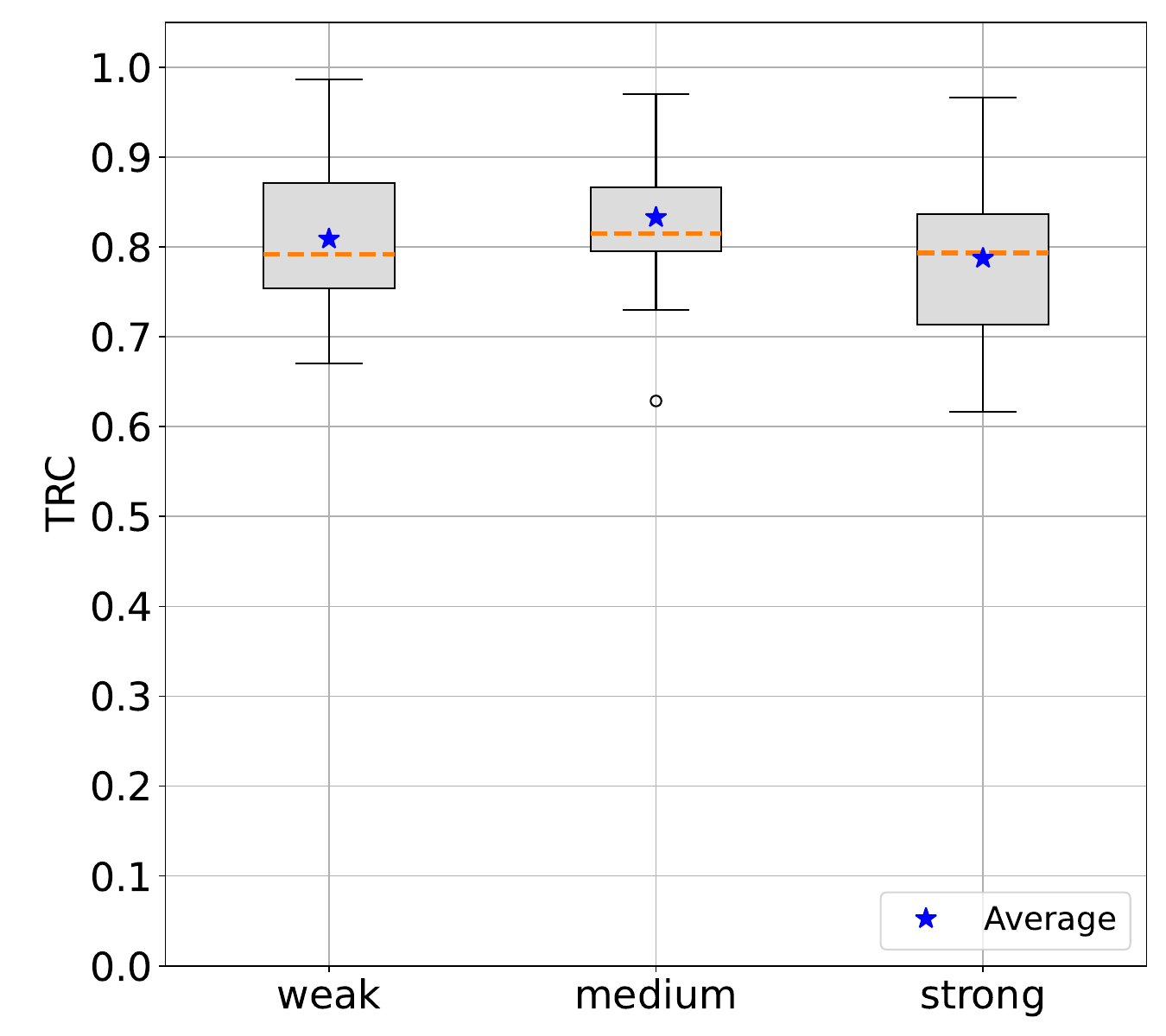}
    \par\vspace{-3pt}
         $b$ = 3\%
  \end{minipage}
  \hfill
  \begin{minipage}[b]{0.48\linewidth}
    \centering
    \includegraphics[width=\linewidth]{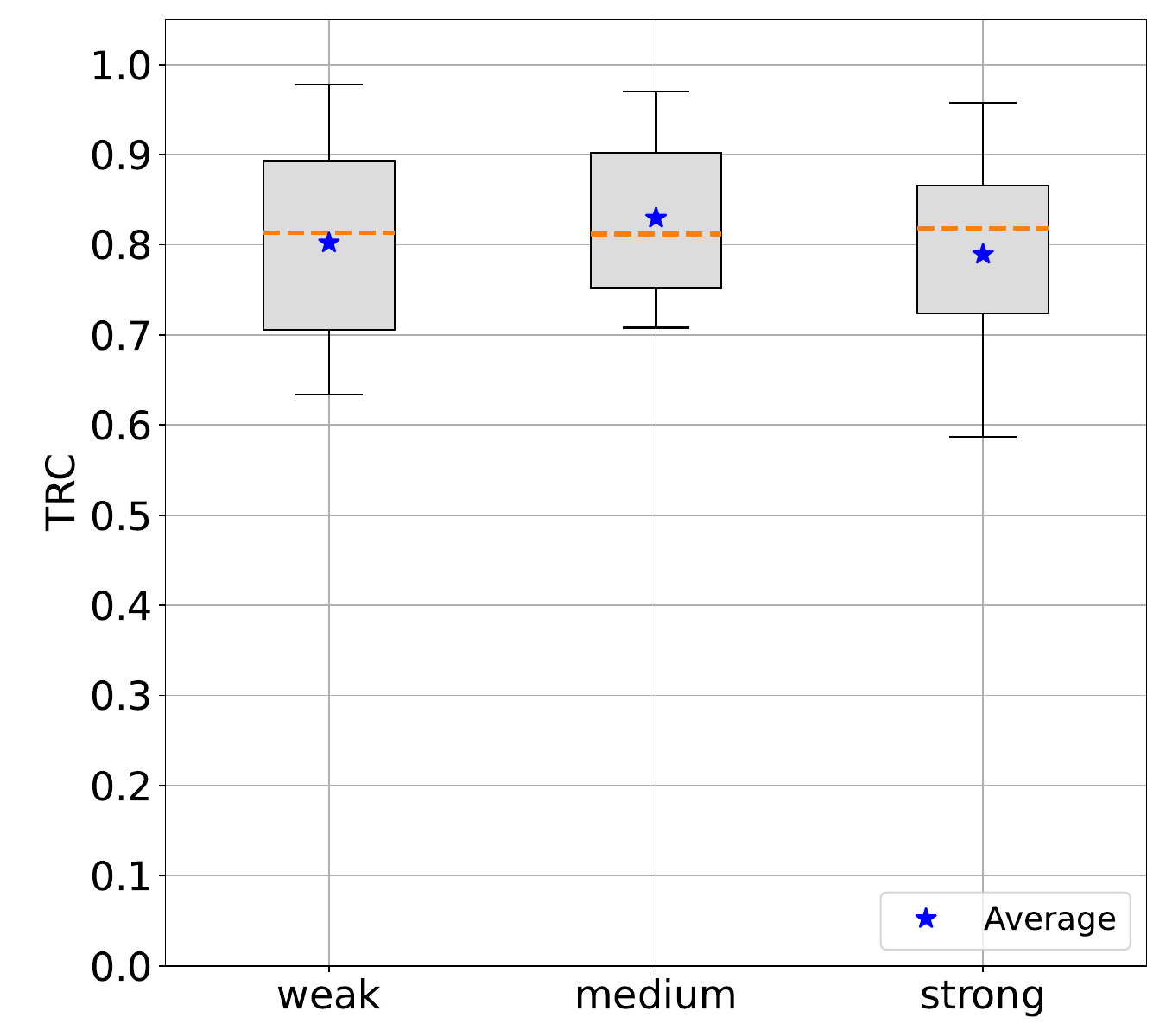}
    \par\vspace{-3pt}
     $b$ = 5\%
  \end{minipage}
  \hfill
  \begin{minipage}[b]{0.48\linewidth}
    \centering
    \includegraphics[width=\linewidth]{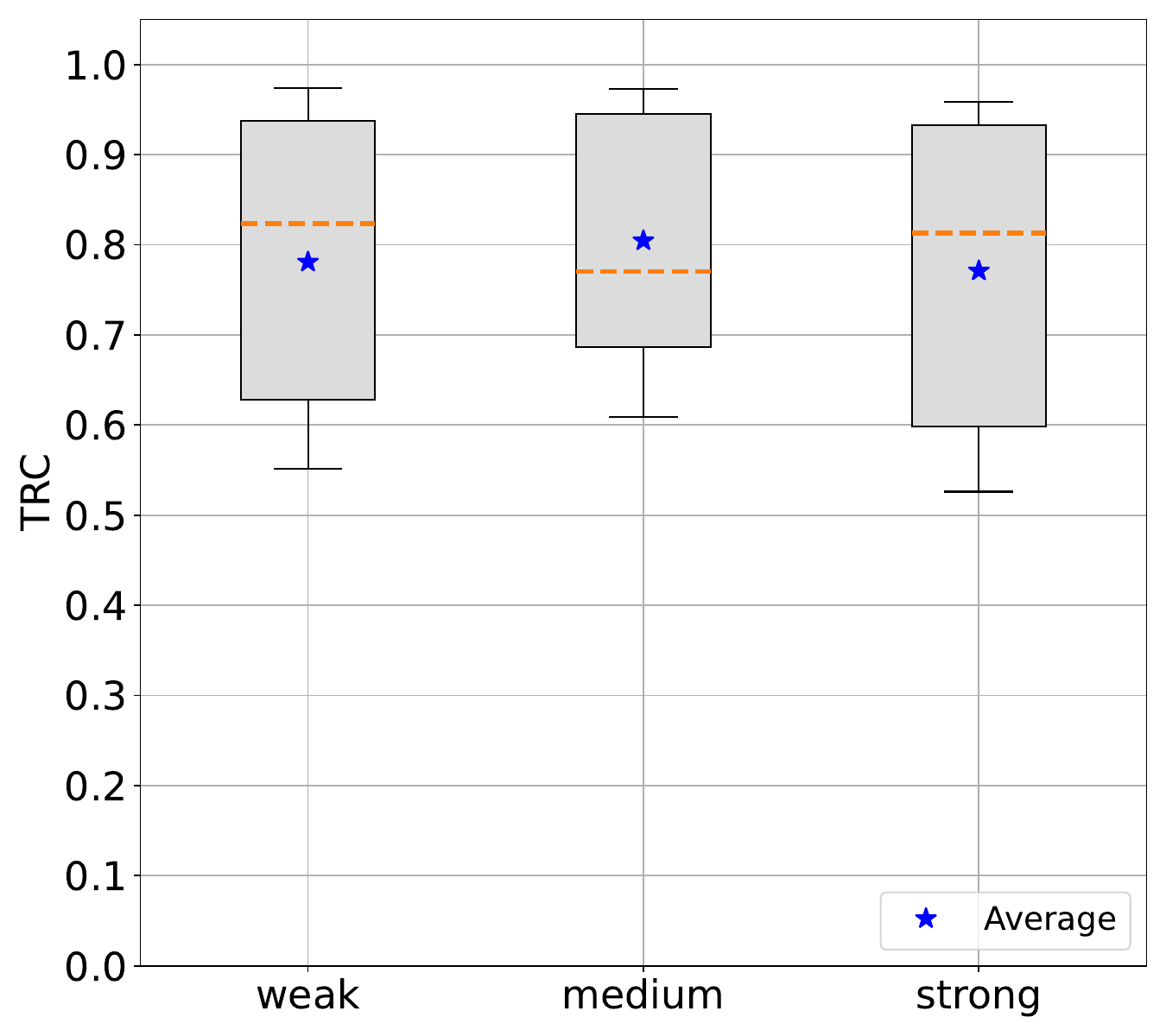}
    \par\vspace{-3pt}
    $b$ = 10\%
  \end{minipage}

    \caption{MetaSel's effectiveness across varying severity levels of distribution shift}
  \label{fig:RQ3MetaSelBoxplots}
\end{figure}

To provide further insights into MetaSel's effectiveness across varying levels of distribution shift, we present in Figure~\ref{fig:RQ3MetaSelBoxplots}, the distribution of TRC values achieved by MetaSel under highly constrained selection budgets across weak, medium, and strong distribution shift severity levels. 
These results show that MetaSel's performance remains unaffected by the degree of distribution shift between source and target sets. 
MetaSel continues to achieve significantly high TRC median, average, and upper quartile values across all budget sizes, demonstrating its effectiveness in test selection.

Despite the increased variance in TRC values for a selection budget of $b = 10\%$,  a trend observed across all baselines and consistent with our findings for RQ1, MetaSel maintains a consistently low variance across severity levels, ensuring stable performance while achieving a high minimum TRC.
These findings further confirm MetaSel’s superiority and reliability, even when applied to subjects with varying levels of distribution shifts between the source and target input sets.

\begin{tcolorbox}   
    \textbf{Answer to RQ3:} MetaSel proves to be a robust solution, consistently maintaining its effectiveness across a wide range of severity levels in distribution shift between source and target input sets.
    In addition to maintaining its effectiveness in absolute terms, by achieving high TRC values, MetaSel also remains beneficial relative to baselines, delivering significant improvements in TRC values across varying severity levels.
\end{tcolorbox}

\subsection{\textbf{RQ4: Contribution of MetaSel's training features}}
\label{sec:ablationStudy}
\TR{C1.4, C2.2, C3.2}{
To address this research question, we conducted a comprehensive ablation study comparing MetaSel with seven ablated variants created by both single-feature and grouped-feature ablation strategies. }

\Rev{
As described in Section~\ref{sec:MetaSelFeatureSet}, MetaSel's training feature set comprises six features obtained from both source and target models, including
the logit vector of the source model ($L_S$), the logit vector of the target model ($L_T$), the absolute difference between the two logits ($L_S - L_T$), the ODIN score based on $M_S$ ($ODIN_S$), the ODIN score based on $M_T$ ($ODIN_T$), and a binary indicator of differential testing outcome (1 if $M_S$ and $M_T$ predict the same class, 0 otherwise). 
We systematically constructed seven variants of MetaSel by excluding one or more features from this set.}

\Rev{
The variants are as follows: 
\begin{itemize}
    \item V1: excludes the difference between logits ($L_S - L_T$), 
    \item V2: excludes the differential testing outcome, 
    \item V3: excludes $ODIN_S$, 
    \item V4: excludes both ODIN scores ($ODIN_S$ and $ODIN_T$), 
    \item V5: excludes $L_S$ and thus $L_S - L_T$, 
    \item V6: excludes logits ($L_S$, $L_T$, and thus $L_S - L_T$), 
    \item V7: excludes all information from $M_S$ and thus includes only $L_T$ and $ODIN_T$.
\end{itemize}}


\Rev{
Variants such as V1, V2, and V3 represent single-feature ablation, where only one feature is removed to isolate its contribution. In contrast, variants such as V4, V5, V6, and V7 represent grouped feature ablation, where multiple features are removed together based on their origin or semantic grouping (e.g., features originated from the source model or OOD-based features). By comparing the performance of these variants to the original MetaSel model, we evaluate the contribution of each feature or group of features to MetaSel's ability to effectively select misclassified target test inputs. }

\begin{figure}[ht!]
  \centering
  \begin{minipage}[b]{0.48\linewidth}
    \centering
    \includegraphics[width=\linewidth]{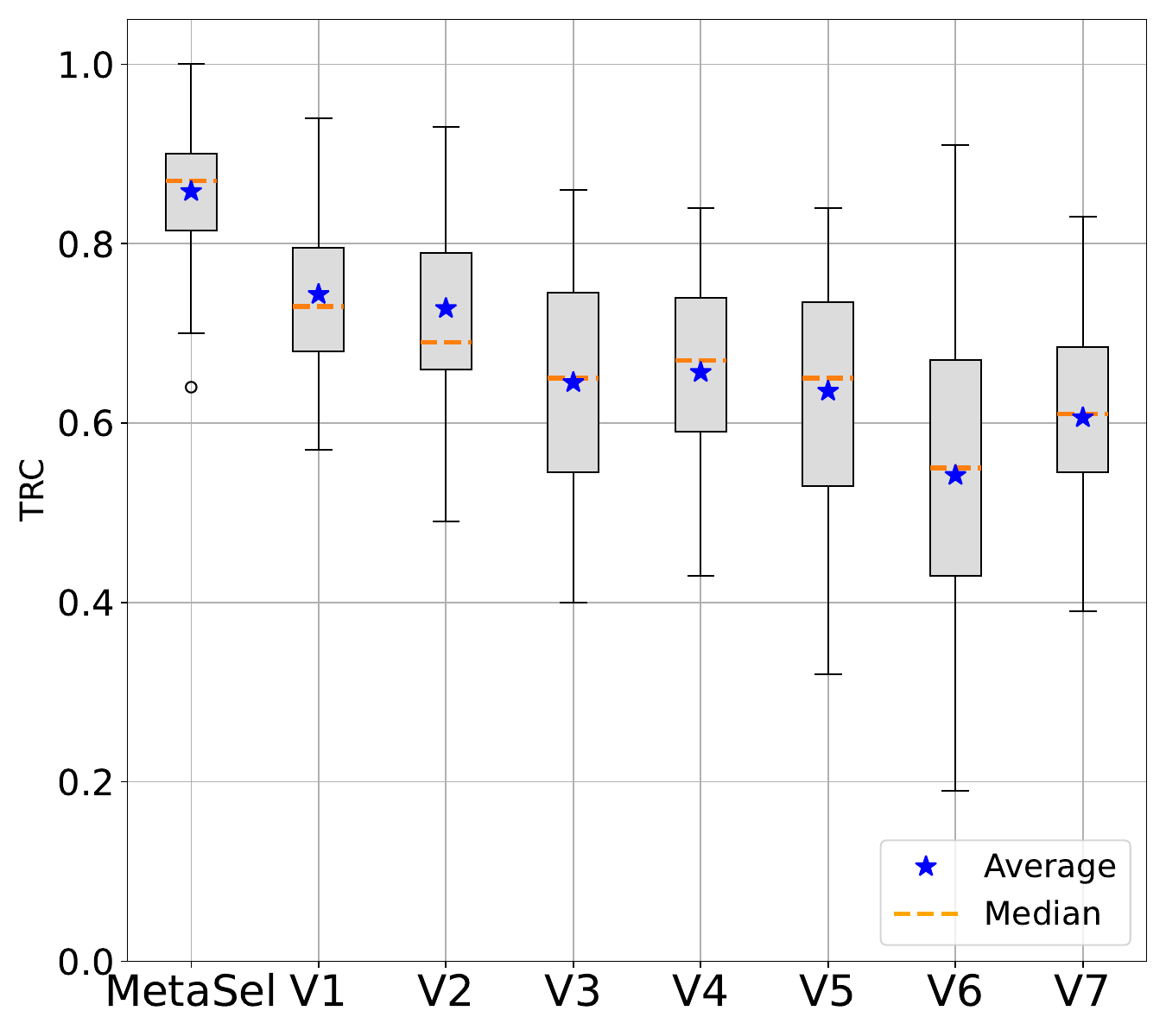}
    \par\vspace{-3pt}
             $b$ = 1\%
  \end{minipage}
  \hfill
  \begin{minipage}[b]{0.48\linewidth}
    \centering
    \includegraphics[width=\linewidth]{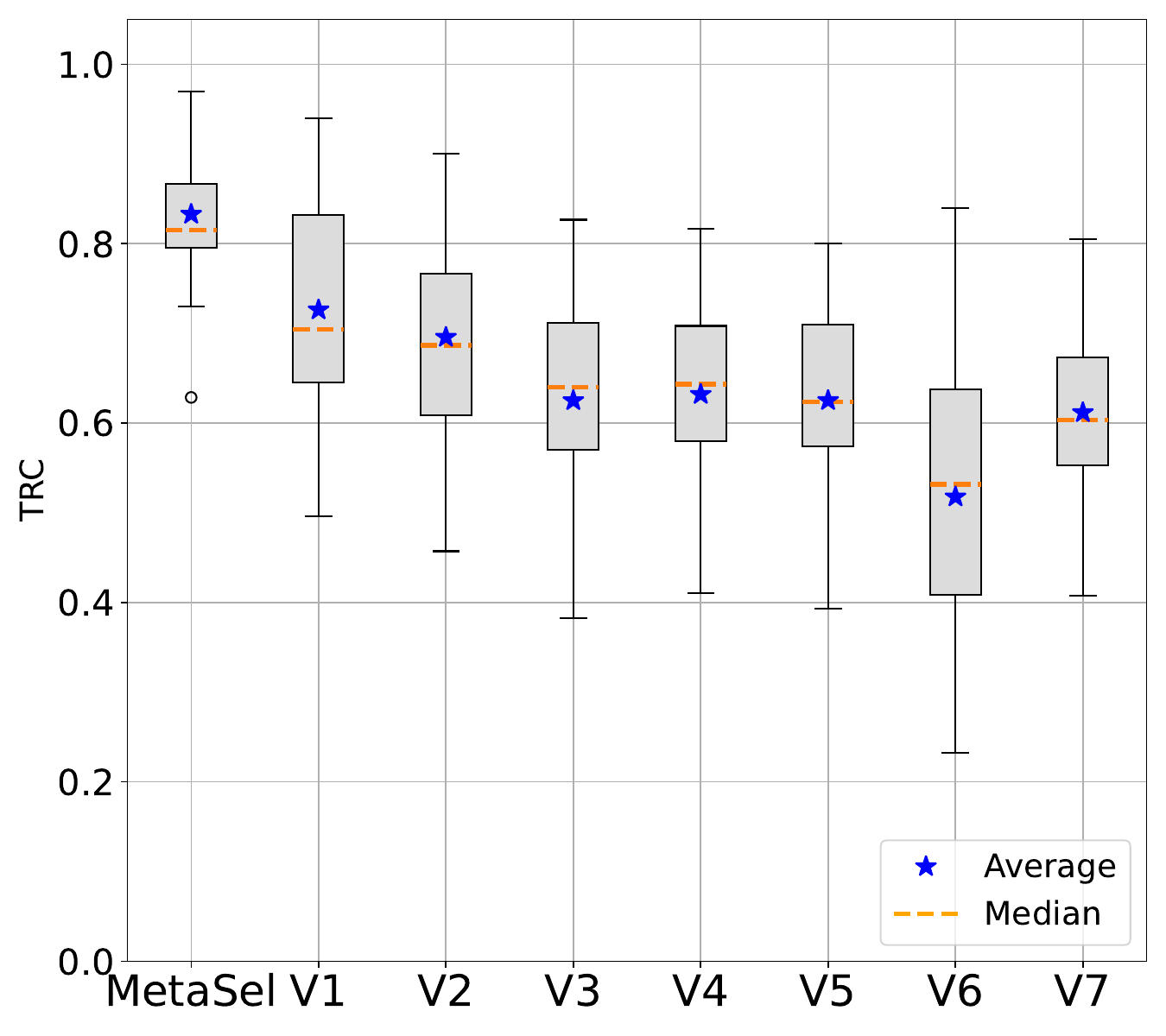}
    \par\vspace{-3pt}
         $b$ = 3\%
  \end{minipage}
  \hfill
    \par\vspace{5pt}
  \begin{minipage}[b]{0.48\linewidth}
    \centering
    \includegraphics[width=\linewidth]{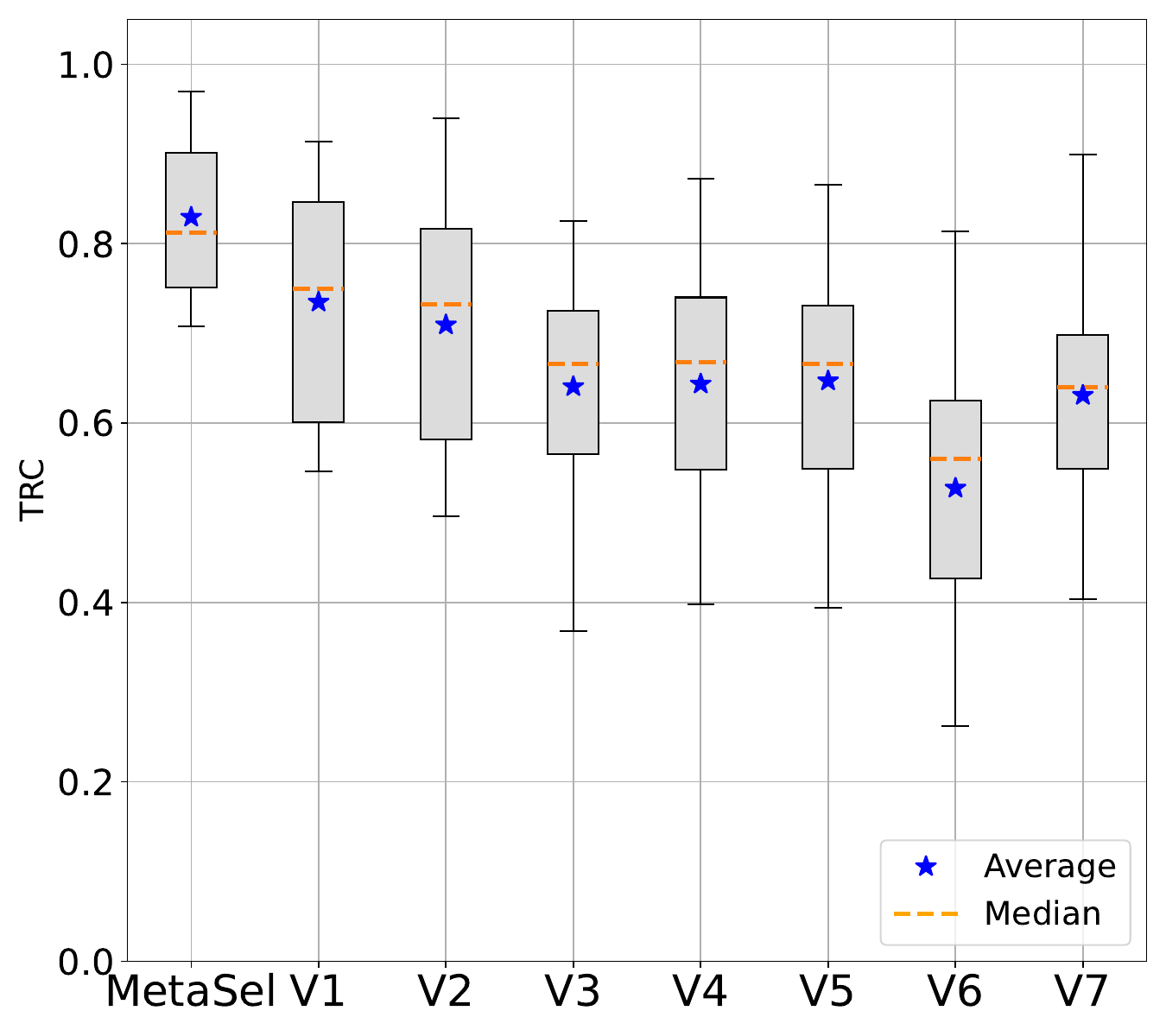}
    \par\vspace{-3pt}
     $b$ = 5\%
  \end{minipage}
  \hfill
  \begin{minipage}[b]{0.48\linewidth}
    \centering
    \includegraphics[width=\linewidth]{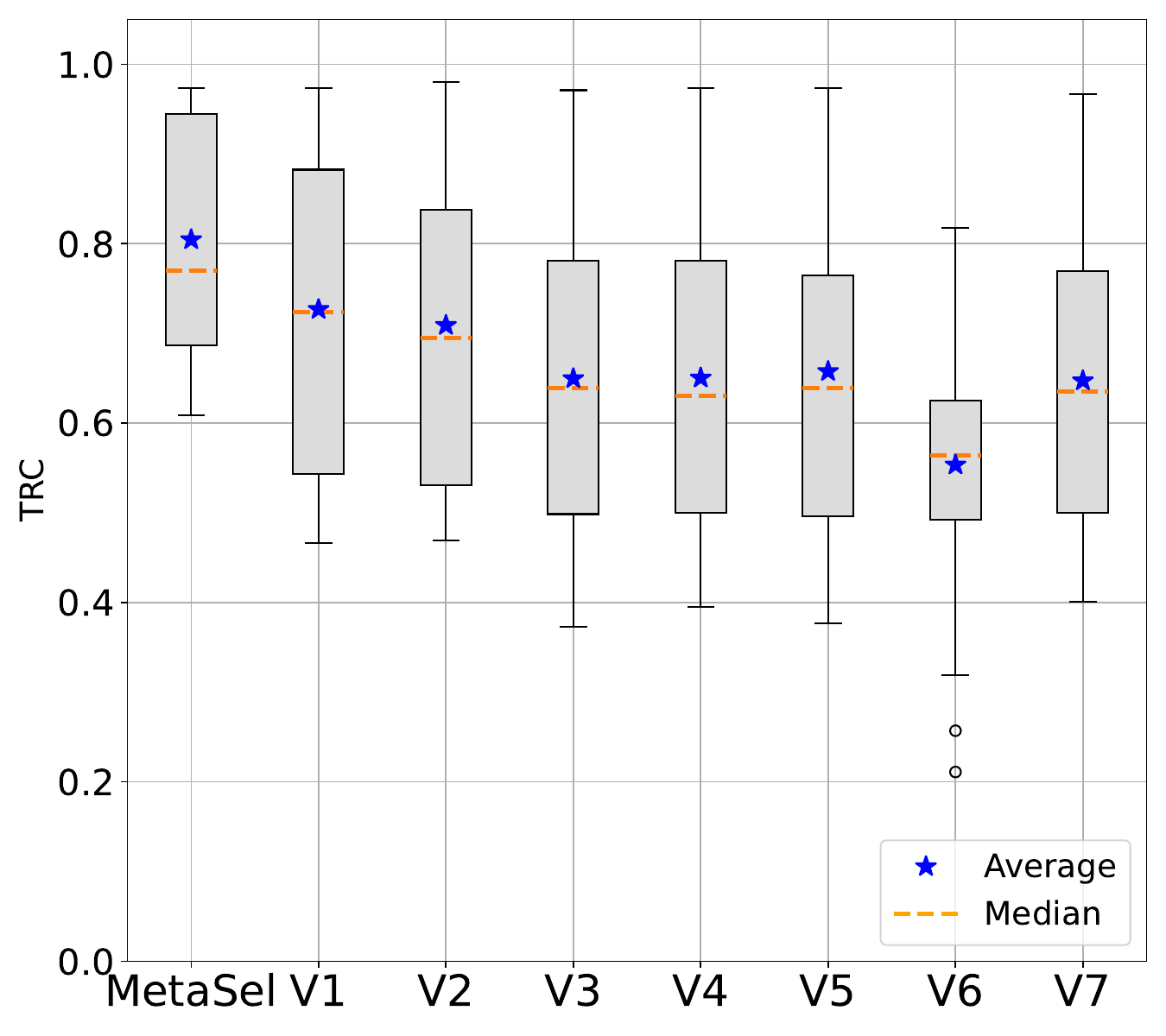}
    \par\vspace{-3pt}
    $b$ = 10\%
  \end{minipage}
    \par\vspace{5pt}

    \caption{TRC distribution achieved by MetaSel and its seven variants across all subjects under highly constrained selection budgets, across subjects representing a medium severity level of distribution shift (severity level = 3)} 
  \label{fig:ResultsRQ4_boxplots}
\end{figure}

\Rev{
Note that we performed our experiments only on subjects representing a medium level of distribution shift (i.e., severity level 3). This choice is based on our earlier findings from RQ3, which demonstrated that MetaSel consistently maintains its effectiveness across different levels of distribution shift severity. %
To answer this question, we compare the TRC results achieved by MetaSel with those of variants across all subjects (i.e., input sets, DNN models, and corruptions presented in Table~\ref{tab:RQ3TRCimprovements}). We compare the TRC distributions achieved by MetaSel and all variants under highly constrained budgets of 1\%, 3\%, 5\%, and 10\% of the target test set, illustrated in Figure~\ref{fig:ResultsRQ4_boxplots}.
To provide a comprehensive insight, we report the complete TRC curves achieved by all variants in our replication package~\cite{replicationpackage}. }

\Rev{Each box plot in Figure~\ref{fig:ResultsRQ4_boxplots}, presents the distribution of TRC achieved by MetaSel and its ablated variants under a given selection budget $b$.
MetaSel consistently achieves the highest TRC median and average values across all budget sizes, validating its superior effectiveness in identifying misclassified inputs compared to any of its variants. MetaSel’s lower TRC variance compared to its variants further supports its reliability across diverse subjects. 
These results confirm that removing any of the training features leads to significant performance degradation, thereby validating the essential contribution of each individual feature used in training MetaSel.
}

\begin{tcolorbox}   
    \TR{C1.4, C2.2, C3.2}{\textbf{Answer to RQ4: Results of our comprehensive ablation study comparing MetaSel with seven ablated variants across all subjects confirm that all features selected for constructing MetaSel significantly contribute to enhancing its performance.} }
\end{tcolorbox}

\subsection{Threats to Validity}
In this section, we discuss the different threats to the validity of our study and describe how we mitigated them.
\textbf{Internal threats to validity.}
One internal threat to validity arises from the parameters and configurations used in our experiments. This includes the architecture and hyperparameters of MetaSel. To address this concern, we performed hyperparameter tuning using a validation set to optimize MetaSel’s performance. Similarly, for the baseline configurations, we adhered to the setups specified in the original papers for each baseline.
For the calculation of SA metrics, we followed the configuration described in the original study~\cite{kim2023evaluating, kim2019guiding}. Specifically, we utilized the deepest layer of the DNN model.

\textbf{Construct threats to validity.}
One potential threat to validity lies in the implementation of test selection baselines. 
Where available, we used the implementations provided by the respective authors. For the SA metrics, however, we utilized an alternative implementation offered by Weiss \textit{et al.}~\cite{weiss2021review}, instead of the original implementation proposed by Kim \textit{et al.}~\cite{kim2019guiding}. This alternative implementation, which is significantly faster than the original, has been thoroughly evaluated by Weiss \textit{et al.}~\cite{weiss2021review}, demonstrating results comparable to those of the original implementation.
For NNS~\cite{bao2023defense}, since the authors did not provide an implementation, we developed the approach ourselves based on its detailed description in the original paper. We carefully reviewed our implementation to ensure its validation.
Another threat relates to the approach we used for measuring the alignment of each test input with the data distribution that the pre-trained and fine-tuned models were trained on. In our experiments, we used the ODIN score, a widely recognized and efficient OOD detection method. However, more advanced OOD detection techniques could potentially improve the results achieved by MetaSel.

\textbf{External threats to validity.}
To thoroughly evaluate the generalizability of MetaSel, we conducted extensive experiments involving 68 fine-tuned models. These models were derived from diverse input sets, including larger datasets like Cifar-100, and featured DNNs with varying internal architectures. To simulate varying degrees of distribution shift between source and target input sets, we incorporated seven types of image corruption, each applied at three severity levels. Furthermore, MetaSel was tested across a wide range of input selection budgets to assess its adaptability to different testing constraints. The results consistently demonstrated significant TRC improvements achieved by MetaSel across all combinations of input sets, model architectures, selection budgets, corruption types, and severity levels, suggesting strong generalizability across diverse testing scenarios.

\section{Related Work}
In this section, we review prior work related to our proposed test selection approach, within the context of DNN models, with a focus on two areas: general test selection for DNNs and learning-based test selection approaches. 
Existing test selection approaches for DNNs are general in nature, as they rely solely on MUT and its associated training set, without leveraging additional contextual information. In contrast, MetaSel is specifically tailored for fine-tuned models.
In addition, MetaSel is a learning-based approach, thus it is essential to specifically consider learning-based methods when evaluating MetaSel's design, requirements, and performance.

\textbf{Test selection for DNNs.} 
In recent years, several test selection approaches have been proposed for DNN models aiming to identify unlabeled inputs with a higher probability of being mispredicted by the model. These approaches are broadly classified into black-box and white-box approaches, depending on whether they require access to the internals of the DNN model~\cite{aghababaeyan2024deepgd}. Neuron coverage (NC) metrics~\cite{pei2017deepxplore, Ma2018DeepGaugeMT}, a class of white-box approaches that have been proposed for DNNs, are inspired by the success of code coverage in traditional software. However, studies have consistently shown that these criteria lack a significant correlation with the number of mispredicted inputs in a test set~\cite{aghababaeyan2023black, li2019structural, yang2022revisiting}.

Another early contribution in this category is SA metrics proposed by Kim \textit{et al.}~\cite{kim2019guiding}. These metrics measure how surprising a test input is with respect to the DNN model's training input set. Test inputs are then ranked in ascending order based on their SA scores, with more surprising inputs given higher priority for testing.  
Our results confirm earlier findings~\cite{Weiss2022SimpleTechniques, bao2023defense} that DSA delivers better performance compared to the other alternative SA metrics. However, MetaSel consistently outperforms all SA metrics across all subjects and distribution shift levels.

Black-box approaches rely on the output of MUT to guide test selection. For instance, ATS introduced by Gao \textit{et al.}~\cite{gao2022adaptive}, is a diversity-driven approach that leverages the variations in model outputs as a behavioral diversity metric. By focusing on these differences, ATS aims to identify and cover a broader range of potential failures. However, ATS involves computing distances between inputs and has been demonstrated to be one of the most computationally intensive approaches~\cite{hu2023evaluating, li2024distance}, making it impractical, particularly when dealing with large input sets. One of the most effective and widely used black-box approaches for test input prioritization is probability-based uncertainty metrics~\cite{feng2020deepgini, Weiss2022SimpleTechniques, hu2024test, ma2021test}. Their simplicity and reliance solely on MUT's output probability vector make them highly efficient.
In training MetaSel, however, we utilize the raw outputs of the logit layer instead of the probability vector, as logits enable more fine-grained comparisons between the pre-trained and fine-tuned DNN models. 
\Rev{Our comprehensive ablation study on selected features further validates this choice, as detailed in Section~\ref{sec:ablationStudy}. }
Our results confirm the efficiency and effectiveness of Probability-based uncertainty metrics. They also demonstrate that MetaSel consistently outperforms these approaches across subjects.

Relying on MUT's output probability vector, probability-based uncertainty metrics like DeepGini, Margin, and Vanilla are reported to be prone to miscalibration~\cite{guo2017Calibration}, where the model may assign high probabilities to incorrect predictions or low probabilities to correct ones~\cite{nguyen2015deep, goodfellow2014explaining, Kuleshov2015}. 
To address this, recent studies have explored leveraging information from a test input's nearest neighbors to provide more accurate uncertainty estimates~\cite{li2024distance, bao2023defense}.
NNS introduced by Bao \textit{et al.}~\cite{bao2023defense}, combines the DNN's prediction on a test input with its predictions for the input's nearest neighbors within the test set. Similarly, DATIS proposed by Li \textit{et al.}~\cite{li2024distance}, calculates uncertainty solely based on the ground truth labels of the test input's nearest neighbors in the labeled training set. Our findings confirm that while these approaches offer slight improvements in test selection effectiveness over probability-based metrics for certain subjects, their performance remains overall comparable.
As shown in Table~\ref{tab:TRCimprovementsSeverity3}, among the 92 combinations of subjects and budget sizes, DATIS and NNS emerged as the second-best approaches in 27 and 15 cases, respectively. However, despite their occasional strong performance, these methods exhibit inconsistencies across different subjects and budget sizes.

These inconsistencies have also been reported by Bao \textit{et al. }~\cite{bao2023defense} in their original study. They also observed that 
NNS's effectiveness heavily depends on the method used for extracting the latent representation of inputs. Their evaluation, which includes using the MUT itself as the representation extractor and employing a dedicated extractor obtained through unsupervised learning with auxiliary pre-trained models, such as BYOL~\cite{grill2020bootstrap}, revealed significant performance variations across input sets and DNN models. 
In contrast, MetaSel's superior performance remains consistent across all subjects, selection budgets, and levels of distribution shift, confirming its reliability as the most effective test selection approach.

\TR{C1.6, C3.8}{
RTS proposed by Sun \textit{et al.}~\cite{sun2023robust} combines probability-based uncertainty with diversity for test prioritization. Like other diversity-based methods, it relies on pairwise similarity computations (based on SSIM similarity~\cite{wang2004image}), resulting in high computational overhead. As shown in our results, RTS is the most time-consuming approach, taking approximately 4 to 15 times longer than MetaSel to prioritize the entire test set across our subjects. Moreover, RTS is consistently outperformed by MetaSel in terms of effectiveness and never appears among the second-best baselines. 
}


\textbf{Learning-based test selection for DNNs.} 
In recent years, learning-based approaches have emerged, leveraging the concept of training an independent model on MUT's outputs to predict misprediction probabilities for each test input. While all learning-based approaches share a common foundation, they differ in the types of features used for training. 
TestRank~\cite{li2021testrank}, for instance, combines intrinsic features from MUT's logit layer and contextual features from the test input's $k$-nearest neighbors.
Meta-model, proposed by Demir \textit{et al.}~\cite{demir2024test}, enhances uncertainty scores by incorporating predictions from deep ensemble models, as described in Section~\ref{sec:BackgroundLearningBased}.
PRIMA, proposed by Wang \textit{et al.}~\cite{wang2021prioritizing}, introduces another learning-based approach leveraging the mutation testing technique. PRIMA employs two types of mutation rules, model mutation and input mutation, to generate mutated outputs for each test input. Features are then extracted by comparing the outputs of the MUT with those of the mutated models, considering also differences in their probability distributions. These features are used to train a ranking model using the XGBoost algorithm, enabling PRIMA to prioritize test inputs effectively by identifying those more likely to reveal faults. However, mutation testing is widely recognized as computationally intensive, and PRIMA’s reliance on this technique makes its execution particularly time-consuming~\cite{hu2023evaluating}. Furthermore, simpler and computationally more efficient approaches such as DeepGini have shown to consistently outperform PRIMA in terms of effectiveness~\cite{hu2023evaluating}. Consequently, we exclude PRIMA from our experiments.

In contrast, MetaSel is uniquely tailored for the context of fine-tuned models, leveraging a richer and more context-specific feature set. It incorporates the logits of both the pre-trained and fine-tuned models, their differences, and the results of comparing their predicted output labels. Additionally, MetaSel uses a measure of how well an input aligns with the data distributions of both the pre-trained and fine-tuned models. By utilizing insights from both DNN models and their underlying data distributions, MetaSel consistently outperforms SOTA methods, including investigated learning-based approaches. Moreover, enhancing MetaSel's training set with inputs from the pre-trained model's test set further improves its accuracy, highlighting the importance of leveraging these additional artifacts specifically available in the context of the fine-tuned models.

All learning-based approaches involve an initial training phase that includes setup and preparation tasks, such as feature extraction and model training.  While execution time is often reported when comparing the efficiency of approaches, the associated effort and computational cost of this initial phase are equally important but often difficult to quantify. These setup costs can vary widely, depending on the complexity of the feature extraction process and the specific requirements for model training. 
For instance, PRIMA, built on mutation testing, necessitates generating both model and input mutations, executing each input against all mutations, and extracting features from the outcomes. However, despite its sophisticated setup, PRIMA has not demonstrated superior performance compared to simpler probability-based uncertainty metrics~\cite{hu2023evaluating}. 
TestRank also entails a significant setup phase, as it involves extracting inputs' representation utilizing pre-trained models such as BYOL~\cite{grill2020bootstrap}. It then creates a similarity graph by calculating distances between each test input and all inputs in both training and test sets to identify its nearest neighbors. Additionally, it trains a GNN to process these features, adding another layer of complexity and increasing the overall training cost. 
Similarly, Meta-model, proposed by Demir \textit{et al.}~\cite{demir2024test} requires training multiple deep ensemble models and then executing each input against all these models to calculate the DE variation score. This adds considerable computational overhead, particularly for deeper DNN models such as ResNet152.
In contrast, MetaSel offers a more straightforward and efficient approach by leveraging the pre-trained model directly for feature extraction. As previously reported, for our most complex subject DNN model, ResNet152 with the Cifar-100 input set, the initial training phase of MetaSel and Meta-model took approximately 15 and 130 minutes, respectively.

Our results demonstrate that MetaSel consistently outperforms all the investigated SOTA general test selection approaches including white-box, black-box, and learning-based baselines across all subjects. This highlights the significant potential of leveraging additional contextual information, when available, to improve test input prioritization and selection.
Furthermore, our empirical study on \Rev{11} baselines reveals that the most frequent second-best approach after MetaSel is Meta-model, a learning-based approach. This further highlights the growing importance and effectiveness of learning-based techniques, suggesting that these approaches can play a pivotal role in enhancing test input selection effectiveness.

\section{Conclusion}
In this paper, we introduce MetaSel, a test input prioritization and selection method specifically designed for fine-tuned DNN models. Our approach leverages information about the relationship between the fine-tuned model and its pre-trained counterpart to estimate misclassification probabilities for unlabeled test inputs. MetaSel relies on the logits of both the pre-trained and fine-tuned models, their differences, and the comparison of their predicted output labels. Additionally, MetaSel employs a measure of how well an input aligns with the data distributions on which both the pre-trained and fine-tuned models were trained.
Our empirical analysis, comparing MetaSel against \Rev{11} SOTA test selection approaches, demonstrates that MetaSel consistently outperforms all baselines, particularly under tightly constrained labeling budgets, which is especially important when testing fine-tuned DNN models.
The results show that MetaSel is capable of detecting a significantly higher number of misclassified inputs within the same selection budget. 
In our extensive empirical study, involving 68 subjects across varying levels of distribution shift severity between pre-trained and fine-tuned models, MetaSel's performance remained consistent regardless of subject, selection budget, and distribution shift severity level. Further, the second-best performing approach after MetaSel varies across different subjects and selection budgets, further confirming that MetaSel is the best solution. Additionally, MetaSel not only achieves higher TRC median and average but also maintains significantly lower variability in TRC values than alternative baselines across all subjects.
Last, when assessing the computational efficiency of MetaSel in comparison to baselines, we demonstrate MetaSel’s practicality and cost-effectiveness, even when applied to large input sets and complex DNN models.
Although MetaSel is inherently adaptable and applicable to a wide range of input types, we plan to extend our investigation to include input sets from diverse application domains, \Rev{such as text inputs}, to determine how well MetaSel maintains its effectiveness across different data domains.
\MinRev{Another direction for our future work is to explore the potential of extending and adapting MetaSel to transfer learning scenarios beyond covariate shift, encompassing different but related source and target domains that still result in successful fine-tuning. Although MetaSel in its original design may not be directly applicable to all such scenarios, adaptation strategies could enable its broader applicability. For instance, when the source and target input sets feature different output classes, a mapping between these classes would enable the extraction of MetaSel’s features, e.g., calculating the differential testing outcome and measuring the differences between the source and target logit vectors. }




\Rev{\section*{Acknowledgements}
This work was supported by a research grant from Huawei Technologies Canada, as well as the Canada Research Chair and Discovery Grant programs of the Natural Sciences and Engineering Research Council of Canada (NSERC) and the Research Ireland grant 13/RC/2094-2. The experiments conducted in this work were enabled in part by support provided by the Digital Research Alliance of Canada~\footnote{https://alliancecan.ca}.}

\bibliographystyle{IEEEtran}
\bibliography{main.bib}

\begin{thebibliography}{10}
\providecommand{\url}[1]{#1}
\csname url@samestyle\endcsname
\providecommand{\newblock}{\relax}
\providecommand{\bibinfo}[2]{#2}
\providecommand{\BIBentrySTDinterwordspacing}{\spaceskip=0pt\relax}
\providecommand{\BIBentryALTinterwordstretchfactor}{4}
\providecommand{\BIBentryALTinterwordspacing}{\spaceskip=\fontdimen2\font plus
\BIBentryALTinterwordstretchfactor\fontdimen3\font minus \fontdimen4\font\relax}
\providecommand{\BIBforeignlanguage}[2]{{%
\expandafter\ifx\csname l@#1\endcsname\relax
\typeout{** WARNING: IEEEtran.bst: No hyphenation pattern has been}%
\typeout{** loaded for the language `#1'. Using the pattern for}%
\typeout{** the default language instead.}%
\else
\language=\csname l@#1\endcsname
\fi
#2}}
\providecommand{\BIBdecl}{\relax}
\BIBdecl

\bibitem{pan2010survey}
S.~Pan and Q.~Yang, ``A survey on transfer learning. ieee transaction on knowledge discovery and data engineering, 22 (10),'' 2010.

\bibitem{hu2024test}
Q.~Hu, Y.~Guo, X.~Xie, M.~Cordy, L.~Ma, M.~Papadakis, and Y.~Le~Traon, ``Test optimization in dnn testing: a survey,'' \emph{ACM Transactions on Software Engineering and Methodology}, vol.~33, no.~4, pp. 1--42, 2024.

\bibitem{aghababaeyan2024deepgd}
Z.~Aghababaeyan, M.~Abdellatif, M.~Dadkhah, and L.~Briand, ``Deepgd: A multi-objective black-box test selection approach for deep neural networks,'' \emph{ACM Transactions on Software Engineering and Methodology}, vol.~33, no.~6, pp. 1--29, 2024.

\bibitem{kim2019guiding}
J.~Kim, R.~Feldt, and S.~Yoo, ``Guiding deep learning system testing using surprise adequacy,'' in \emph{2019 IEEE/ACM 41st International Conference on Software Engineering (ICSE)}.\hskip 1em plus 0.5em minus 0.4em\relax IEEE, 2019, pp. 1039--1049.

\bibitem{kim2023evaluating}
{J. Kim}, {R. Feldt}, and {S. Yoo}, ``Evaluating surprise adequacy for deep learning system testing,'' \emph{ACM Transactions on Software Engineering and Methodology}, vol.~32, no.~2, pp. 1--29, 2023.

\bibitem{feng2020deepgini}
Y.~Feng, Q.~Shi, X.~Gao, J.~Wan, C.~Fang, and Z.~Chen, ``Deepgini: prioritizing massive tests to enhance the robustness of deep neural networks,'' in \emph{Proceedings of the 29th ACM SIGSOFT International Symposium on Software Testing and Analysis}, 2020, pp. 177--188.

\bibitem{Weiss2022SimpleTechniques}
M.~Weiss and P.~Tonella, ``Simple techniques work surprisingly well for neural network test prioritization and active learning,'' in \emph{Proceedings of the 31th ACM SIGSOFT International Symposium on Software Testing and Analysis}, 2022.

\bibitem{ma2021test}
W.~Ma, M.~Papadakis, A.~Tsakmalis, M.~Cordy, and Y.~L. Traon, ``Test selection for deep learning systems,'' \emph{ACM Transactions on Software Engineering and Methodology (TOSEM)}, vol.~30, no.~2, pp. 1--22, 2021.

\bibitem{li2024distance}
Z.~Li, Z.~Xu, R.~Ji, M.~Pan, T.~Zhang, L.~Wang, and X.~Li, ``Distance-aware test input selection for deep neural networks,'' in \emph{Proceedings of the 33rd ACM SIGSOFT International Symposium on Software Testing and Analysis}, 2024, pp. 248--260.

\bibitem{bao2023defense}
S.~Bao, C.~Sha, B.~Chen, X.~Peng, and W.~Zhao, ``In defense of simple techniques for neural network test case selection,'' in \emph{Proceedings of the 32nd ACM SIGSOFT International Symposium on Software Testing and Analysis}, 2023, pp. 501--513.

\bibitem{demir2024test}
D.~Demir, A.~Betin~Can, and E.~Surer, ``Test selection for deep neural networks using meta-models with uncertainty metrics,'' in \emph{Proceedings of the 33rd ACM SIGSOFT International Symposium on Software Testing and Analysis}, 2024, pp. 678--690.

\bibitem{li2021testrank}
Y.~Li, M.~Li, Q.~Lai, Y.~Liu, and Q.~Xu, ``Testrank: Bringing order into unlabeled test instances for deep learning tasks,'' \emph{Advances in Neural Information Processing Systems}, vol.~34, pp. 20\,874--20\,886, 2021.

\bibitem{computecanada}
``Digital research alliance of canada,'' \url{https://alliancecan.ca/}, 2016, accessed: March 3, 2025.

\bibitem{wang2021prioritizing}
Z.~Wang, H.~You, J.~Chen, Y.~Zhang, X.~Dong, and W.~Zhang, ``Prioritizing test inputs for deep neural networks via mutation analysis,'' in \emph{2021 IEEE/ACM 43rd International Conference on Software Engineering (ICSE)}.\hskip 1em plus 0.5em minus 0.4em\relax IEEE, 2021, pp. 397--409.

\bibitem{hu2023evaluating}
Q.~Hu, Y.~Guo, X.~Xie, M.~Cordy, W.~Ma, M.~Papadakis, and Y.~L. Traon, ``Evaluating the robustness of test selection methods for deep neural networks,'' \emph{arXiv preprint arXiv:2308.01314}, 2023.

\bibitem{wand1994kernel}
M.~P. Wand and M.~C. Jones, \emph{Kernel smoothing}.\hskip 1em plus 0.5em minus 0.4em\relax CRC press, 1994.

\bibitem{gao2022adaptive}
X.~Gao, Y.~Feng, Y.~Yin, Z.~Liu, Z.~Chen, and B.~Xu, ``Adaptive test selection for deep neural networks,'' in \emph{Proceedings of the 44th International Conference on Software Engineering}, 2022, pp. 73--85.

\bibitem{dang2023graphprior}
X.~Dang, Y.~Li, M.~Papadakis, J.~Klein, T.~F. Bissyand{\'e}, and Y.~Le~Traon, ``Graphprior: Mutation-based test input prioritization for graph neural networks,'' \emph{ACM Transactions on Software Engineering and Methodology}, vol.~33, no.~1, pp. 1--40, 2023.

\bibitem{hu2022empirical}
Q.~Hu, Y.~Guo, M.~Cordy, X.~Xie, L.~Ma, M.~Papadakis, and Y.~Le~Traon, ``An empirical study on data distribution-aware test selection for deep learning enhancement,'' \emph{ACM Transactions on Software Engineering and Methodology (TOSEM)}, vol.~31, no.~4, pp. 1--30, 2022.

\bibitem{sun2023robust}
W.~Sun, M.~Yan, Z.~Liu, and D.~Lo, ``Robust test selection for deep neural networks,'' \emph{IEEE Transactions on Software Engineering}, vol.~49, no.~12, pp. 5250--5278, 2023.

\bibitem{wang2004image}
Z.~Wang, A.~C. Bovik, H.~R. Sheikh, and E.~P. Simoncelli, ``Image quality assessment: from error visibility to structural similarity,'' \emph{IEEE transactions on image processing}, vol.~13, no.~4, pp. 600--612, 2004.

\bibitem{replicationpackage}
\BIBentryALTinterwordspacing
A.~Abbasishahkoo, M.~Dadkhah, L.~Briand, and D.~Lin. [Online]. Available: \url{https://figshare.com/s/b4b1338873351ca01734}
\BIBentrySTDinterwordspacing

\bibitem{liang2017enhancing}
S.~Liang, Y.~Li, and R.~Srikant, ``Enhancing the reliability of out-of-distribution image detection in neural networks,'' \emph{arXiv preprint arXiv:1706.02690}, 2017.

\bibitem{xiao2020likelihood}
Z.~Xiao, Q.~Yan, and Y.~Amit, ``Likelihood regret: An out-of-distribution detection score for variational auto-encoder,'' \emph{Advances in neural information processing systems}, vol.~33, pp. 20\,685--20\,696, 2020.

\bibitem{aigrain2019detecting}
J.~Aigrain and M.~Detyniecki, ``Detecting adversarial examples and other misclassifications in neural networks by introspection,'' \emph{arXiv preprint arXiv:1905.09186}, 2019.

\bibitem{simonyan2014very}
K.~Simonyan and A.~Zisserman, ``Very deep convolutional networks for large-scale image recognition,'' \emph{arXiv preprint arXiv:1409.1556}, 2014.

\bibitem{zhao2022uniform}
F.~Zhao, C.~Zhang, N.~Dong, Z.~You, and Z.~Wu, ``A uniform framework for anomaly detection in deep neural networks,'' \emph{Neural Processing Letters}, vol.~54, no.~4, pp. 3467--3488, 2022.

\bibitem{deng2012mnist}
L.~Deng, ``The mnist database of handwritten digit images for machine learning research [best of the web],'' \emph{IEEE Signal Processing Magazine}, vol.~29, no.~6, pp. 141--142, 2012.

\bibitem{krizhevsky2009learning}
A.~Krizhevsky and G.~Hinton, ``Learning multiple layers of features from tiny images,'' University of Toronto, Tech. Rep., 2009.

\bibitem{shen2020multiple}
W.~Shen, Y.~Li, L.~Chen, Y.~Han, Y.~Zhou, and B.~Xu, ``Multiple-boundary clustering and prioritization to promote neural network retraining,'' in \emph{Proceedings of the 35th IEEE/ACM International Conference on Automated Software Engineering}, 2020, pp. 410--422.

\bibitem{abbasishahkoo2024teasma}
A.~Abbasishahkoo, M.~Dadkhah, L.~Briand, and D.~Lin, ``Teasma: A practical methodology for test adequacy assessment of deep neural networks,'' \emph{IEEE Transactions on Software Engineering}, 2024.

\bibitem{berend2020cats}
D.~Berend, X.~Xie, L.~Ma, L.~Zhou, Y.~Liu, C.~Xu, and J.~Zhao, ``Cats are not fish: Deep learning testing calls for out-of-distribution awareness,'' in \emph{Proceedings of the 35th IEEE/ACM international conference on automated software engineering}, 2020, pp. 1041--1052.

\bibitem{lecun1998gradient}
Y.~LeCun, L.~Bottou, Y.~Bengio, and P.~Haffner, ``Gradient-based learning applied to document recognition,'' \emph{Proceedings of the IEEE}, vol.~86, no.~11, pp. 2278--2324, 1998.

\bibitem{he2016deep}
K.~He, X.~Zhang, S.~Ren, and J.~Sun, ``Deep residual learning for image recognition,'' in \emph{Proceedings of the IEEE conference on computer vision and pattern recognition}, 2016, pp. 770--778.

\bibitem{hendrycks2019benchmarking}
D.~Hendrycks and T.~Dietterich, ``Benchmarking neural network robustness to common corruptions and perturbations,'' \emph{arXiv preprint arXiv:1903.12261}, 2019.

\bibitem{mu2019mnist}
N.~Mu and J.~Gilmer, ``Mnist-c: A robustness benchmark for computer vision,'' \emph{arXiv preprint arXiv:1906.02337}, 2019.

\bibitem{pan2009survey}
S.~J. Pan and Q.~Yang, ``A survey on transfer learning,'' \emph{IEEE Transactions on knowledge and data engineering}, vol.~22, no.~10, pp. 1345--1359, 2009.

\bibitem{xu2020transfer}
W.~Xu, J.~He, and Y.~Shu, ``Transfer learning and deep domain adaptation,'' \emph{Advances and applications in deep learning}, vol.~45, 2020.

\bibitem{rothermel2001prioritizing}
G.~Rothermel, R.~H. Untch, C.~Chu, and M.~J. Harrold, ``Prioritizing test cases for regression testing,'' \emph{IEEE Transactions on software engineering}, vol.~27, no.~10, pp. 929--948, 2001.

\bibitem{liang2018redefining}
J.~Liang, S.~Elbaum, and G.~Rothermel, ``Redefining prioritization: continuous prioritization for continuous integration,'' in \emph{Proceedings of the 40th International Conference on Software Engineering}, 2018, pp. 688--698.

\bibitem{weiss2021review}
M.~Weiss, R.~Chakraborty, and P.~Tonella, ``A review and refinement of surprise adequacy,'' in \emph{2021 IEEE/ACM Third International Workshop on Deep Learning for Testing and Testing for Deep Learning (DeepTest)}.\hskip 1em plus 0.5em minus 0.4em\relax IEEE, 2021, pp. 17--24.

\bibitem{pei2017deepxplore}
K.~Pei, Y.~Cao, J.~Yang, and S.~Jana, ``Deepxplore: Automated whitebox testing of deep learning systems,'' in \emph{proceedings of the 26th Symposium on Operating Systems Principles}, 2017, pp. 1--18.

\bibitem{Ma2018DeepGaugeMT}
L.~Ma, F.~Juefei-Xu, F.~Zhang, J.~Sun, M.~Xue, B.~Li, C.~Chen, T.~Su, L.~Li, Y.~Liu, J.~Zhao, and Y.~Wang, ``Deepgauge: Multi-granularity testing criteria for deep learning systems,'' \emph{2018 33rd IEEE/ACM International Conference on Automated Software Engineering (ASE)}, pp. 120--131, 2018.

\bibitem{aghababaeyan2023black}
Z.~Aghababaeyan, M.~Abdellatif, L.~Briand, S.~Ramesh, and M.~Bagherzadeh, ``Black-box testing of deep neural networks through test case diversity,'' \emph{IEEE Transactions on Software Engineering}, vol.~49, no.~5, pp. 3182--3204, 2023.

\bibitem{li2019structural}
Z.~Li, X.~Ma, C.~Xu, and C.~Cao, ``Structural coverage criteria for neural networks could be misleading,'' in \emph{2019 IEEE/ACM 41st International Conference on Software Engineering: New Ideas and Emerging Results (ICSE-NIER)}.\hskip 1em plus 0.5em minus 0.4em\relax IEEE, 2019, pp. 89--92.

\bibitem{yang2022revisiting}
Z.~Yang, J.~Shi, M.~H. Asyrofi, and D.~Lo, ``Revisiting neuron coverage metrics and quality of deep neural networks,'' in \emph{2022 IEEE International Conference on Software Analysis, Evolution and Reengineering (SANER)}.\hskip 1em plus 0.5em minus 0.4em\relax IEEE, 2022, pp. 408--419.

\bibitem{guo2017Calibration}
C.~Guo, G.~Pleiss, Y.~Sun, and K.~Q. Weinberger, ``On calibration of modern neural networks,'' in \emph{Proceedings of the 34th International Conference on Machine Learning}, ser. Proceedings of Machine Learning Research, D.~Precup and Y.~W. Teh, Eds., vol.~70.\hskip 1em plus 0.5em minus 0.4em\relax PMLR, 06--11 Aug 2017, pp. 1321--1330.

\bibitem{nguyen2015deep}
A.~Nguyen, J.~Yosinski, and J.~Clune, ``Deep neural networks are easily fooled: High confidence predictions for unrecognizable images,'' in \emph{Proceedings of the IEEE conference on computer vision and pattern recognition}, 2015, pp. 427--436.

\bibitem{goodfellow2014explaining}
I.~J. Goodfellow, J.~Shlens, and C.~Szegedy, ``Explaining and harnessing adversarial examples,'' \emph{arXiv preprint arXiv:1412.6572}, 2014.

\bibitem{Kuleshov2015}
V.~Kuleshov and P.~S. Liang, ``Calibrated structured prediction,'' in \emph{Advances in Neural Information Processing Systems}, C.~Cortes, N.~Lawrence, D.~Lee, M.~Sugiyama, and R.~Garnett, Eds., vol.~28.\hskip 1em plus 0.5em minus 0.4em\relax Curran Associates, Inc., 2015.

\bibitem{grill2020bootstrap}
J.-B. Grill, F.~Strub, F.~Altch{\'e}, C.~Tallec, P.~Richemond, E.~Buchatskaya, C.~Doersch, B.~Avila~Pires, Z.~Guo, M.~Gheshlaghi~Azar \emph{et~al.}, ``Bootstrap your own latent-a new approach to self-supervised learning,'' \emph{Advances in neural information processing systems}, vol.~33, pp. 21\,271--21\,284, 2020.

\end{thebibliography}

\end{document}